\documentclass[10pt,twocolumn,letterpaper]{article}

\usepackage[pagenumbers]{cvpr} % To force page numbers, e.g. for an arXiv version

\usepackage{graphicx}
\usepackage{amsmath}
\usepackage{amssymb}
\usepackage{booktabs}
\usepackage[accsupp]{axessibility}  % Improves PDF readability for those with disabilities.

\usepackage[pagebackref,breaklinks,colorlinks]{hyperref}

\usepackage{mathtools}
\usepackage{multirow}
\usepackage{threeparttable}
\usepackage{colortbl}
\DeclarePairedDelimiter\floor{\lfloor}{\rfloor}

\usepackage[capitalize]{cleveref}
\crefname{section}{Sec.}{Secs.}
\Crefname{section}{Section}{Sections}
\Crefname{table}{Table}{Tables}
\crefname{table}{Tab.}{Tabs.}

\def\cvprPaperID{6107} % *** Enter the CVPR Paper ID here
\def\confName{CVPR}
\def\confYear{2022}

\begin{document}

%%%%%%%%% TITLE - PLEASE UPDATE
\title{Deep Depth from Focus with Differential Focus Volume}

\author{Fengting Yang \quad  Xiaolei Huang \quad Zihan Zhou\\
The Pennsylvania State University\\
% Institution1 address\\
{\tt\small \{fuy34, suh972, zuz22\}@psu.edu}
}
% For a paper whose authors are all at the same institution,
% omit the following lines up until the closing ``}''.
% Additional authors and addresses can be added with ``\and'',
% just like the second author.
% To save space, use either the email address or home page, not both
% \and
% Xiaolei Huang\\
% Institution2\\
% First line of institution2 address\\
% {\tt\small secondauthor@i2.org}
% }
\maketitle

%%%%%%%%% ABSTRACT
\begin{abstract}
  Depth-from-focus (DFF) is a technique that infers depth using the focus change of a camera. In this work, we propose a convolutional neural network (CNN) to find the best-focused pixels in a focal stack and infer depth from the focus estimation.  The key innovation of the network is the novel deep differential focus volume (DFV). By computing the first-order derivative with the stacked features over different focal distances, DFV is able to capture both the focus and context information for focus analysis. Besides, we also introduce a probability regression mechanism for focus estimation to handle sparsely sampled focal stacks and provide uncertainty estimation to the final prediction.  Comprehensive experiments demonstrate that the proposed model achieves state-of-the-art performance on multiple datasets with good generalizability and fast speed. 
\end{abstract} %Our network consists of a shared 2D CNN that first extracts image features and a 3D CNN that aggregates the extracted features to infer the focus distribution. The depth is later computed through probability regression. 

%%%%%%%%% BODY TEXT
\section{Introduction}
\label{sec:intro}
 Recovering depth using a single RGB camera is a critical problem in 3D vision. Many applications can benefit from this technique, including 3D reconstruction, virtual and augmented reality, and image editing.  In the literature, various cues have been explored to tackle the problem, such as  focus~\cite{pentland1987firstDFF}, ego-motion~\cite{schonberger2016sfmColmap}, and structured-light patterns~\cite{albitar2007structurelight}. But cues like ego-motion and structured-light patterns require either extra motion or additional devices, which limit their applications on smart devices and hand-held cameras.  In contrast, the focus (or defocus) cue is what we can gain almost for ``free.'' To capture a well-focused image, many digital cameras rapidly sweep a focal plane in its focus range, resulting in a series of images (\ie, a focal stack) with different focal distances.  A pioneering work~\cite{suwajanakorn2015mobiledepth} has shown that depth can be inferred from focal stacks taken by a mobile phone.  
 
Image context is another ``free'' cue we can get from pictures. Based on it, a series of works~\cite{eigen2014depth, godard2019monov2, Ranftl2020Midas, ramamonjisoa2021single} have shown the ability to infer depth from a single image. But it is still a challenge to generalize these single-view methods to unknown scenes. Plus, due to the scale ambiguity, one cannot estimate the absolute depth from a single image without scene priors, even if the camera is well-calibrated.  
 
In this work, we utilize both the focus and context cues and develop a deep depth-from-focus (DFF) network for focus analysis and depth inference. We prefer DFF over the other focus-based technique, namely depth-from-defocus (DFD), because of its generalizability. For DFD, a mathematical model between the depth and the defocus pattern needs to be established. These methods commonly assume the object point is centralized at a small plane and the point spread function (PSF) follows a certain distribution~\cite{mannan2016good}. But such assumptions may not hold in real-world scenarios. DFF only assumes there is one and only one best-focused frame for a pixel in a focal stack, which is guaranteed in theory~\cite{schechner2000depth} for thin lens cameras if the sampling focal distances are dense enough. The focal distance of the best-focused frame can serve as the pixel depth estimation. 
 
Several major challenges still exist in DFF. The first one is the focus measure design. A large number of focus measures have been proposed, but none of them is perfect in the wild environment~\cite{pertuz2013overview}. The second one is regarding textureless regions. As the responses to focus measures stay low in textureless regions, the context information has to be used to infer the focus status~\cite{fan2018novel, ali2021guidedFinSFF}. The third one is the high sampling frequency requirement. In theory, for an object point falls into the depth of field of a camera, there must be one sharpest pixel in the focal sweep. But it may not be visible if the sampling frequency is low. Thus, many traditional DFF methods~\cite{surh2017noise, moeller2015variational} take tens of frames as input, which limits their running speed.
 
To overcome these challenges, we propose to learn a deep focus measure with the convolutional neural network (CNN). The global information embedded in the deep features extracted by the CNN can help focus analysis in textureless regions. Further, the focus status is represented as a probability distribution of the best-focused pixels, which can be learned from a limited number of frames. Note that earlier works~\cite{hazirbas2018DFFdataset, wang2021bridging, chen2021deep} have also tried to tackle the DFF problem with CNN. But their networks are adopted from either the general dense prediction tasks~\cite{hazirbas2018DFFdataset, chen2021deep} or the video representation task~\cite{wang2021bridging}, not specifically designed for DFF. 
 
% Our work is inspired the close relationship between stereo matching and DFF~\cite{schechner2000depth, lee2019complex}. A common practice in traditional stereo matching and DFF is to use a volume representation to store the matching score or in-focus score and process it with aggregation operations. Previous work~\cite{surh2017noise} has shown that we can utilize the same aggregation methods from stereo matching to enhance the DFF performance. Therefore, to design a network specifically for DFF, the first question we ask is {\it if we can adopt an existing design from deep stereo matching}. 
%  by the recent success of deep cost volume in stereo matching~\cite{chang2018PSM, yang2019hsm, mao2021uasnet} and
Our network is inspired by the close relationship between stereo matching and DFF~\cite{schechner2000depth, lee2019complex} and the recent success of deep stereo matching~\cite{chang2018PSM, yang2019hsm, mao2021uasnet}. The network employs a 2D CNN to learn the deep focus and context representations, stacks them into 4D focus volumes (FVs), and computes the differences in their frame dimension to build deep differential focus volumes (DFVs). The DFVs are later processed by a 3D CNN to predict the best-focused probability of pixels. The probabilistic prediction helps locate the best-focused frames for pixels even if they are not visible in the focal stack, thus decreasing the required focal stack size.  Finally, the depth estimation is obtained by probability regression.  Our pipeline is analogous to traditional FV-based DFF approaches~\cite{mahmood2012nonlinear, surh2017noise, fan2018novel}, which first compute the in-focus score using hand-crafted measures and then aggregate the score volume to obtain the final focus analysis for depth estimation.
Besides depth prediction, our model also provides an uncertainty estimation that shows the reliability of the prediction. To the best of our knowledge, this is the first work that introduces the deep 4D focus volume to DFF.

We have conducted comprehensive experiments to evaluate our method on both synthetic and real-world datasets. Our method achieves state-of-the-art performance while consuming a small number of frames (\eg, five) per focal stack, outperforming all DFF and DFD baselines. % the existing 
Our method also generalizes well to unknown scenes without fine-tuning. Further, the model runs at about 18.2 ms/stack with $256\times256$ input resolution and about 33.3 ms/stack with $383\times552$ input resolution on an NVIDIA 1080Ti GPU, making it suitable for potential real-time applications.

\begin{figure*}[htp]
\centering
\includegraphics[width=0.95\linewidth]{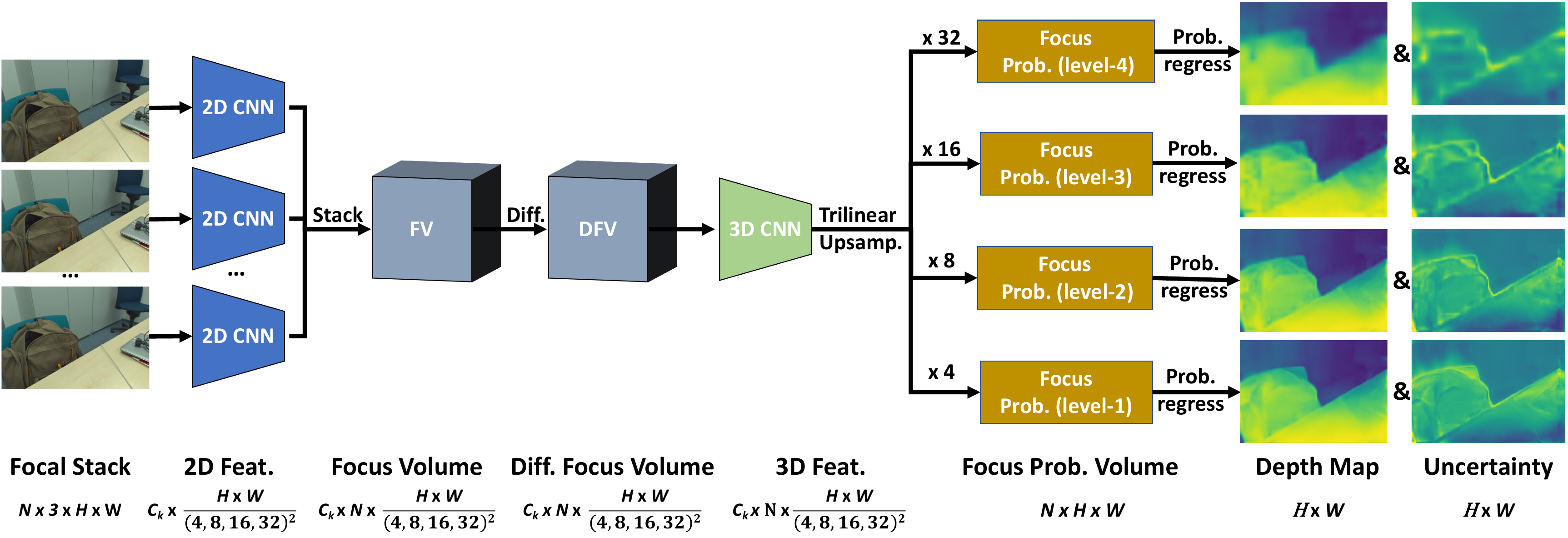}
\caption{Our multi-scale DFF network. Given an $N$-frame focal stack with $H\times W$ resolution, a shared 2D CNN first processes the frames to build the 4D FVs, where each layer corresponds to the features from one frame. The volume is then differentiated to obtain the DFVs, and sent to 3D CNN to predict the focus probabilities. The depth is estimated by the weighted sum in the probability regression, and the standard deviation is calculated to indicate the prediction uncertainty. The denominator in the feature dimension indicates the resolution scale of the corresponding level.} 
\vspace{-3mm}
\label{fig:pipeline}
\end{figure*}

\section{Related Works}
\label{sec:related_work}

%\smallskip
\noindent{\bf Depth-from-Focus (DFF)}. DFF is also known as shape-from-focus. Due to physical constraints, most cameras can only capture a clear image of objects in a certain range called depth of field (DoF). A point outside the range appears to be blurry and forms a circle of confusion (CoC). Researchers have been utilizing this phenomenon to infer depth for a long time~\cite{nair1992robust, subbarao1998selecting, nourbakhsh1996obstacle, pertuz2013overview}. According to the thin lens model~\cite{schechner2000depth}, given a densely sampled focal stack, there must be one and only one best-focused (sharpest) frame for a pixel. The focal distance of the frame can serve as the depth estimation of the pixel with the uncertainty of DoF. 

In practice, however, if the sample rate of the focal stack is low, such a frame may not always be visible. Therefore, traditional DFF methods usually take tens of frames in a focal stack as input and focus on finding good focus measures to identify the sharpest pixels. A number of focus measures have been proposed, including gradient-based measures~\cite{nair1992robust}, Laplacian-based measures~\cite{nayar1994SFF, ahmad20073dLapl}, frequency transformation-based measures~\cite{xie2006SFFwavelet}, and statistic-based measures~\cite{krotkov1988focusing}. We refer readers to~\cite{pertuz2013overview} for a comprehensive review. More recently,~\cite{surh2017noise} proposes a ring difference filter measure,~\cite{sakurikar2017composite} combines multiple measures into a composite, and~\cite{moeller2015variational} attempts to solve the noisy prediction with a variational framework. Deep DFF methods are proposed in~\cite{hazirbas2018DFFdataset, chen2021deep, wang2021bridging}. However,~\cite{hazirbas2018DFFdataset} and~\cite{chen2021deep} adopt their network from general dense prediction tasks, and the network of~\cite{wang2021bridging} is from a video representation work~\cite{alayrac2019visual}. None of them fully considers the specialty of DFF.

%But the networks  of~\cite{hazirbas2018DFFdataset} is directly adopted from the general dense prediction task~\cite{simonyan2014very}, and the network of~\cite{wang2021bridging} is from the video representation task~\cite{alayrac2019visual}. Neither of them fully considers the specialty of DFF. 

% In this work, we keep the specialty of DFF in mind, and propose the DFV. The 3D CNN used to aggregate the information is analogous to the 3D filters in traditional DFF algorithms~\cite{fan2018novel, ali2021robust}, and the focus probability volume output plays the same role as the traditional focus volume~\cite{mahmood2012nonlinear} which indicates the best-focused pixel distribution. 

\medskip
\noindent{\bf Focus Volume and Cost Volume}. 
In DFF, focus volume (FV) is commonly used to store the ``in-focus" score computed by focus measures. Once the FV is built, a naive approach can take the focal distance of the pixels with the highest scores as the depth estimation. But such an estimation is likely to be noisy due to the imperfect focus measurements. Thus, most works take the initial FV as features, and aggregate it with the average filter~\cite{nayar1994SFF, hosni2012fast}, bilateral filter~\cite{shoji2006shape}, and guided filter~\cite{jeon2019ring}  for better depth estimation.
%To alleviate the noise, early works~\cite{nayar1994SFF, nair1992robust} average the initial in-focus scores over a local window. More recent works employ iterative non-linear filter~\cite{mahmood2012nonlinear}, bilateral filter~\cite{shoji2006shape}, and guided filter~\cite{jeon2019ring} to refine the focus volume. %surh2017noise, sakurikar2017composite, 

% Cost volume is widely used in stereo matching~\cite{}, optical flow~\cite{}, and DFF~\cite{sakurikar2017composite}. While, in the first two tasks, the c 

It is worth noting that a similar concept called ``cost volume'' is popular in traditional matching tasks, \eg, stereo matching~\cite{scharstein1998stereo, yang2012CVstereo} and optical flow~\cite{xu2017accurate}, where the matching score rather than the in-focus score is stored and later aggregated.~\cite{surh2017noise} shows that the traditional cost aggregation method proposed for stereo matching~\cite{yang2012CVstereo} can be adapted to aggregate FV in DFF. With the advance in deep learning, many methods~\cite{kendall2017gcNet, chang2018PSM, yang2019hsm, cheng2019cspn, dosovitskiy2015flownet, ilg2017flownet, sun2018pwc, mao2021uasnet} build deep cost volume to incorporate context information and perform cost aggregation with 2D or 3D CNN to improve the estimation accuracy. In this work, we show that the core idea of deep cost volume and deep cost aggregation is also suitable for DFF. Moreover, we propose a deep differential focus volume by considering the special characteristics of DFF.
%and use the 3D CNN to aggregate the focus volume as stereo matching methods do.

\section{Methodology}
Figure~\ref{fig:pipeline} presents the overall pipeline of our method. As with~\cite{hazirbas2018DFFdataset, wang2021bridging}, we assume the input focal stack is pre-aligned by optical flow~\cite{suwajanakorn2015mobiledepth}, homography~\cite{jeon2019ring}, or any other methods. The network first extracts image features using a shared 2D CNN, and then builds differential focus volumes by computing the first-order derivative on the stacked features. This volume is further processed by a 3D CNN to predict the best-focused probabilities. Finally, multi-scale depth predictions are obtained by probability regression.  
 
In the following, we first briefly review the deep cost volume (Sec.~\ref{sec:cv}) and describe the proposed deep focus volume (Sec.~\ref{sec:fv}) and deep differential focus volume (Sec.~\ref{sec:dfv}) for the DFF task. Then, we introduce our probability regression and uncertainty estimation approach (Sec.~\ref{sec:dpth_reg}) and discuss the implementation details (Sec.~\ref{sec:implementation}).

% In the following, we first briefly review the connection between DFF and stereo matching, and the popular cost volume module in the deep stereo matching methods. Then, we introduce the proposed differential focus volume for the DFF task, followed with our probability regression and uncertainty estimation approach. The implementation details will be discussed at the end. 

% \subsection{Connection Between DFF and Stereo Matching}
% Consider a thin lens system as shown in Figure~\ref{fig:optical}. The diameter of the aperture is $D$, and the image plane is located at distance $v_1$ behind the lens. The system generates an in-focused image of point $P_0$ at distance $z_0$ in front of the lens, and a circle of confusion (CoC) of $P_1$. Based on geometry~\cite{}, we have 
% \begin{equation}
%     CoC = D\frac{|v_0 z_1 - v_0 F - z_1 F|}{Fz_1}
% \end{equation}
% where $F$ is the focal length. Now, if we block the entire lens except its top and bottom points, then for point $P_1$, the system can be analogized as stereo system with two pin-hole cameras whose image planes are perfectly overlapped (Fig~\ref{fig:optical}b).The detailed derivation can be found in~\cite{schechner2000depth}.

% % The focal length $f$ of these two camera is equal to $v_1$, the baseline $B$ equals to $D$, and the disparity of $P_1$ is $x_R - x_L$. We will found 
% % \begin{equation}
% %     x_R - x_L = CoC = D\frac{|v_0 z_ - v_0 F - z_0 F|}{fz_0} = B\frac{|f z_1 - f F - z_0 F|}{fz_1}
    
\subsection{A Brief Review of Deep Cost Volume}
\label{sec:cv}

Deep cost volume is widely used in stereo matching~\cite{mayer2016large, chang2018PSM, yang2019hsm, mao2021uasnet} and optical flow~\cite{dosovitskiy2015flownet, sun2018pwc, teed2020raft} nowadays. Although the exact implementation varies, there are two main types of volume designs: 3D cost volume and 4D cost volume.

In both designs, input frames are first processed by a 2D CNN to extract deep features (feature channel $\times$ height $\times$ width). The main difference is, for 3D cost volume~\cite{mayer2016large, dosovitskiy2015flownet}, a hand-crafted similarity measure (\eg, cross-correlation) is applied to the deep features to generate the cost volume, whereas for 4D cost volume~\cite{chang2018PSM, yang2019hsm, mao2021uasnet}, the deep features are stacked in a new ``disparity" dimension, resulting in a 4D representation (feature channel $\times$ disparity $\times$ height $\times$ width).  Every index in the disparity dimension represents a disparity proposal. Another difference is that the 3D cost volume is aggregated by 2D CNN to directly predict disparity values, but the 4D volume is followed by 3D CNN to produce a probability for each proposal. Although the 3D cost volume involves less computation, the 4D cost volume has been reported as a better implementation to integrate the context information and improve the model accuracy. 
% The left figure presents the in-focus score (feature) of a pixel in a 10-frame focal stack, and the right figure shows the differences between adjacent frames. 

\subsection{Deep Focus Volume}
\label{sec:fv}

As discussed in Sec.~\ref{sec:related_work}, there is a long history of using focus volume (FV) in DFF.
Compared to cost volume in stereo matching and optical flow, the only difference is that the FV is used to find the best-focused (sharpest) pixels rather than the best-matched pairs. % Previous work~\cite{surh2017noise} has shown that we can utilize the same aggregation methods from stereo matching to process the focus volume in DFF to improve the model performance. 
Given the two deep cost volume designs described in Sec.~\ref{sec:cv}, we can naturally develop two ways to build the deep FV in a DFF network. On the one hand, we can construct a 3D FV using hand-crafted focus measures and process it with 2D CNN to predict best-focused frames. On the other hand, we can construct a 4D FV by stacking the features in a new ``frame" dimension and use 3D CNN to produce a probability for each best-focused frame proposal.  As we wish to eliminate the use of hand-crafted measures, we adopt the latter approach. %Here, the 3D CNN can be viewed as learning a focus measure.

With this, our network would be the same as the one in Figure~\ref{fig:pipeline}, except that the DFVs are excluded. In Sec.~\ref{sec:exp}, we demonstrate that this simple approach already achieves state-of-the-art performance on multiple datasets. 

\subsection{Deep Differential Focus Volume}
\label{sec:dfv}
Despite the excellent performance of deep FV, we argue that such a design does not fully utilize the characteristics of DFF. In theory, in textured regions, DFF does not suffer from matching ambiguity if the comparing patch is larger than the widest CoC~\cite{nourbakhsh1996obstacle, schechner2000depth}. This means that if we have a good focus measure and operate it on a sufficiently large patch, there is always a single extremum on the pixel's in-focus curve, indicating the best-focused status. This is different from stereo matching or optical flow, where a reference patch may be matched with multiple targets due to repeated patterns. 

\begin{figure}[tp]
\centering
\includegraphics[width=1.0\linewidth]{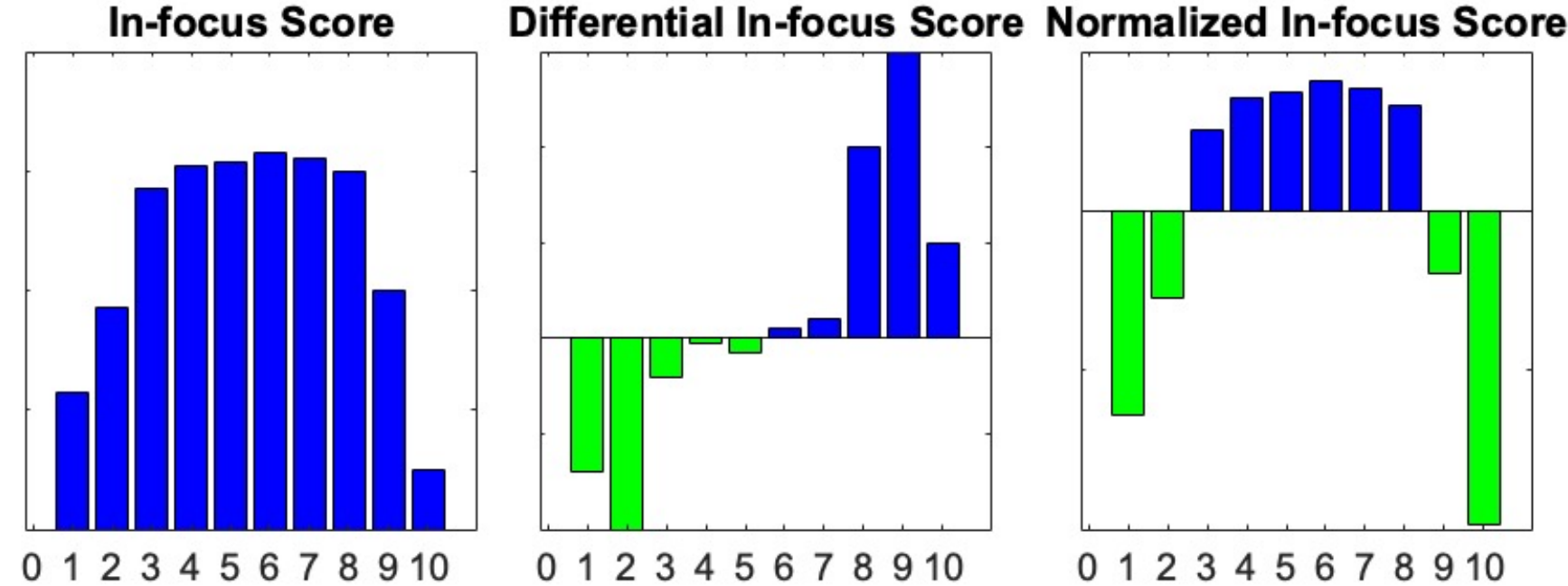}
\caption{Illustration of differential features and normalized features. The blue and green colors indicate positive and negative values, respectively.}
\vspace{-3mm}
\label{fig:toy}
\end{figure}

We utilize this single extremum and propose the deep differential focus volume. It is well-known that the gradient is a good indicator for the single extremum. This is illustrated in Figure~\ref{fig:toy}. For a single pixel in a 10-frame focal stack, we pick the positive in-focus score as the feature. Sometimes, due to weak texture, the feature magnitude of the sharpest pixel may not be salient - see the scores of frames 4 to 8 in the left figure. Consequently, it could be hard for the network to locate the sharpest pixel. However, if we take the difference in the feature, the sharpest pixel will correspond to the zero-crossing, as the middle figure shows. This is a more significant pattern to distinguish. We also compare the effect between differential and normalization (the right figure). While normalization can enlarge the relative score gaps, it does not make the sharpest pixel as prominent as the differential one does.

Therefore, we differentiate the focus volume $Q$ along the frame dimension to build the differential focus volume: 
 \begin{equation}
    V^i = 
    \left\{
    \begin{array}{ll}
        Q^{i} - Q^{i+1},  &  i = 1, \ldots, N-1   \\
        Q^i,   & i=N
    \end{array}
 \right.
 \end{equation}
where $i$ is the index of the frame dimension, and $N$ is the total number of frames. Note that the first $N-1$ dimensions of $V$ contain the first-order feature derivatives of adjacent frames, whereas the last dimension is equal to the features of frame $N$, containing the original context information. Thus, the 3D CNN has access to both differential features and context features for focus analysis. This is important, especially for textureless regions. Arguably, because of defocus, some parts of the last frame may be blurry. But it can still serve as ``context.'' As the purpose of the context is to capture global relationships among image regions, the lack of certain fine details will not impair its effectiveness. Nevertheless, better ways to gather the context information can be studied in future work. For example, we may perform an average pooling or convolution over $Q$ along the frame dimension to learn context from the entire stack.
% Arguably, in the extreme defocus case, the $N$ frame may originally fail to capture the whole context  do not capture the  well present the context due to the defocus blur,  perfect \todo: it is not functionally the same to normalization. 

According to the universal approximation theorem~\cite{hornik1989multilayer}, the network might be able to eventually learn a similar representation by itself. But we believe explicitly introducing the known prior to the network will help the learning process, just as the deep cost volume for the matching tasks. 

\subsection{Depth Regression and Uncertainty Estimation}
\label{sec:dpth_reg}
Besides textureless regions, the required number of frames is another issue in classic DFF. Because classic DFF methods infer the depth by locating the sharpest pixels in a focal stack, they usually need tens of frames as input. Otherwise, the sharpest pixels may be invisible, and the depth inference will be inaccurate. In this work, we locate the sharpest pixel in sub-frame accuracy by learning a probability distribution. 

The final output of our network is a focus probability volume $P$, where $p_j^i$ indicates the probability that the pixel $x_j$ in the $i^{th}$ frame is the  best-focused. The sum of the probability of pixel $x_j$ is constrained to $1$ by softmax activation. Thus, the best-focused frame ID of pixel $x_j$ is $\hat{i}_j = \sum_{i=1}^{N} p_i^j \cdot i$. Note, $\hat{i}_j$ can be a float value that indicates the best-focused frame locates between $\lfloor \hat{i}_j \rfloor$ and $\lceil \hat{i}_j \rceil$. Similarly, we can obtain the depth $\hat{d}_j$ as
\begin{equation}
    \hat{d}_j =  \sum_{i=1}^{N} p^i_j \cdot l^i,
\end{equation}
where $l^i$ is the focal distance of the $i^{th}$ frame. If the camera is calibrated, we take the actual value of $l^i$. Otherwise, we assume the frames are sorted with ascending focal distances and set the distance from 0 to 1. In practice, the sorting can be achieved by first estimating the up-to-scale focal distances with the method proposed in~\cite{suwajanakorn2015mobiledepth}. 

The prediction confidence of the network can be revealed as the uncertainty $\Phi$. We compute the uncertainty of pixel $x_j$ using the weighted standard deviation:
\begin{equation}
    \phi_j = \sqrt{\sum_{i=1}^{N} p^i_j \cdot (l^i - \hat{d}_j)^2}.
\end{equation}
Besides serving as a confidence indicator, this uncertainty measure may also be used in scenarios such as multi-task learning~\cite{kendall2018multi} and measurement fusion~\cite{yeo2021robustness}.

\begin{figure*}[ht]
\centering
\begin{tabular}{ccccc|c}
Image/GT &  \hspace{-2.5mm}DDFF & \hspace{-2.5mm} DefocusNet &\hspace{-2.5mm}  Ours-FV &\hspace{-2.5mm}  Ours-DFV &\hspace{-2.0mm}  {\small Uncer. Ours-FV/DFV} \\

\includegraphics[height =0.98in]{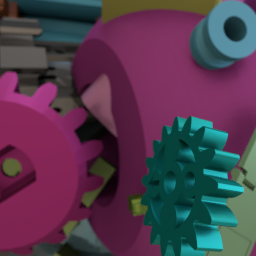} &
% \hspace{-2.5mm}\includegraphics[height =0.8in]{figures//unknown_res/telephone_img.png} &
\hspace{-2.5mm}\includegraphics[height =0.98in]{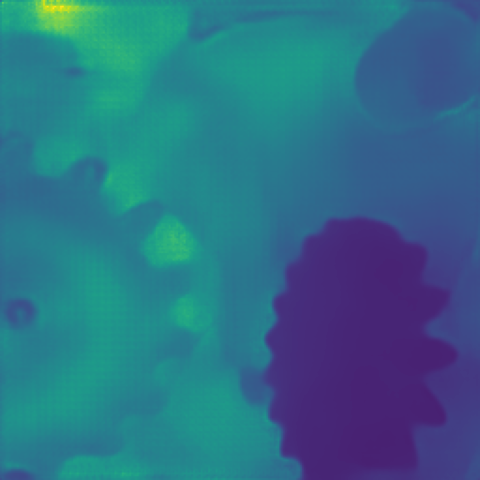} &
 \hspace{-2.5mm}\includegraphics[height =0.98in]{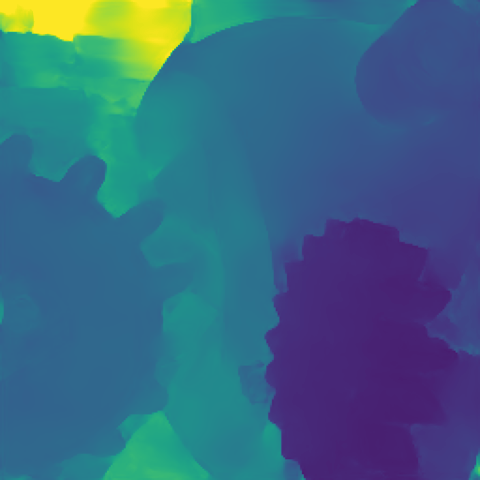}&
\hspace{-2.5mm}\includegraphics[height =0.98in]{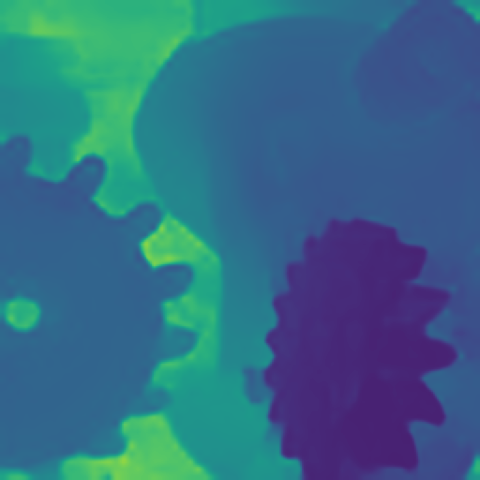} &
\hspace{-2.5mm}\includegraphics[height =0.98in]{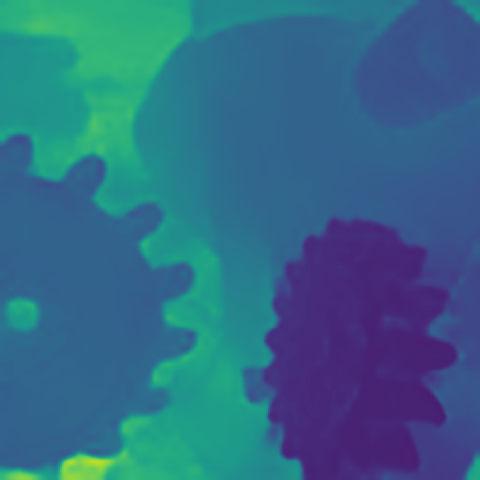} & 
\hspace{-1.2mm}\includegraphics[height =0.98in]{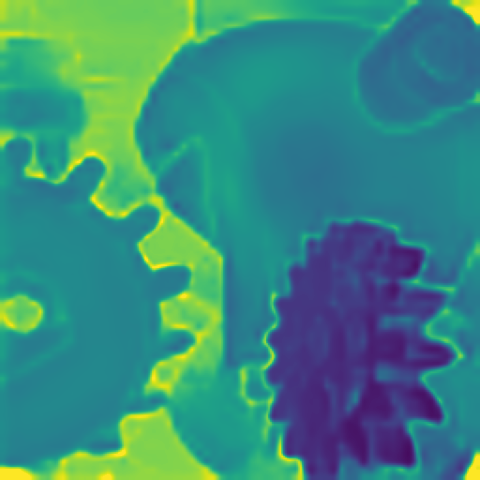} \\
\includegraphics[height= 0.98in]{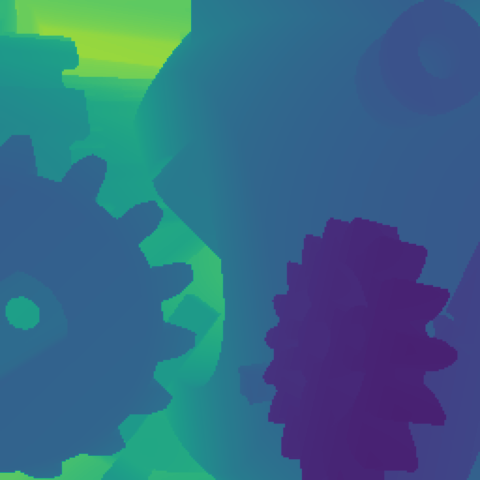} &
% \hspace{-2.5mm}\includegraphics[height= 0.8in]{figures/unknown_res/telephone_pred_viz.png} &
\hspace{-2.5mm}\includegraphics[height= 0.98in]{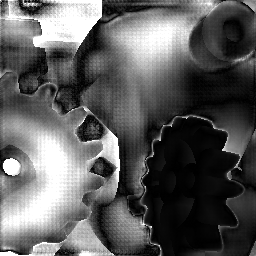} &
\hspace{-2.5mm}\includegraphics[height= 0.98in]{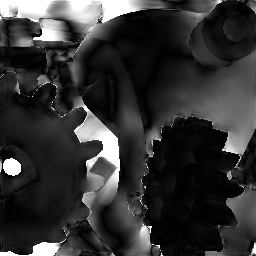} &
\hspace{-2.5mm}\includegraphics[height =0.98in]{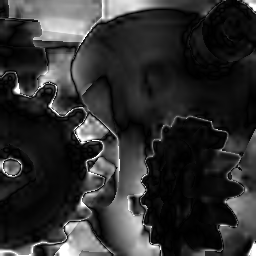} &
\hspace{-2.5mm}\includegraphics[height= 0.98in]{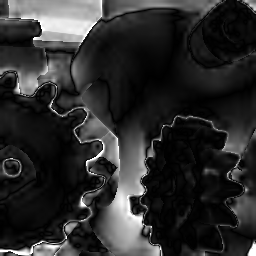} & 
\hspace{-1.2mm}\includegraphics[height= 0.98in]{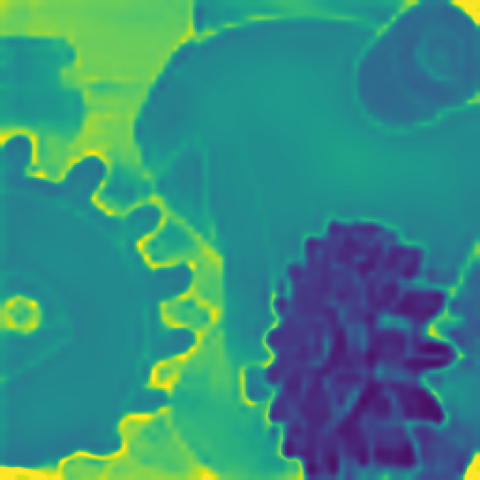}\\

\includegraphics[height =0.68in]{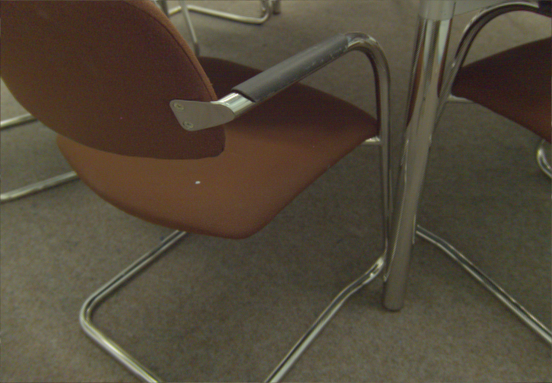} &
% \hspace{-2.5mm}\includegraphics[height =0.8in]{figures//unknown_res/telephone_img.png} &
\hspace{-2.5mm}\includegraphics[height =0.68in]{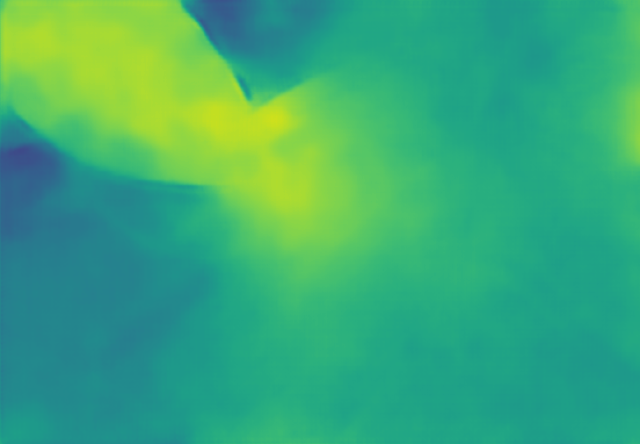} &
 \hspace{-2.5mm}\includegraphics[height =0.68in]{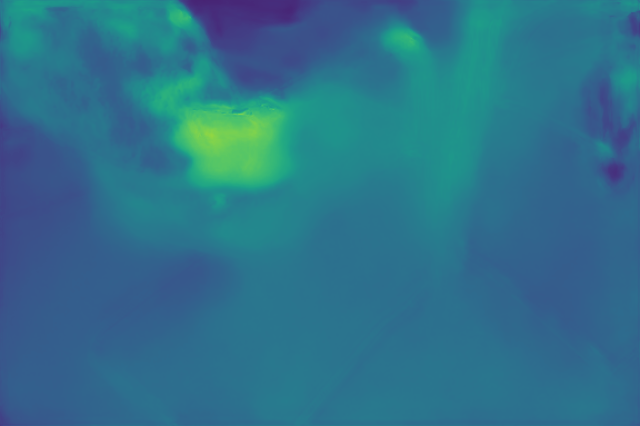}&
\hspace{-2.5mm}\includegraphics[height =0.68in]{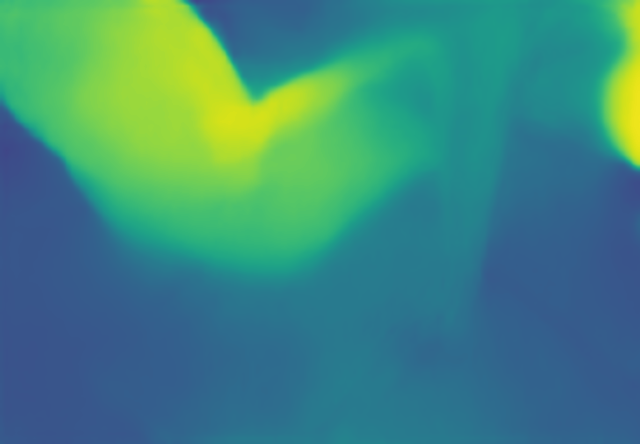} &
\hspace{-2.5mm}\includegraphics[height =0.68in]{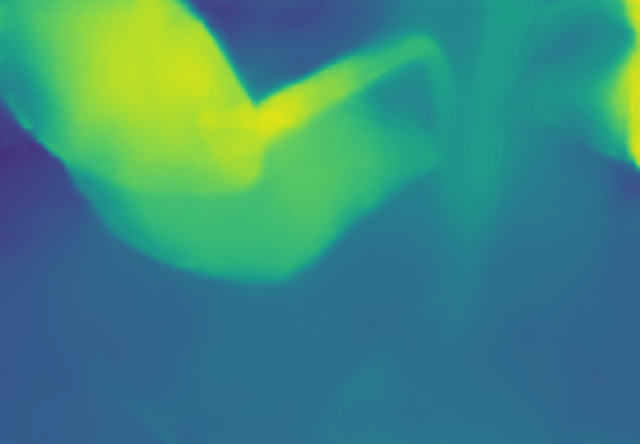} & 
\hspace{-1.2mm}\includegraphics[height =0.68in]{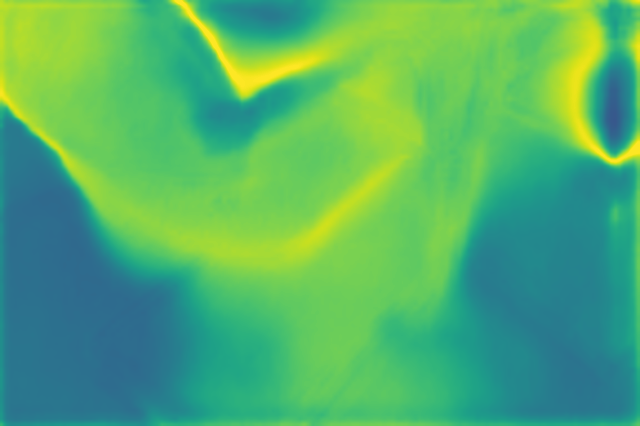}\\

\includegraphics[height= 0.68in]{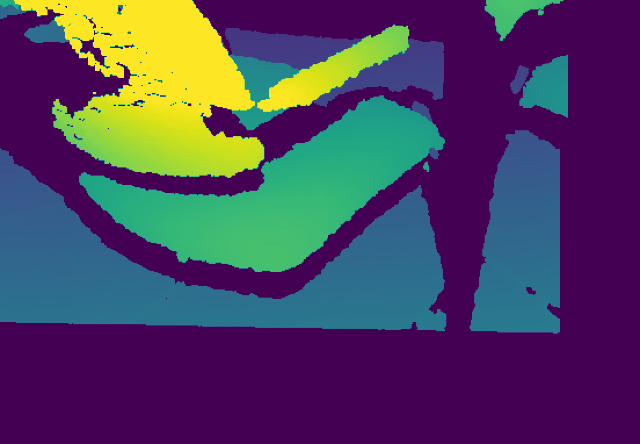} &
% \hspace{-2.5mm}\includegraphics[height= 0.8in]{figures/unknown_res/telephone_pred_viz.png} &
\hspace{-2.5mm}\includegraphics[height= 0.68in]{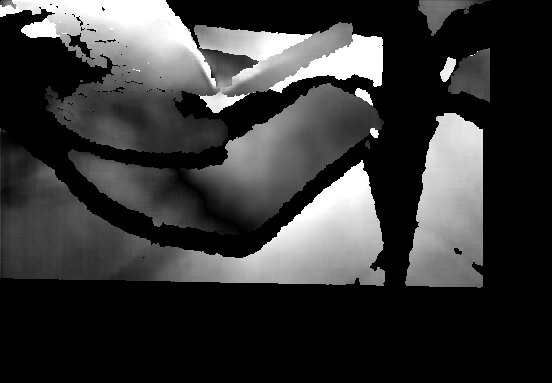} &
\hspace{-2.5mm}\includegraphics[height= 0.68in]{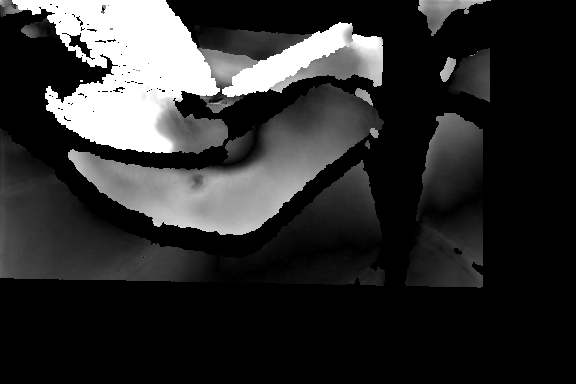} &
\hspace{-2.5mm}\includegraphics[height =0.68in]{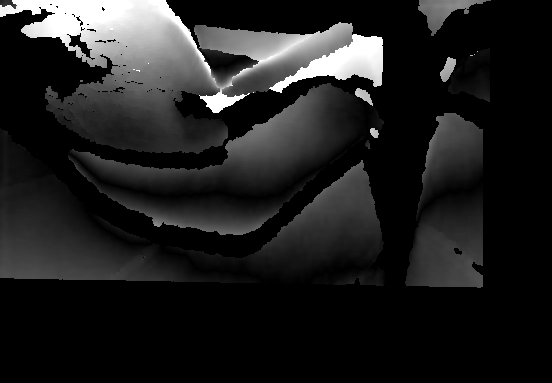} &
\hspace{-2.5mm}\includegraphics[height= 0.68in]{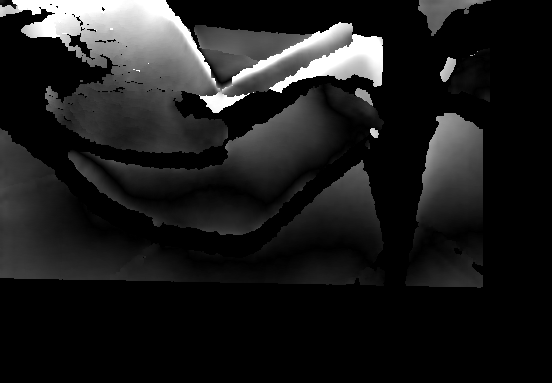} & 
\hspace{-1.2mm}\includegraphics[height =0.68in]{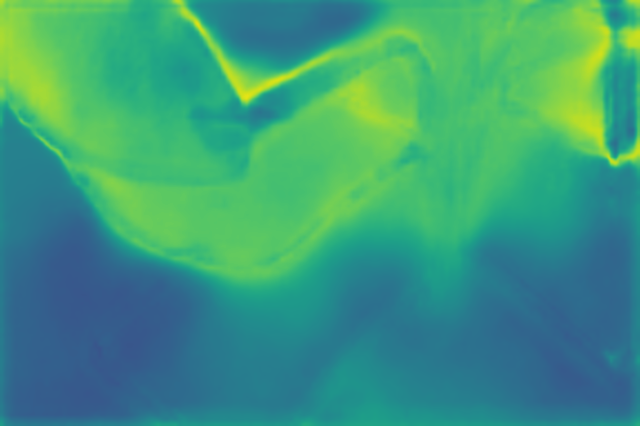}
 \end{tabular}
\caption{Qualitative results on FoD500 (first two rows) and DDFF-12 (last two rows). The first column shows the first image in the input focal stack and the corresponding ground truth. The next 4 columns show depth (Row 1) and disparity (Row 3) predictions, and the corresponding error maps (Rows 2 and 4). The last column shows the uncertainty maps of Ours-FV (Rows 1 and 3) and Ours-DFV (Rows 2 and 4). The warmer or brighter the color, the higher the value.} % For error map, the darker the color, the lower the value.}
\vspace{-3mm}
\label{fig:fod_ddff_res_viz}
\end{figure*}

\subsection{Implementation Details}
\label{sec:implementation}
Our network design is adopted from~\cite{yang2019hsm} and optimized for the DFF task. For the 2D CNN, we employ a ResNet-18-FPN~\cite{lin2017fpn} which is pre-trained on ImageNet. A spatial pyramid pooling (SPP)~\cite{zhao2017SPP} module is inserted between the encoder and decoder to better capture local and global information. We take the last 4-level features and build four DFVs at corresponding resolutions. For each DFV, we use two 3D-ResNet blocks~\cite{hara2017Res3D} followed by a 3D-SPP for aggregation. A 2-layer 3D convolution is used to predict the focus probability. The aggregated DFV is also sent to the next scale level after upsampling and a 3D convolution. Please refer to our supplementary material for more details. 

We implement our model using Pytorch and optimize it with Adam ($\beta_1=0.9$ and $\beta_2=0.999$) on one NVIDIA 1080Ti GPU for 700 epochs. The batch size is 20, and the learning rate is $1\times10^{-4}$. Given a training focal stack, we first randomly select 5 frames and crop them into 224$\times$224 resolution with random-flipping afterward. All the frames are organized with the ascending focal distance order. %We have also tried to randomly shuffle the frames, but the results are not as good. We suspect that this is because all the test focal stacks are in ascending focal distance order. 

At training time, we take all 4-level outputs and compare the predicted pixel depth $\hat{d}_j^s$ with the ground truth depth $d_j^s$, using a multi-scale smooth L1 loss, $L$. At test time, we only output the largest scale (level-1 in Figure~\ref{fig:pipeline}) focus probability volume for depth regression. 
 \begin{equation}
L = \sum_{s=1} ^{4} \alpha_s \Big(  \frac{1}{M} \sum_{j=1}^{M} smooth_{L_1}(d^s_j - \hat{d}^s_j) \Big),
\label{eqn:final}
\end{equation}
where %$d_j^s$ and $\hat{d}_j^s$ are the ground truth and predicted depths at scale $s$, respectively. 
$M$ is the total number of pixels, and we set $\alpha_s=\{\frac{8}{15}, \frac{4}{15},\frac{2}{15},\frac{1}{15}\}$ for all experiments.\footnote{Code is available at \url{https://github.com/fuy34/DFV}}

\section{Experiments}
\label{sec:exp}
We conduct comprehensive experiments to study the model performance. First,  we use two annotated datasets, FoD500~\cite{maximov2020focusondefocus} and DDFF-12~\cite{hazirbas2018DFFdataset}, for quantitative and qualitative comparison. Next,  we conduct ablation studies on the model's sensitivity to the focal stack size and the multi-scale architecture. Finally,  we test the model generalizability on an unlabeled dataset, Mobile depth~\cite{suwajanakorn2015mobiledepth}. %the model's generalizability.%  Since no ground truth depth is provided in this dataset, only qualitative results are provided for comparison.

\begin{table*}[ht]
\centering
 \setlength{\tabcolsep}{4pt}
    {\small
    \begin{tabular}{l|cccccccccccc}
        \hline
        Method & MSE $\downarrow$  & RMS$\downarrow$ & log RMS $\downarrow$ & Abs. rel.$\downarrow$& Sqr. rel.$\downarrow$ & $\delta\uparrow$ & $\delta^2\uparrow$ & $\delta^3\uparrow$ & Bump.$\downarrow$ & avgUnc.$\downarrow$ & Time(ms)$\downarrow$ \\
        \hline 
        VDFF~\cite{moeller2015variational} & 29.66$e^{-2}$ &5.05$e^{-1}$ & 0.87 & 1.18 & 85.62$e^{-2}$ & 17.92 & 32.66 & 50.31 & {1.12} & -- & -- \\
        RDF~\cite{jeon2019ring} & 11.15$e^{-2}$ &3.22$e^{-1}$ & 0.71 & 0.46 & 23.95$e^{-2}$ & 39.48 & 64.65 & 76.13 & {1.54}& -- & -- \\
        DDFF~\cite{hazirbas2018DFFdataset}&  3.34$e^{-2}$ &1.67$e^{-1}$ & 0.27 & 0.17 & 3.56$e^{-2}$ & 72.82 & 89.96 & 96.26 & {1.74}& -- & 50.6 \\
        DefocusNet~\cite{maximov2020focusondefocus} & 2.18$e^{-2}$ &1.34$e^{-1}$ & 0.24 & 0.15 & 3.59$e^{-2}$ & 81.14 & 93.31 & 96.62 & {2.52}& --& 24.7 \\
        {Ours-FV} & {\bf 1.88$e^{-2}$}  & {\bf 1.25$e^{-1}$} & {\bf 0.21} & {0.14} & 2.43$e^{-2}$ & {81.16} & {\bf 94.97}& {\bf 98.08} & {1.45} &0.24 & {\bf 18.1} \\ %\textbf{1.26$e^{-2}$}\\  \textbf
        {Ours-DFV} & 2.05$e^{-2}$ & { 1.29$e^{-1}$} & { 0.21} & {\bf 0.13} & {\bf 2.39$e^{-2}$} & {\bf 81.90} & {94.68} & 98.05 & {\bf 1.43} & {\bf 0.17} & 18.2 \\
        \hline
    \end{tabular}}
\caption{Evaluation results on FoD500 test set.}
% \vspace{-3mm}
\label{tab:fod_res}
\end{table*}

\subsection{Datasets} 
% \smallskip
\noindent{\bf FoD500}~\cite{maximov2020focusondefocus} is a synthetic DFD dataset\footnote{\url{https://github.com/dvl-tum/defocus-net}}, containing 400 training samples and 100 test samples. Every sample has a 5-frame focal stack with known focal distances and a ground truth depth map. The image resolution is 256$\times$256. As the dataset is originally designed for DFD, the focal distance range of a sample does not always cover its ground truth depth range. We mask pixels whose depth values are out of the focal distance range at both training and test time.  % accessed on 03/10/2021

\smallskip
\noindent{\bf DDFF-12}~\cite{hazirbas2018DFFdataset} is a real-world DFF dataset captured by a light-field camera from 12 different scenes. Six scenes, {\it glassroom, kitchen, office41, seminaroom, socialcorner, studentlab}, are selected as the training set, each containing 100 samples. The other six scenes, {\it cafeteria, library, lockeroom, magistrale, office44, spencerlab}, are chosen as the test set with 20 samples per scene. Each sample contains a 10-frame focal stack with known focal disparities and a ground truth disparity map. The image resolution is 383$\times$552. We further divide the original training set into 4 training scenes ({\it kitchen, seminaroom, socialcorner, studentlab}) and 2 validation scenes ({\it glassroom,  office41}) by random selection. Following~\cite{hazirbas2018DFFdataset}, we evaluate the disparity accuracy on this dataset. 

\smallskip
\noindent{\bf Mobile depth}~\cite{suwajanakorn2015mobiledepth} is a real-world DFF dataset captured by a mobile phone. The dataset consists of 11 aligned focal stacks and 2 unaligned stacks from 11 scenes. The image resolution varies between $360\times640$ and $518\times774$, and the number of frames ranges from 14 to 33 per stack. Neither ground truth depth nor the focal distance is provided. As~\cite{suwajanakorn2015mobiledepth} only releases their results on the 11 aligned focal stacks and no code is available, we only evaluate our method on the aligned scenes for qualitative comparison. 

\begin{table*}[ht]
\centering
 \setlength{\tabcolsep}{4pt}
{\small
\begin{tabular}{l|cccccccccccc}
\hline
Method & MSE $\downarrow$  & RMS$\downarrow$ & log RMS $\downarrow$ & Abs. rel.$\downarrow$& Sqr. rel.$\downarrow$ & $\delta\uparrow$ & $\delta^2\uparrow$ & $\delta^3\uparrow$ & Bump.$\downarrow$ & avgUnc.$\downarrow$ & Time(ms)$\downarrow$\\
\hline 
VDFF~\cite{moeller2015variational} & 156.55$e^{-4}$ & 12.14$^{-2}$ & 0.98 & 1.38 & 241.2$e^{-3}$ & 15.26 & 29.46 & 44.89 & 0.43 & -- & --\\
RDF~\cite{jeon2019ring} &  91.81$e^{-4}$ & 9.41$e^{-2}$ & 0.91 & 1.00 & 139.4$e^{-3}$ & 15.65 & 33.08 & 47.48 & 1.33 & -- & --\\
DDFF~\cite{hazirbas2018DFFdataset} & 8.97$e^{-4}$ & 2.76$e^{-2}$ & 0.28 & 0.24 & 9.47$e^{-3}$ & 61.26 & 88.70 & 96.49 & 0.52 & -- & 191.7 \\
DefocusNet~\cite{maximov2020focusondefocus} & 8.61$e^{-4}$ & 2.55$e^{-2}$ & 0.23 & 0.17 & {\bf 6.00}$e^{-3}$ & 72.56 & 94.15 & 97.92 & 0.46 & -- & 34.3\\

Ours-FV   & 6.49$e^{-4}$ & 2.28$e^{-2}$ & 0.23 & 0.18 & 7.10$e^{-3}$ & 71.93 & 92.80 & 97.86 & 0.42 & 5.20$e^{-2}$ & {\bf 33.2}\\
Ours-DFV & {\bf 5.70$e^{-4}$} & {\bf 2.13$e^{-2}$} & {\bf 0.21} & {\bf 0.17} & {6.26$e^{-3}$} & {\bf 76.74} & {\bf 94.23} & {\bf 98.14} & {\bf 0.42} & {\bf 4.99$e^{-2}$} & 33.3 \\
% Ours(dcv_{DGF})  & {6.2$e^{-4}$} & {2.22$e^{-2}$} & {0.21} & {0.17} & {6.67$e^{-3}$} & {75.96} & {93.80} & {97.87} & {0.43} & {5.12$e^{-2}$}\\
\hline
\end{tabular}}
\caption{Evaluation results on DDFF-12 {validation} set.}
\vspace{-3mm}
\label{tab:ddff12_res}
\end{table*}

\begin{table*}[htp]
\centering
% \begin{threeparttable}
 \setlength{\tabcolsep}{4pt}
  {\small
    \begin{tabular}{l|ccccccccccccc}
        \hline
        Method & MSE $\downarrow$  & RMS$\downarrow$ & log RMS $\downarrow$ & Abs. Rel.$\downarrow$& Sqr. rel.$\downarrow$ & $\delta\uparrow$ & $\delta^2\uparrow$ & $\delta^3\uparrow$ & Bump.$\downarrow$\\
        \hline 
        DDFF~\cite{hazirbas2018DFFdataset} & 9.68$e^{-4}$ & 9.01$e^{-2}$ & 0.32 & 0.29 & {0.01} & 61.95 & 85.14 & 92.98 & 0.59\\
        AiFDepthNet~\cite{wang2021bridging} & 8.6$e^{-4}$ & -- & 0.29 & 0.25 & 0.01 & 68.33 & 87.40 & 93.96 & 0.63 \\
        % DefocusNet~\cite{maximov2020focusondefocus} & 9.1$e^{-4}$ & -- & -- & -- & -- & -- & -- & -- & --\\
        Ours-FV   & 6.54$e^{-4}$ & 7.55$e^{-2}$ & 0.25 & 0.20 & 0.01 & 68.58 & 91.26 & 97.36 & 0.58 \\
        Ours-DFV & \textbf{5.58$e^{-4}$} & \textbf{6.87$e^{-2}$} & \textbf{0.23} & \textbf{0.19} & \textbf{0.01} & \textbf{74.26} & \textbf{92.38}& \textbf{97.39} & \textbf{0.57}\\ %\textbf{1.26$e^{-2}$}\\
        \hline
    \end{tabular}}
\caption{Evaluation results on DDFF-12 {test} set. All values are from DDFF-12 leaderboard except that AiFDepthNet's results are from~\cite{wang2021bridging}.}
\vspace{-1mm}
\label{tab:ddff12_test}
\end{table*}

\subsection{Comparison to the State-of-the-Art}
\label{sec:exp_res}
To evaluate the model performance, we compare our differential focus volume network, denoted as ``{Ours-DFV},'' with five existing methods, VDFF~\cite{moeller2015variational}, RDF~\cite{jeon2019ring}, DDFF~\cite{hazirbas2018DFFdataset}, DefocusNet~\cite{maximov2020focusondefocus}, and AiFDepthNet~\cite{wang2021bridging}. We also compare to a variant of our method called ``{Ours-FV},'' which uses the deep focus volume as stated in Sec.~\ref{sec:fv}. 

VDFF and RDF are classic DFF methods. VDFF uses a Laplacian focus measure and a variational framework to generate smooth depth estimation. RDF uses the ring difference filter as the focus measure and aggregates the focus volume with guided filter. We use the code provided by the authors and run it on the test sets with default parameters.

DDFF and DefocusNet are two deep learning methods for DFF and DFD, respectively. We train the models with the original code from scratch in two different ways: (1) train on the mixed dataset as our methods do, and (2) first train on FoD500 dataset and then fine-tune it on DDFF-12 dataset as DefocusNet did. For both ways, we train it until the validation loss converges and pick the one with better performance on the DDFF-12 validation set for testing. As our model learns to predict the best-focused probability  instead of the depth or disparity, in the mixed dataset training, there is no need to unify the ground truth from disparity to depth, or vice versa. But we find that, for DDFF and DefocusNet, such a unification improves their performance. Therefore, we convert the ground truth of the two datasets to the methods' original settings for training and convert their predictions back for evaluation. That is, at training time, we convert the ground truth disparity in DDFF-12 to depth for DefocusNets and convert the ground truth depth in FoD500 to disparity for DDFF. As a result, DDFF favors the first training approach, and DefocusNet prefers the second. 

AiFDepthNet is a recent deep learning method trained on a large amount of data, including DDFF-12, FoD500, FlyingThings3D~\cite{mayer2016large}, and 4D Light Field~\cite{honauer2016dataset}. The network supervision can be either the ground truth depth or the corresponding all-in-focus image. Because it does not release the training code, we cannot reproduce their results under our experiment settings. Thus, we only include AiFDepthNet in the comparison on the DDFF-12 test set, where its results are directly copied from the AiFDepthNet paper~\cite{wang2021bridging}.

For a fair comparison,  all methods take five frames from the input focal stack by random sampling at training time and by equidistant sampling at test time. The frame with the smallest or the largest focal distance/disparity will always be sampled at test time. We employ the same metrics used in~\cite{hazirbas2018DFFdataset} and introduce a new metric called average uncertainty (avgUnc.): $\frac{1}{M}\sum^M_{j=1}\phi_j$ to compare the prediction confidence of Ours-FV and Ours-DFV.

% We employ the same metrics as~\cite{hazirbas2018DFFdataset} for evaluation, including MSE, RMS, log RMS, absolute relative (Abs. rel.), squared relative (Sqr. rel.), three accuracy percentages ($\delta$, $\delta^2$, $\delta^3$), and bumpiness (Bump.). The first 8 metrics reflect the estimation accuracy from absolute and relative perspectives, and the last one evaluates the smoothness of results. Besides, to compare the prediction confidence between {Ours-FV} and {Ours-DFV}, we introduce a new metric called average uncertainty (avg. Uncer.): $\frac{1}{M}\sum^M_{j=1}\phi_j$. The lower the value, the higher the confidence. 

\smallskip
\noindent{\bf Result on FoD500}. Table~\ref{tab:fod_res} presents the quantitative results on FoD500. Because of the limited sample frequency (5 frame/stack), both classic methods fail to deliver competitive results. In contrast, all deep methods work well. Compared to DDFF and DefocusNet, our methods provide more accurate and smooth results, which can also be seen from the qualitative results in the first two rows in Figure~\ref{fig:fod_ddff_res_viz}. Our methods better preserve the object boundaries, such as the cog and the central hole of the left gear, and show more smooth depth inference on the object surfaces. This observation verifies the power of the deep 4D FV. The effectiveness of the DFV is not significant in terms of the accuracy (first eight) and smoothness (Bump.) metrics, due to the dataset simplicity. But the avgUnc. of Ours-DFV is lower than Ours-FV, showing the higher confidence of Ours-DFV.

\smallskip
\noindent{\bf Result on DDFF-12}. DDFF-12 is a more challenging dataset because of the large textureless regions in real-world scenarios. To reach high accuracy on this dataset, methods have to capture the weak defocus signal and utilize the context information to make a reasonable estimation.  As DDFF-12 leaderboard only allows participants to submit their own methods, we first perform the comparison on the validation set. Table~\ref{tab:ddff12_res} presents the quantitative results. As with the FoD500 results, deep learning methods outperform the classic approaches with large margins. %due to the ability to capture more context information with convolution operations. 
Both of our methods consistently outperform the other deep learning methods. Thanks to the DFV module, Ours-DFV is able to better identify the weak defocus signal, and is more accurate and more confident than {Ours-FV}. For example, the MSE of Ours-DFV is 36.4\% lower than DDFF, 33.8\% lower than DefocusNet, and 12.2\% lower than Ours-FV.  The qualitative results are presented in the last two rows in Figure~\ref{fig:fod_ddff_res_viz}. More examples are available in our supplementary material. In the material, we also illustrate the focus probability distribution of Ours-DFV, which empirically shows the network does learn to locate the best-focused pixels.

Finally, we train Ours-FV and {Ours-DFV} models on the training and validation sets with the same scheme and submit the test result to the leaderboard\footnote{\url{https://competitions.codalab.org/competitions/17807\#results}}. Table~\ref{tab:ddff12_test} shows the results, where all values are from the leaderboard, except the one of AiFDepthNet is from their paper~\cite{wang2021bridging}.
%A consistent trend with the validation result can be observed.
At the time of submission, Ours-DFV ranks $1^{st}$ on the leaderboard. % and our disparity MSE error is 34.8\% lower than the second best one. 

\smallskip
\noindent{\bf Runtime}. We compare the running speed among the deep learning methods.  All models are tested with 5-frame focal stack inputs on an NVIDIA 1080Ti GPU. The results are shown in Table~\ref{tab:fod_res} and Table~\ref{tab:ddff12_res} with unit ms. Our methods are slightly faster than the DefocusNet and about three to six times faster than DDFF. In addition, {Ours-FV} and {Ours-DFV} have almost the same running time ($\pm$0.1ms), which shows the efficiency of the DFV module. 

\subsection{Ablation Study}
\begin{figure}[htp]
\centering
\includegraphics[width=0.75\linewidth]{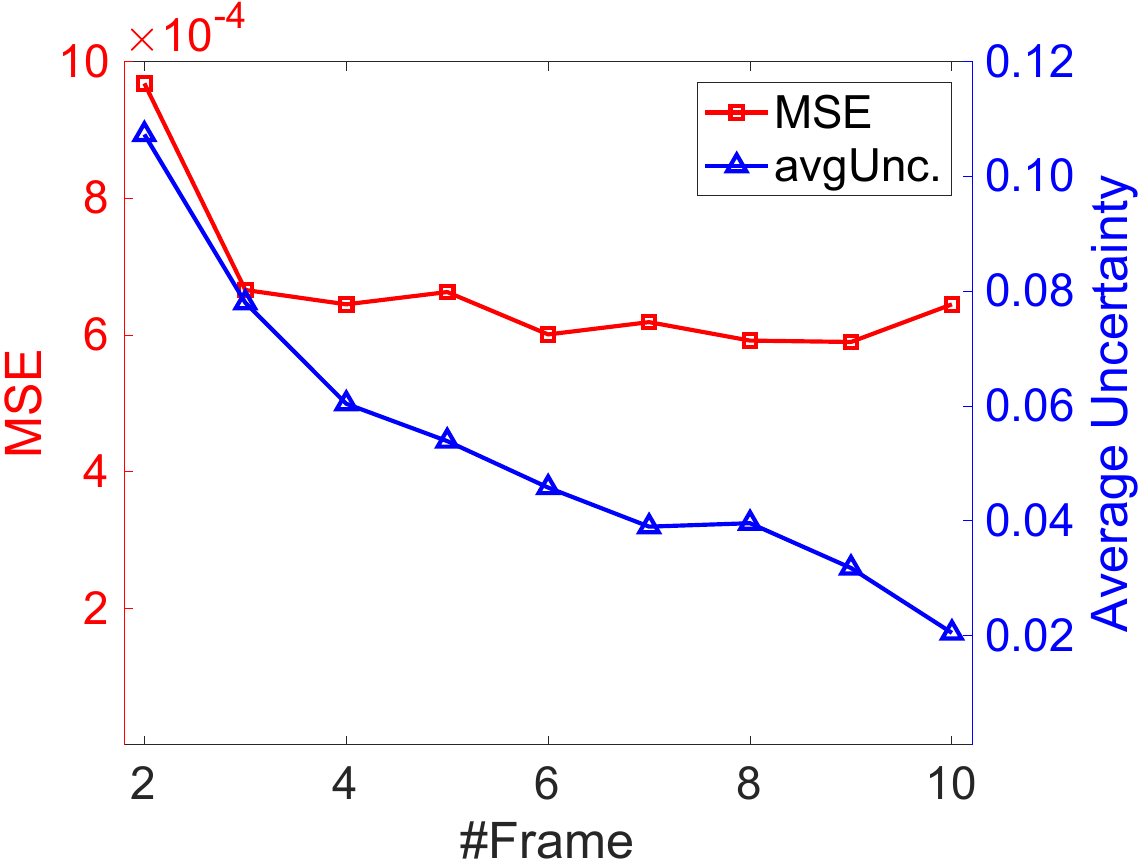}
\caption{Model performance w.r.t. focal stack sizes.}
\label{fig:frameNum}
\vspace{-3mm}
\end{figure}

\smallskip
\noindent {\bf Focal Stack Size}. In DFF, the input focal stack size is a key variable to the method performance. We evaluate our model's sensitivity to the stack size by training {Ours-DFV} with different sizes, $N = 2, \ldots, 10$, on DDFF-12 training set with the same training scheme and testing on the validation set. We exclude FoD500 in the experiment, because its focal stacks only contain 5 frames. Figure~\ref{fig:frameNum} shows the MSE and avgUnc. changes w.r.t. the input frame number. The full metrics table is available in the supplementary material. We can see the model starts to deliver fairly accurate results in terms of MSE with only 3 frames/stack. The avgUnc. keeps decreasing with increasing stack size, showing the model's growing confidence with more input frames. However, we notice that the impact of additional frames is diminishing as the frame number increases. The authors of~\cite{hazirbas2018DFFdataset} also report that their network performance on DDFF-12 dataset stops improving after the frame number reaches $10$ (which is the stack size they finally released to the public). Further studies with a new dataset containing more frames per stack are needed to find the actual reason.

\smallskip
\noindent {\bf Multi-scale Architecture}. Our network uses a multi-scale architecture, and deep supervision (Eq.~\ref{eqn:final}) is applied at training time. To investigate the impact of this design, we train three variants of Ours-DFV model named DFV-L$k$, where $k=1, 2, 3$. In each variant, we only use the $k$ largest scale DFVs and remove the rest DFVs and the associated 3D CNN modules from the network. For example, for DFV-L2, only the 2 largest-scale DFVs ($\frac{1}{4}$ and $\frac{1}{8}$) and their associated 3D CNN modules are used. The training scheme is the same as Ours-DFV. The evaluation results on the DDFF-12 validation set are shown in Table~\ref{tab:multiscale}. The depth accuracy increases as more DFV modules are used, verifying the effectiveness of the multi-scale design. Compared with Table~\ref{tab:ddff12_res}, even with a single DFV module (DFV-L1), the model still outperforms existing DFF and DFD methods by large margins. This demonstrates the effectiveness of DFV.

\begin{table}[bt]
\centering
{\small
\begin{tabular}{l|cccc}
\hline
Method & MSE $\downarrow$  & RMS$\downarrow$  & Abs.rel.$\downarrow$ &  Time(ms) $\downarrow$  \\
\hline 
DFV-L1 & 6.65$e^{-4}$ &2.30$e^{-2}$ &  0.18 &  {\bf 24.1} \\
DFV-L2 & 6.16$e^{-4}$ & 2.20$e^{-2}$ & 0.17 &  28.0 \\
DFV-L3 & 5.94$e^{-4}$ & 2.16$e^{-2}$ & 0.17 &  30.9 \\
\hline 
Ours-DFV & {\bf 5.70$e^{-4}$} & {\bf 2.13$e^{-2}$} & {\bf 0.17} &  33.3 \\
\hline
\end{tabular}}
\caption{Performance of different multi-scale variants.}
\label{tab:multiscale}
\vspace{-3mm}
\end{table}

\begin{figure*}[ht]
\centering
\begin{tabular}{ccccccc}
%\hspace{-2mm}\includegraphics[height =0.41in]{figures/scheme1.pdf} &
\small{Image} & \hspace{-1.5mm} \small{MobileDFF} &  \hspace{-1.5mm} \small{DDFF} & \hspace{-1.5mm} \small{DefocusNet}  &\hspace{-1.5mm} \small{AiFDepthNet} &\hspace{-1.8mm} \small{Ours-FV} & \hspace{-1.5mm} \small{Ours-DFV} \\
\includegraphics[height =1.14in]{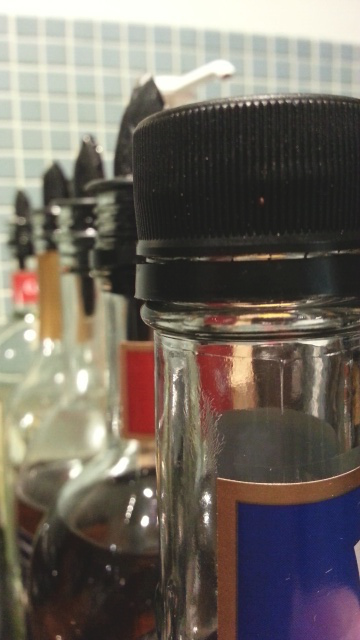} &
% \hspace{-2.5mm}\includegraphics[height =0.8in]{figures//unknown_res/telephone_img.png} &
\hspace{-1.5mm}\includegraphics[height =1.14in]{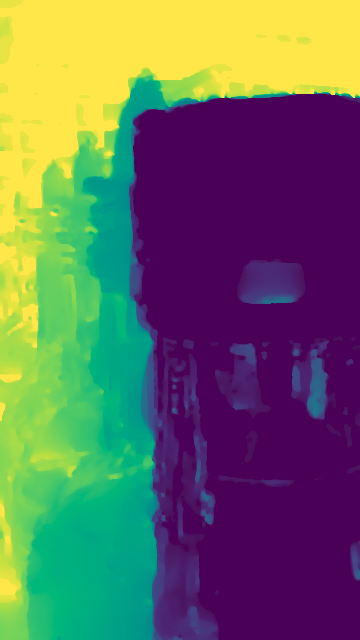} &
 \hspace{-1.5mm}\includegraphics[height =1.14in]{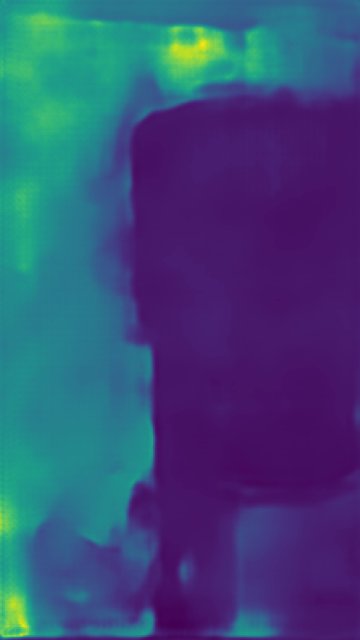}&
\hspace{-1.5mm}\includegraphics[height =1.14in]{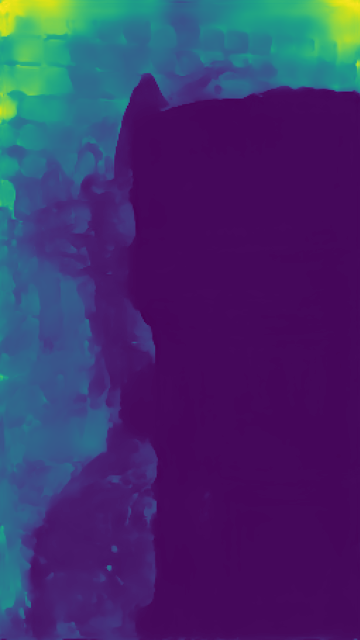} &
\hspace{-1.5mm}\includegraphics[height =1.14in]{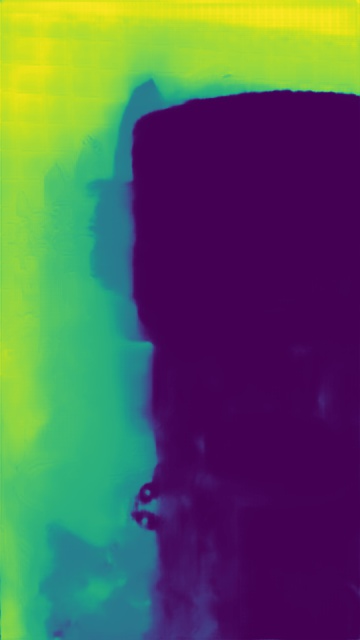} & 
\hspace{-1.8mm}\includegraphics[height =1.14in]{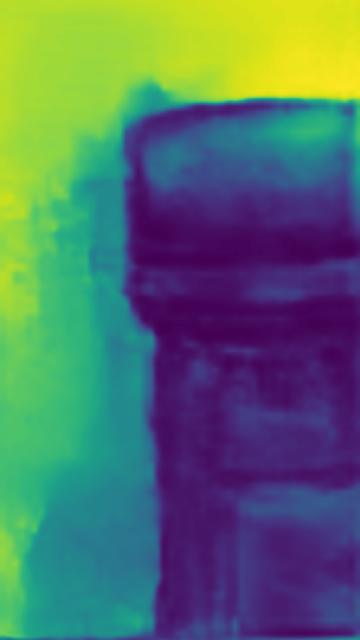}&
\hspace{-1.5mm}\includegraphics[height =1.14in]{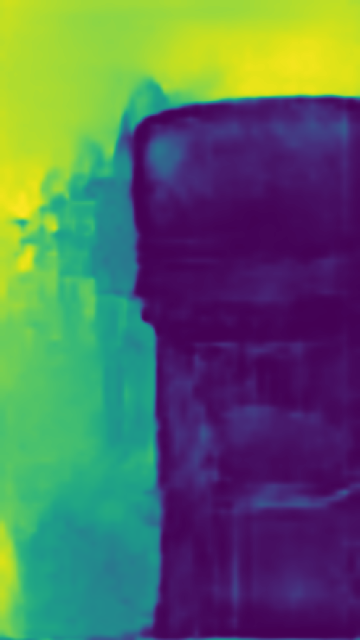} \\

\includegraphics[height=1.1in]{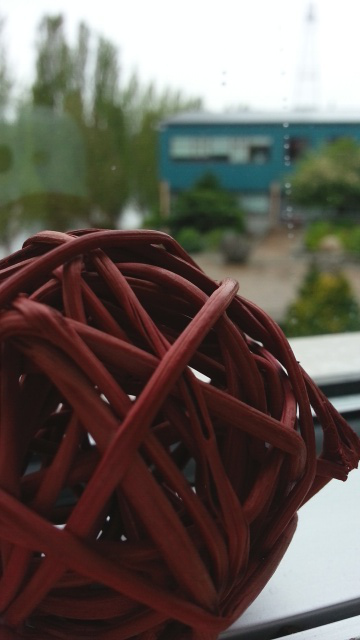} &
% \hspace{-2.5mm}\includegraphics[height =0.8in]{figures//unknown_res/telephone_img.png} &
\hspace{-1.5mm}\includegraphics[height =1.14in]{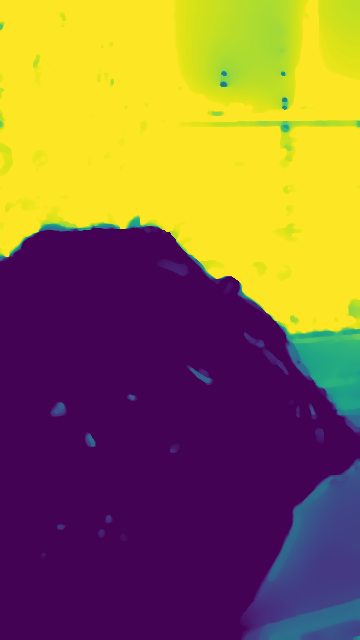} &
 \hspace{-1.5mm}\includegraphics[height =1.14in]{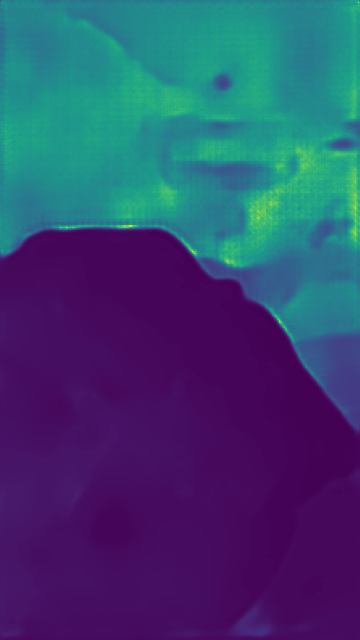}&
\hspace{-1.5mm}\includegraphics[height =1.14in]{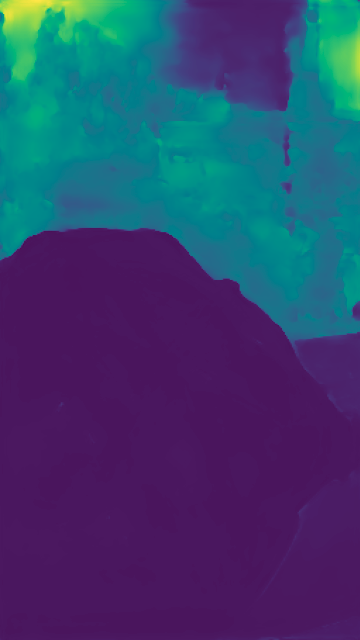} &
\hspace{-1.5mm}\includegraphics[height =1.14in]{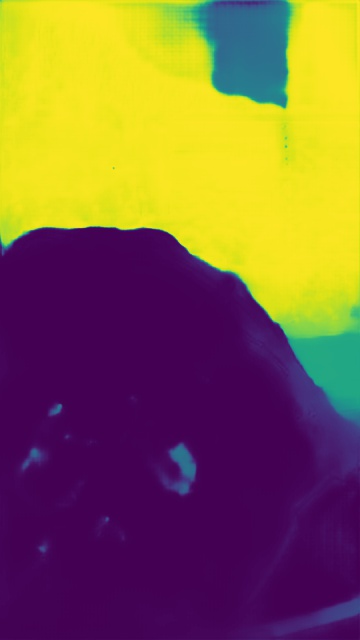} &
\hspace{-1.8mm}\includegraphics[height =1.14in]{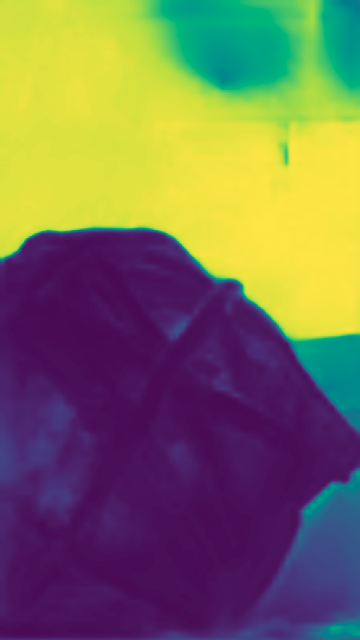}& 
\hspace{-1.5mm}\includegraphics[height =1.14in]{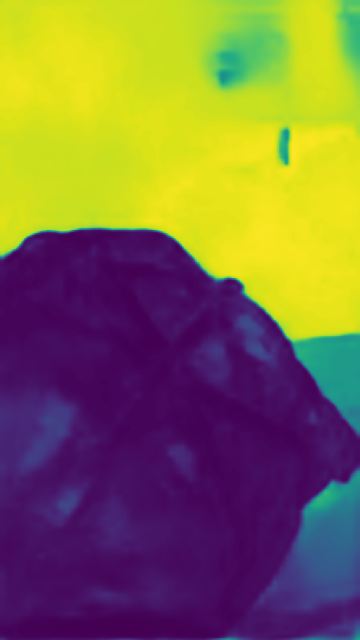}   \\
\includegraphics[height
=1.14in]{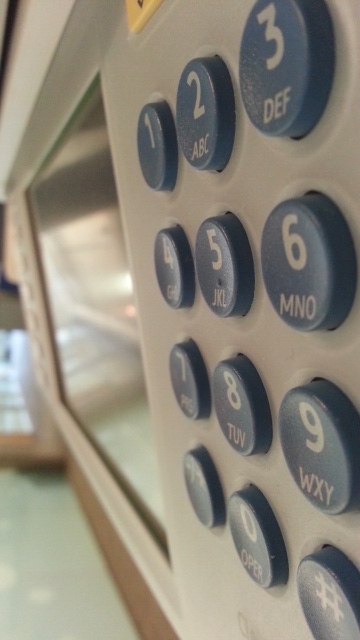} &
% \hspace{-2.5mm}\includegraphics[height =0.8in]{figures//unknown_res/telephone_img.png} &
\hspace{-1.5mm}\includegraphics[height =1.14in]{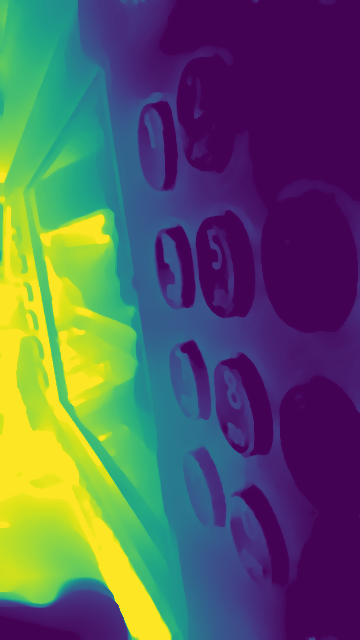} &
 \hspace{-1.5mm}\includegraphics[height =1.14in]{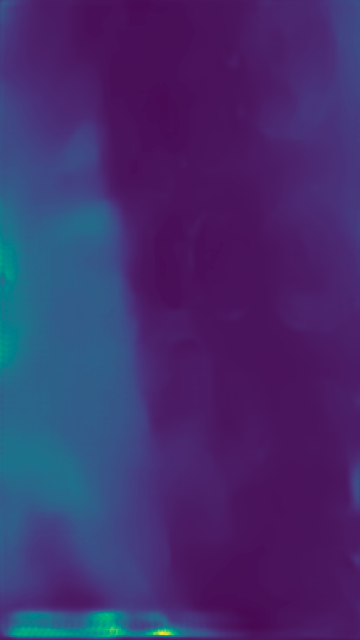}&
\hspace{-1.5mm}\includegraphics[height =1.14in]{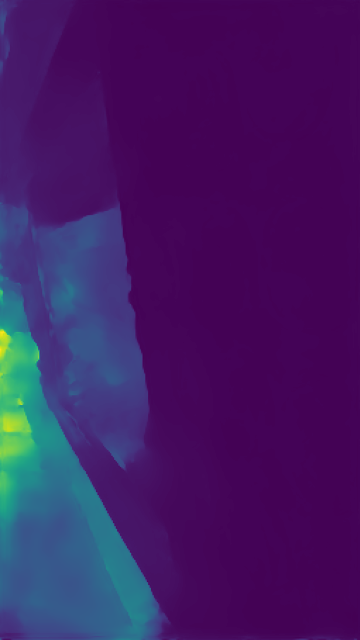} &
\hspace{-1.5mm}\includegraphics[height =1.14in]{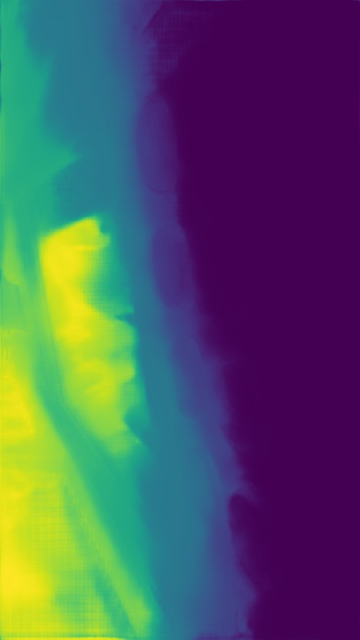}  &
\hspace{-1.8mm}\includegraphics[height =1.14in]{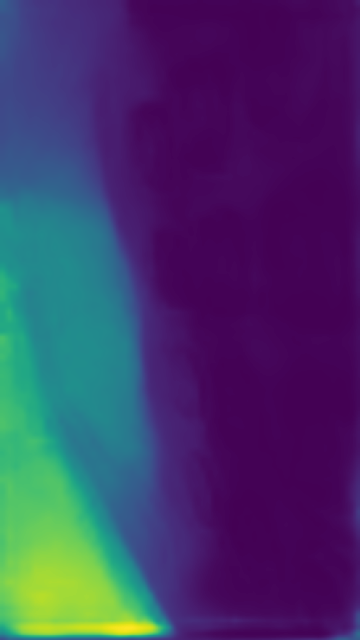}&
\hspace{-1.5mm}\includegraphics[height =1.14in]{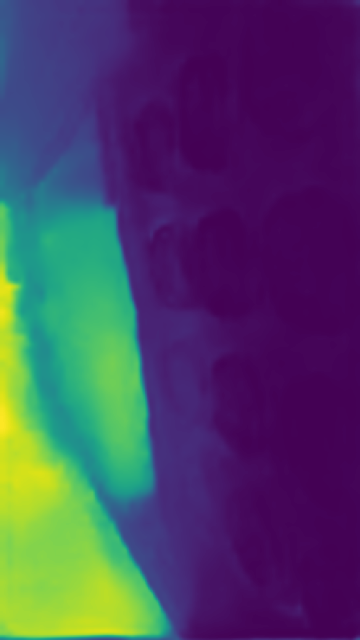}
\\

\end{tabular}
\vspace{-2mm}
\caption{Qualitative results on Mobile depth dataset. The warmer the color, the larger the depth value.}
\label{fig:mobileDepth}
\vspace{-3mm}
\end{figure*}

\subsection{Generalization to Unknown Scenes}
Model generalizability to unknown scenes is important for deep DFF methods. To evaluate our model's generalizability, we directly test the pre-trained DDFF~\cite{hazirbas2018DFFdataset},  DefocusNet~\cite{maximov2020focusondefocus}, Ours-FV, and {Ours-DFV} models, from Sec.~\ref{sec:exp_res}, on the Moblie depth dataset, and compare them with MobileDFF~\cite{suwajanakorn2015mobiledepth} and AiFDepthNet~\cite{wang2021bridging}.  MobileDFF is a traditional method that combines the merits of DFF and DFD. The results are directly from the authors.  
%AiFDepthNet is a recent deep learning method pre-trained on a large amount of data including DDFF-12, FoD500, FlyingThings3D~\cite{mayer2016large}, and 4D Light Field~\cite{honauer2016dataset}. 
For AiFDepthNet, as the authors only release their pre-trained models, we apply their depth supervised Mobile depth model for the experiment. Because of the lack of ground truth depth, we normalize all the results and compare them visually. 

Part of the results are shown in Figure~\ref{fig:mobileDepth}, and the rest are available in the supplementary material. All the deep methods provide reasonable depth estimation. Compared to the other deep methods, Ours-DFV better preserves object boundaries, \eg, the front bottle in the first row and the ball in the second row. Ours-DFV is also less sensitive to background patterns, such as the tile in the first row. However, all methods (except DDFF) are sensitive to mirror reflection (the screen in the third row). One possible reason is that the effective distance between the camera and the object shown on the screen is equal to the sum of the distance between the camera and the screen and the distance between the screen and the object. The CoC becomes larger, so all five methods infer it as an object at a farther distance. The DDFF network is adopted from the general dense prediction task, which may focus more on the context information, and thus gives a better estimation in such a region. % More context information has to be utilized to make a right depth inference in such a region.  

Among all the methods, MobileDFF appears to recover the most details. This is partially because MobileDFF takes the whole focal stack (ranging from 14 to 33 frame/scene) as input, whereas the deep models (including ours) only use 5 frames. Besides, MobileDFF is an optimization method which takes several minutes to process one scene~\cite{suwajanakorn2015mobiledepth}. Meanwhile, our methods run in tens of milliseconds.

\section{Conclusion}
In this work, we developed a novel CNN architecture for the DFF task. To some extent, our network design is analogous to traditional DFF methods and recent deep stereo matching methods. The proposed deep differential focus volume module is able to combine both the focus and the context information for focus estimation. Experiments on various datasets demonstrated the superior performance of our models in terms of accuracy, efficiency, and generalizability. Currently, our models can only work on well-aligned focal stacks in static scenes. In the future, it is interesting to explore specific data augmentation methods for imperfectly aligned stacks. We also plan to integrate the optical flow~\cite{suwajanakorn2015mobiledepth} or homography~\cite{jeon2019ring} alignment mechanism into the model and extend 
it to dynamic environments.

\noindent{\bf Acknowledgement.} This work is supported in part by NSF
award \#1815491.

%%%%%%%%% REFERENCES
{\small
\bibliographystyle{ieee_fullname}
\bibliography{egbib}
}

%%%%%%%%% appendix 
\appendix
\section{Appendix}
\subsection{Network Architecture}
\label{sec:net}
\begin{figure*}[htp]
\centering
\includegraphics[width=1.0\linewidth]{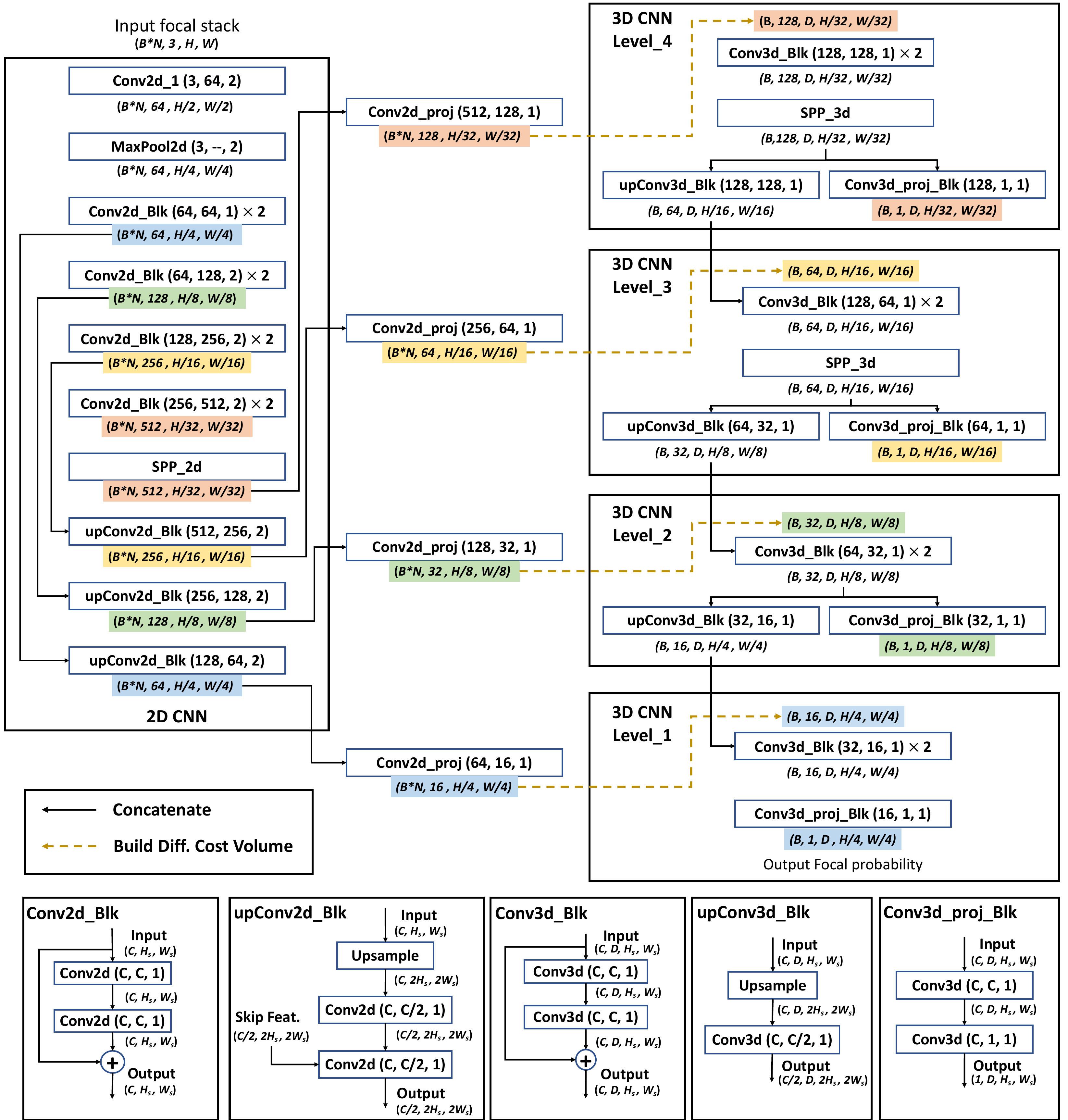}
\caption{Our differential focus volume network architecture. The $B$, $N$, $H$, and $W$ are the batch size, the input frame number the height, and the width, respectively. The subscript $s$ indicates the scale level.}
\label{fig:fullNet}
\end{figure*}

Figure~\ref{fig:fullNet} presents our architecture design. The overall framework is similar to~\cite{yang2019hsm}, but we optimize the 2D encoder and the 3D CNN decoders for the DFF task. Given $B$ focal stacks with $N$ frames in each, we first reshape them into a $B \cdot N \times 3 \times H \times W$ tensor and pass it to the 2D CNN to extract features in four different scales.  Four differential focus volumes (DFVs) are then built based on the features, which later are used to produce the focus probability distributions volume in the corresponding scale as outputs. In the end, all these outputs are upsampled with linear interpolation to the full resolution ($B \times N \times H \times W$) followed by depth probability regression for deep supervision at training time.  At test time, only the largest (Level\_1 in Figure~\ref{fig:fullNet}) scale output is upsampled for the depth regression. Here, $B, N, H$ and $W$  denote the batch dimension, the frame dimension, the height dimension, and the width dimension, respectively.

For all the convolution layers in the figure, the three numbers in the in-box parenthesis indicate the in-feature channel number, the out-feature channel number, and the convolution stride, respectively. The parenthesis below the box presents the output size. All convolution layers use kernel size 3, except ``Conv2d\_1" ,  ``Conv2d\_proj" and the last convolution in  ``upConv3d\_Blk" and ``Conv3d\_proj\_Blk". The former one uses kernel size 7, and the latter three use size 1. We use batch normalization followed with ReLU for all convolution operations, except the final convolution in ``Conv3d\_proj\_Blk" where softmax is applied to the $N$ dimension. No activation function is applied to  ``Conv2d\_proj" or the last convolution in ``upConv3d\_Blk". For 3D spatial pyramid pooling (SPP) module, we use four pooling scales $m_a$ linearly sampled from $1$ to $\floor*{\frac{min(N, H, W)}{2}}$, where $a=1, ..., 4$, and the corresponding pooling kernel size $k_a = \floor*{\frac{N_a}{m_a},\frac{H_a}{m_a},\frac{W_a}{m_a}}$. For example, given an input in the shape $(10, 14, 14)$, the 4 scales will be $\{1, 2, 3, 5\}$, and the corresponding pooling kernel sizes are $\{(10, 14, 14), (5, 7, 7), (3, 4, 4), (2, 2, 2)\}$. We do not include 3D SPP in the last two levels for the speed and accuracy trade-off. The 2D SPP kernel sizes are the same as the 3D SPP, except that the $N$ dimension is excluded.

\begin{table*}[ht]
\centering

\caption{Ours-DFV results on DDFF-12 {validation} set with various number of input frames.}
\label{tab:var_nfrm}
 \setlength{\tabcolsep}{3pt}
{\small
\begin{tabular}{cc|cccccccccccc}
\hline
Method & \#Frm & MSE $\downarrow$  & RMS$\downarrow$ & log RMS $\downarrow$ & Abs. rel.$\downarrow$& Sqr. rel.$\downarrow$ & $\delta\uparrow$ & $\delta^2\uparrow$ & $\delta^3\uparrow$ &   Bump.$\downarrow$ & avgUnc.$\downarrow$ & Time(ms)$\downarrow$\\
\hline 

\multirow{9}{*}{Ours-DFV} & 2 & 9.68$e^{-4}$ & 2.89$e^{-2}$ & 2.87$e^{-1}$  &2.47$e^{-1}$ & 10.26$e^{-3}$ &60.79 & 88.53 & 96.41 & 4.35$e^{-1}$ & 10.73$e^{-2}$ & {\bf 19.9}\\
& 3 &  6.66$e^{-4}$ & 2.35$e^{-2}$ & 2.36$e^{-1}$  & 2.00$e^{-1}$ & 7.22$e^{-3}$ & 71.13 & 93.28 & 97.91 & 4.26$e^{-1}$ & 7.80$e^{-2}$ & 22.8 \\
& 4 &  6.45$e^{-4}$ & 2.29$e^{-2}$ & 2.34$e^{-1}$  & 1.90$e^{-1}$ & 6.92$e^{-3}$ & 72.66 & 93.46 & 97.84 & {\bf 4.10$e^{-1}$} & 6.05$e^{-2}$ & 27.8 \\
& 5 & 6.63$e^{-4}$ & 2.35$e^{-2}$ & 2.39$e^{-1}$  & 1.86$e^{-1}$ & 6.92$e^{-3}$ & 70.17 & 92.94 &97.96 & 4.21$e^{-1}$ & 5.39$e^{-2}$ & 33.3 \\
& 6 & 6.01$e^{-4}$ & 2.20$e^{-2}$ & 2.25$e^{-1}$  & 1.73$e^{-1}$ & 6.50$e^{-3}$ & 75.65 & 93.53 &97.80 & {4.16$e^{-1}$} & 4.58$e^{-2}$ & 38.6 \\
& 7 & 6.19$e^{-4}$ & 2.22$e^{-2}$ & 2.27$e^{-1}$  & 1.77$e^{-1}$ & 6.43$e^{-3}$ & 74.77 & 93.45 &97.80 & 4.19$e^{-1}$ & 3.90$e^{-2}$ & 44.5 \\
& 8 & {5.92$e^{-4}$} & {\bf 2.16$e^{-2}$} & {\bf 2.14$e^{-1}$ } & {\bf 1.68$e^{-1}$} & 6.00$e^{-3}$ & {\bf 77.92} & 94.67 & 98.03 & 4.22$e^{-1}$ & 3.96$e^{-2}$ & 48.4\\
& 9 & {\bf 5.90$e^{-4}$} & {2.18$e^{-2}$} & 2.23$e^{-1}$  & 1.86$e^{-1}$ & 6.48$e^{-3}$ & 74.02 & 94.58 & {\bf 98.20} & 4.20$e^{-1}$ & 3.18$e^{-2}$ & 55.1\\
& 10 & 6.45$e^{-4}$ & 2.24$e^{-2}$ & 2.18$e^{-1}$  & 1.68$e^{-1}$ & {\bf 5.57$e^{-3}$} & 75.52 & {\bf 94.77} & {98.08} & {4.17$e^{-1}$} & {\bf 2.05$e^{-2}$} & 59.6\\
\hline
DDFF~\cite{hazirbas2018DFFdataset} & 5 & 11.84$e^{-4}$ & 3.05$e^{-2}$ & 2.85$e^{-1}$ & 2.19$e^{-1}$ & 8.36$e^{-3}$ & 56.00 & 87.60 & 97.11 & 4.37$e^{-1}$ & -- & 191.7\\
DefocusNet~\cite{maximov2020focusondefocus} &5 & 8.57$e^{-4}$ & 2.56$e^{-2}$ & 2.48$e^{-1}$  & 1.80$e^{-1}$  & {6.94}$e^{-3}$ & 73.16 & 92.04 & 96.86 & 4.45$e^{-1}$  & --&34.4\\
\hline
\end{tabular}}
\end{table*}

\subsection{Focus Probability Visualization}
\label{sec:focus_viz_sec}
As Figure~\ref{fig:fullNet} illustrates, the direct output of our network in the scale level $s$ is the focus probability distribution $P^s$ (batch $\times$ 1 $\times$ frame\_ID $\times$ height $\times$ width), where $p^s_{(b, 0, i, u, v)}$ indicates the probability of the pixel at coordinate $(u, v)$ in the frame $i$ sample $b$ to be the best-focused pixel. From this perspective, the whole network can be viewed as a deep focus measure.  

\begin{figure*}[htp]
\begin{tabular}{cccccc}
\label{tab:focal_size}
    \centering
    \hspace{-0.5mm}Image &  \hspace{-0.5mm}Frame-1 & \hspace{-0.5mm} Frame-2 &\hspace{-0.5mm}  Frame-3 &\hspace{-0.5mm}  Frame-4 &\hspace{-2.0mm}  Frame-5 \\
\hspace{-2.5mm}\includegraphics[height=1.10in]{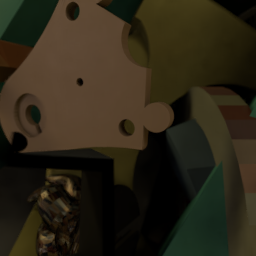} &
 \hspace{-2.7mm}\includegraphics[height =1.10in]{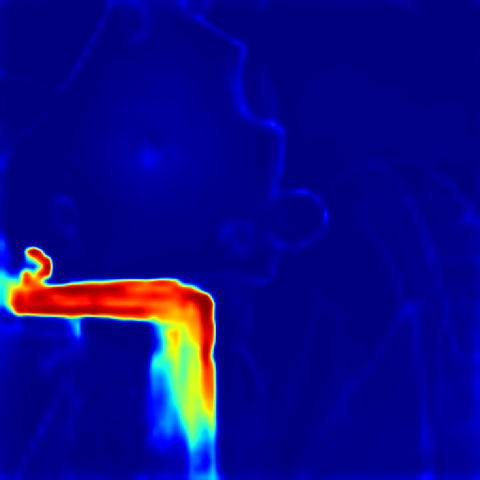}&
\hspace{-2.7mm}\includegraphics[height =1.10in]{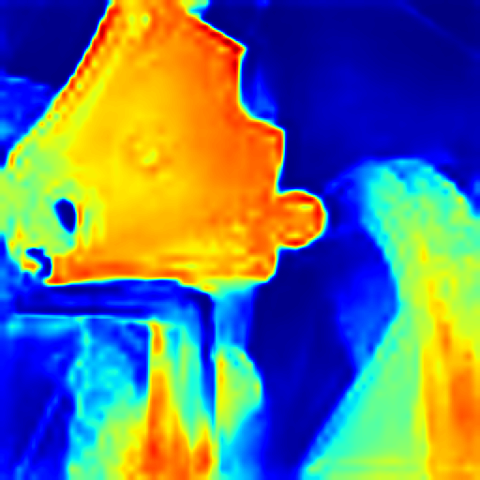}&
\hspace{-2.7mm}\includegraphics[height =1.10in]{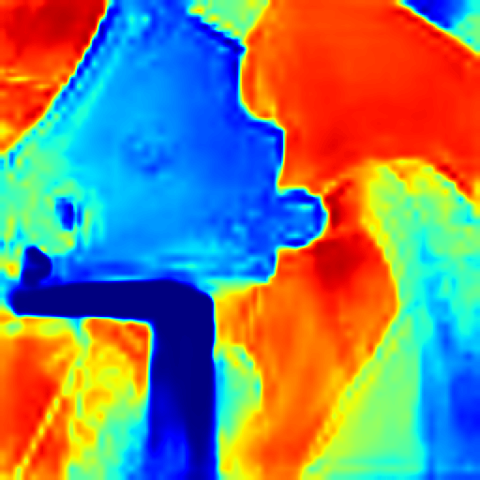}&
\hspace{-2.7mm}\includegraphics[height =1.10in]{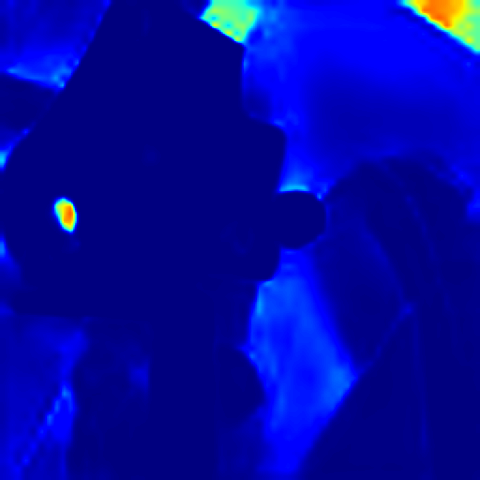} & 
\hspace{-2.7mm}\includegraphics[height =1.10in]{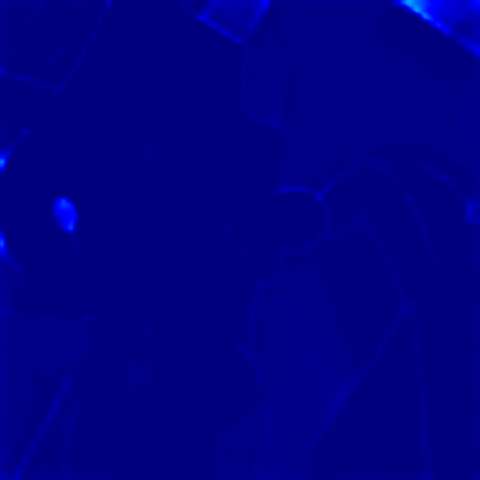}\\

\hspace{-2.5mm}\includegraphics[height=1.10in]{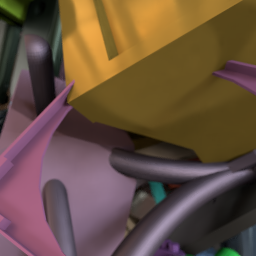} &
 \hspace{-2.7mm}\includegraphics[height =1.10in]{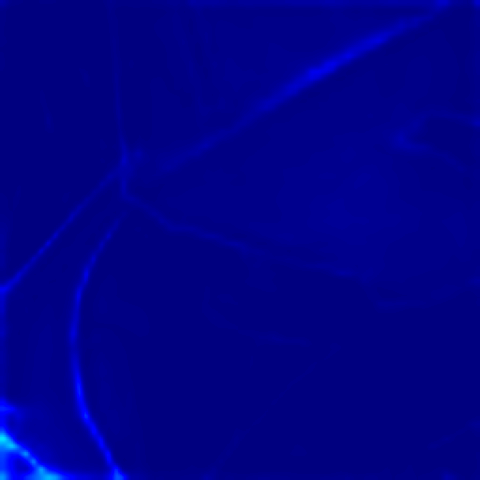}&
\hspace{-2.7mm}\includegraphics[height =1.10in]{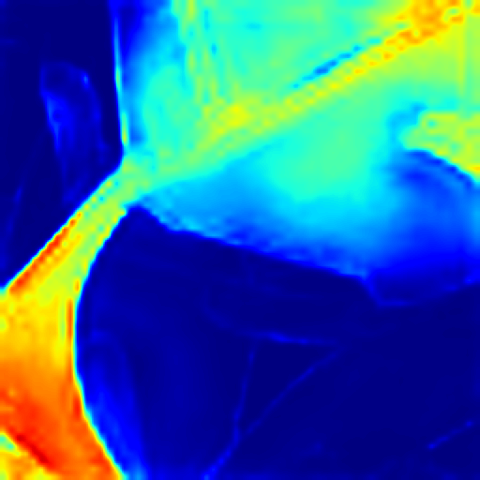}&
\hspace{-2.7mm}\includegraphics[height =1.10in]{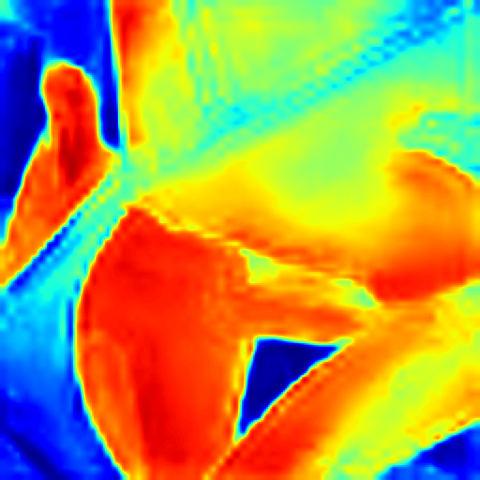}&
\hspace{-2.7mm}\includegraphics[height =1.10in]{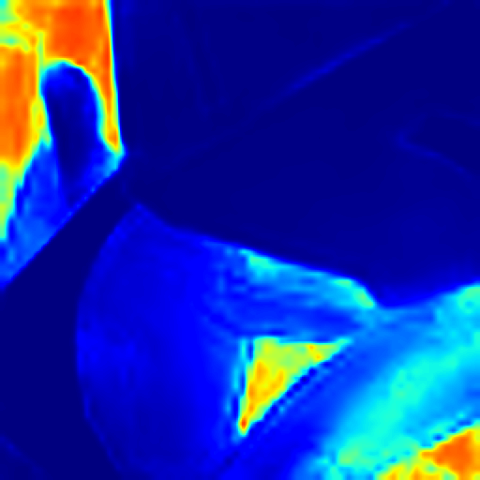} & 
\hspace{-2.7mm}\includegraphics[height =1.10in]{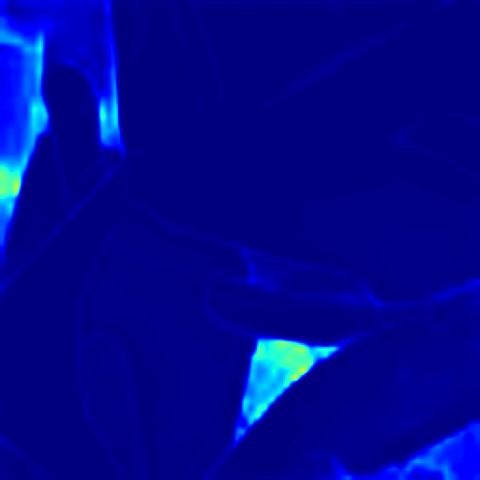}\\

\hspace{-2.5mm}\includegraphics[height=1.10in]{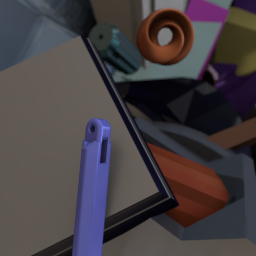} &
 \hspace{-2.7mm}\includegraphics[height =1.10in]{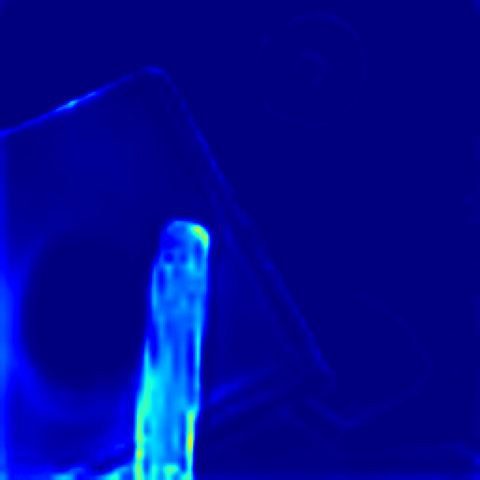}&
\hspace{-2.7mm}\includegraphics[height =1.10in]{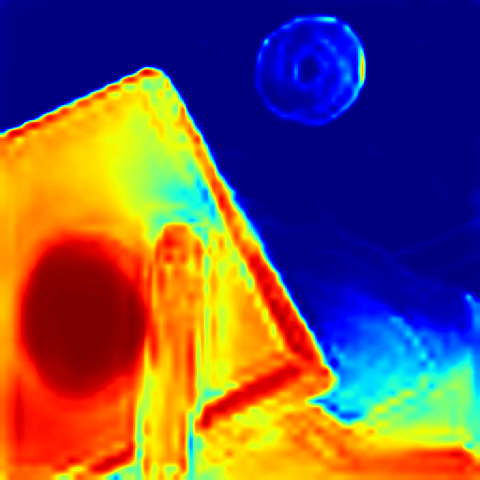}&
\hspace{-2.7mm}\includegraphics[height =1.10in]{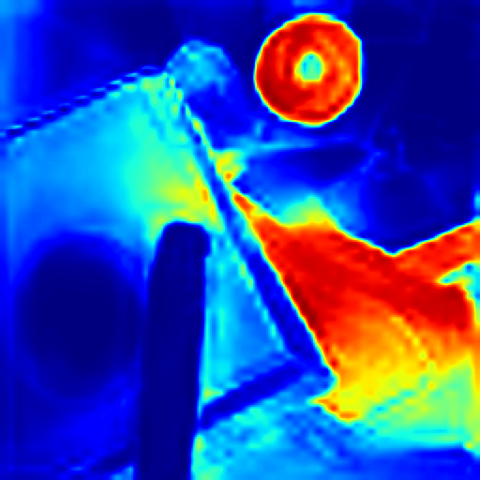}&
\hspace{-2.7mm}\includegraphics[height =1.10in]{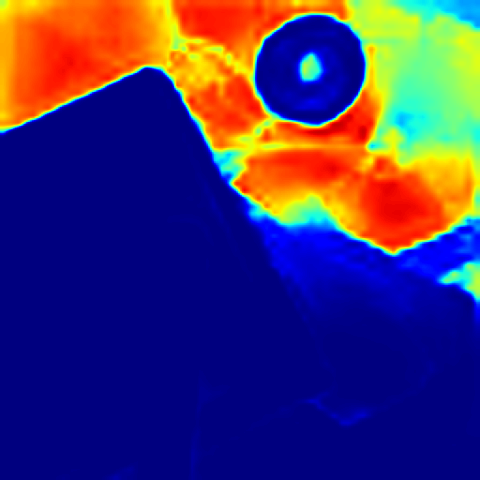} & 
\hspace{-2.7mm}\includegraphics[height =1.10in]{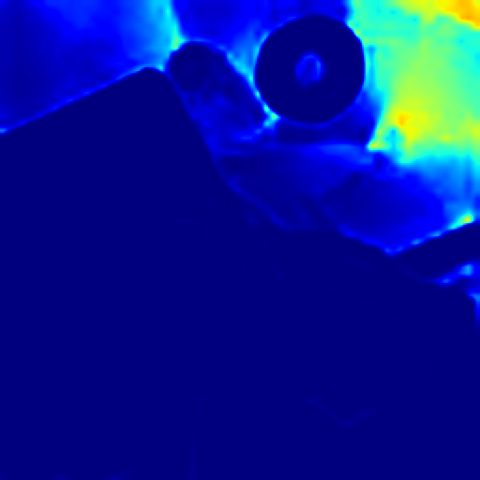}\\

% ===
\hspace{-2.5mm}\includegraphics[height=.75in]{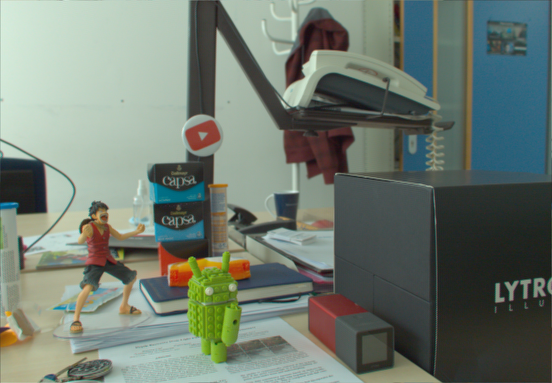} &
 \hspace{-2.5mm}\includegraphics[height =0.75in]{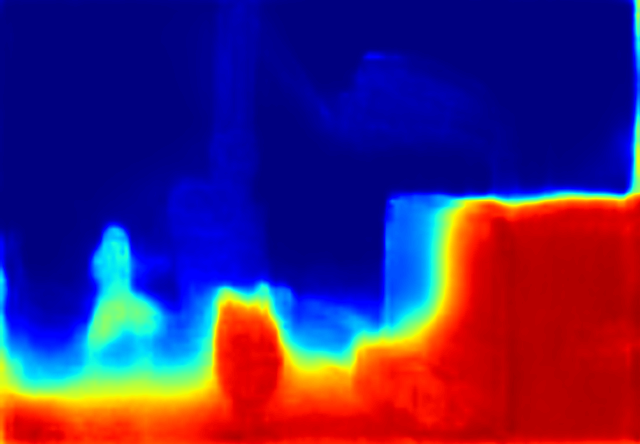}&
\hspace{-2.5mm}\includegraphics[height =0.75in]{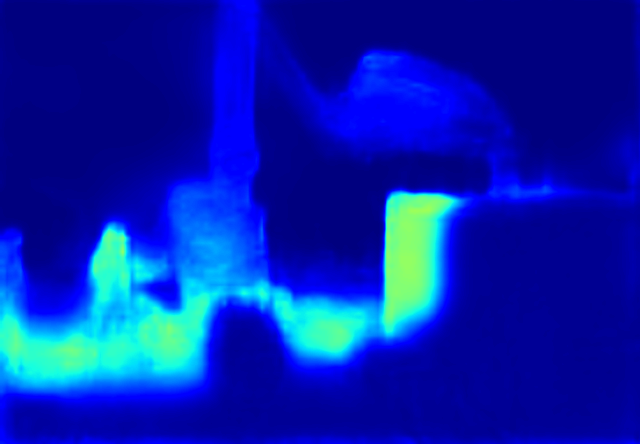}&
\hspace{-2.5mm}\includegraphics[height =.75in]{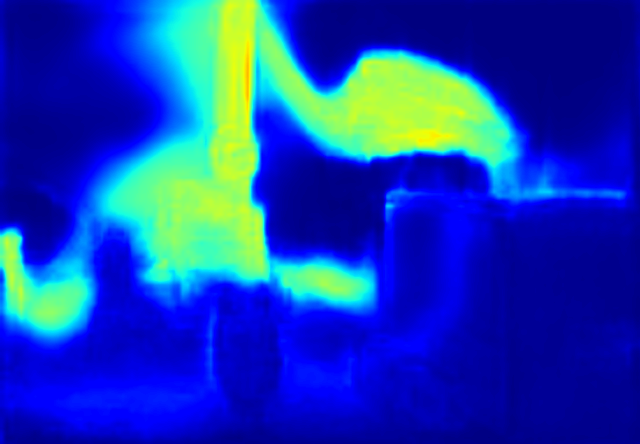}&
\hspace{-2.5mm}\includegraphics[height =.75in]{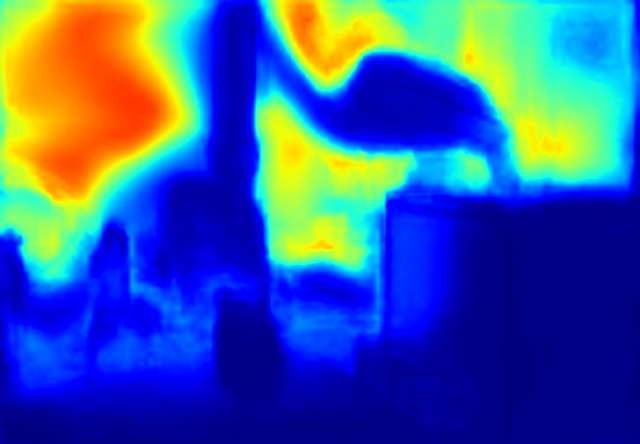} & 
\hspace{-2.5mm}\includegraphics[height =.75in]{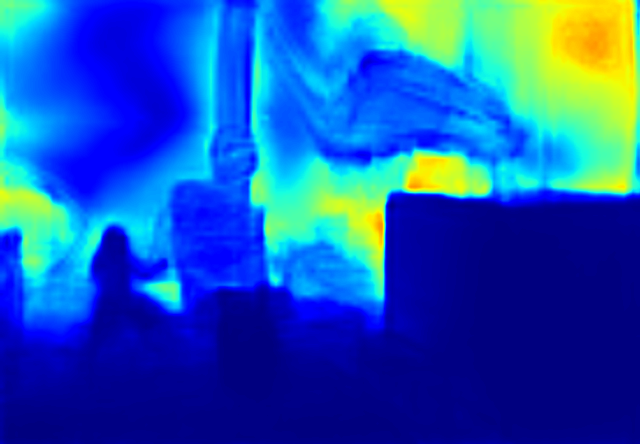}\\

\hspace{-2.5mm}\includegraphics[height=.75in]{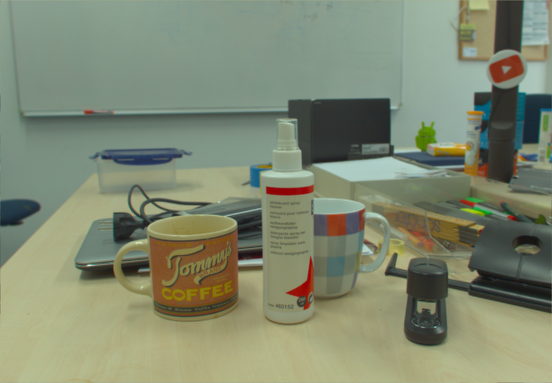} &
 \hspace{-2.5mm}\includegraphics[height =0.75in]{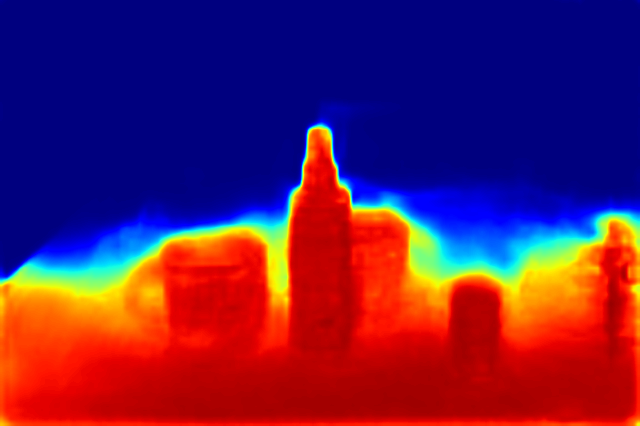}&
\hspace{-2.5mm}\includegraphics[height =0.75in]{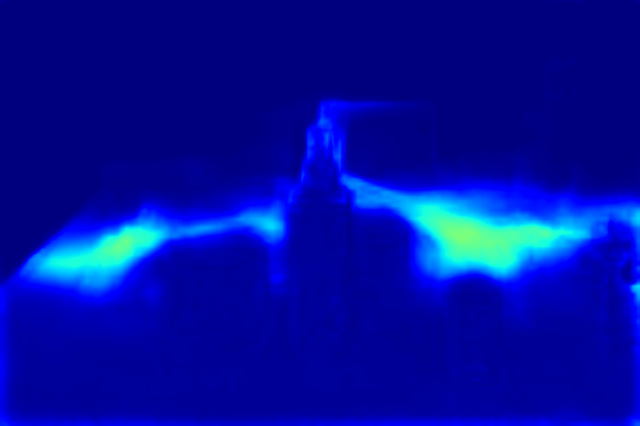}&
\hspace{-2.5mm}\includegraphics[height =.75in]{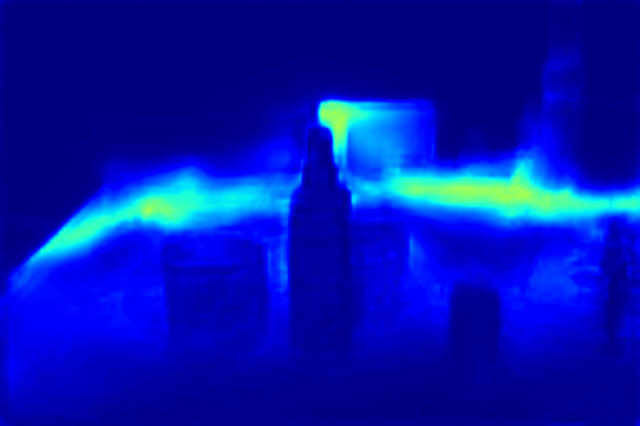}&
\hspace{-2.5mm}\includegraphics[height =.75in]{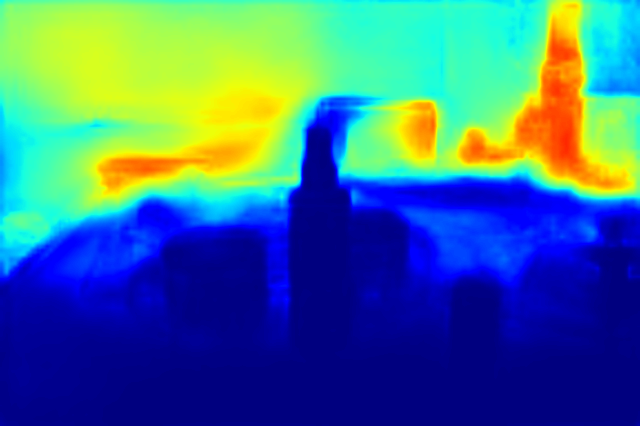} & 
\hspace{-2.5mm}\includegraphics[height =.75in]{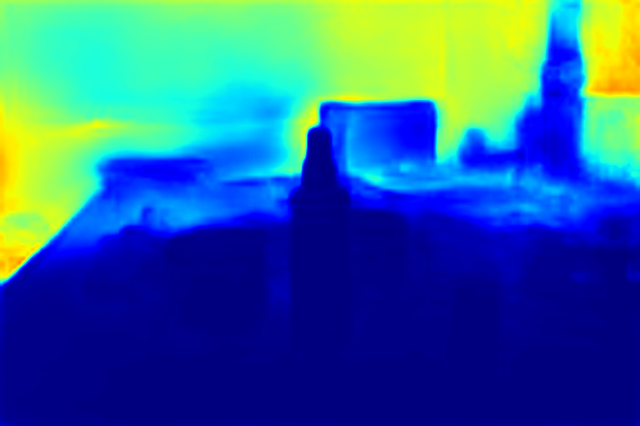}\\

\hspace{-2.5mm}\includegraphics[height=.75in]{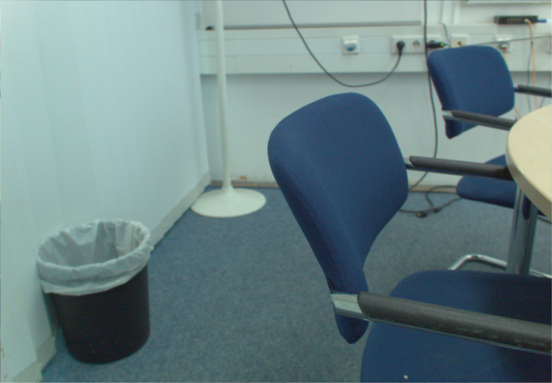} &
 \hspace{-2.5mm}\includegraphics[height =0.75in]{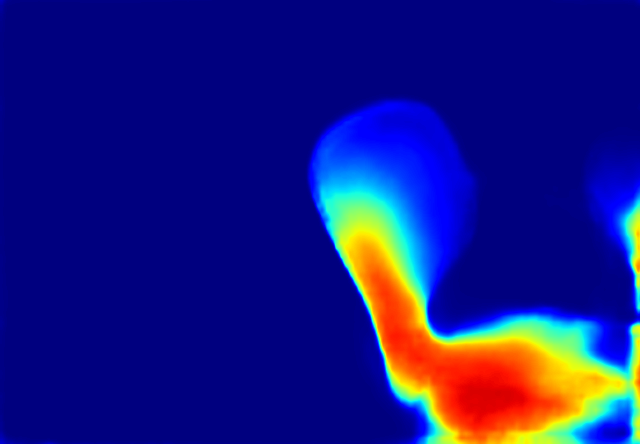}&
\hspace{-2.5mm}\includegraphics[height =0.75in]{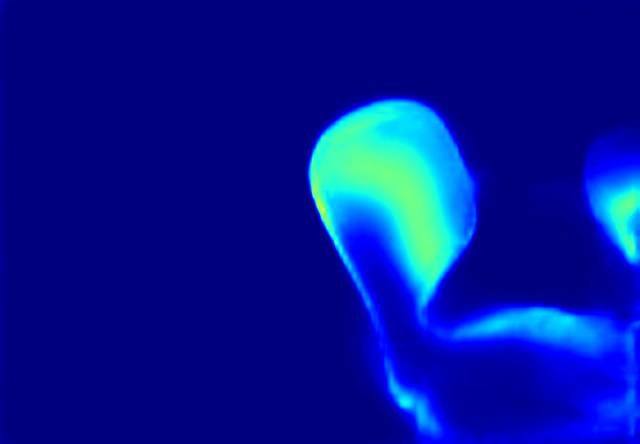}&
\hspace{-2.5mm}\includegraphics[height =.75in]{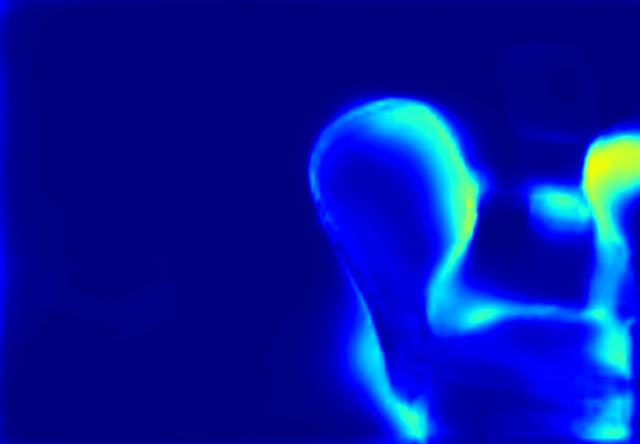}&
\hspace{-2.5mm}\includegraphics[height =.75in]{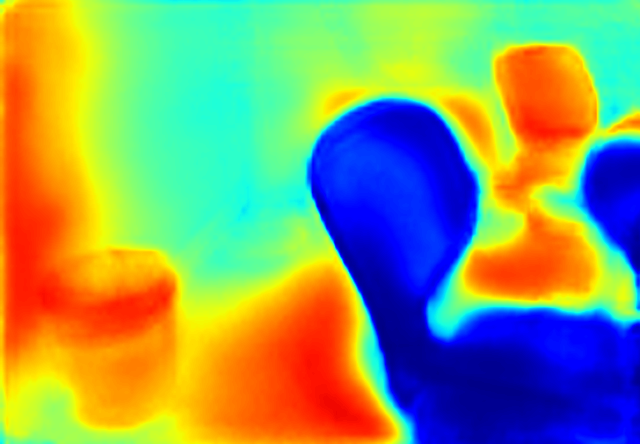} & 
\hspace{-2.5mm}\includegraphics[height =.75in]{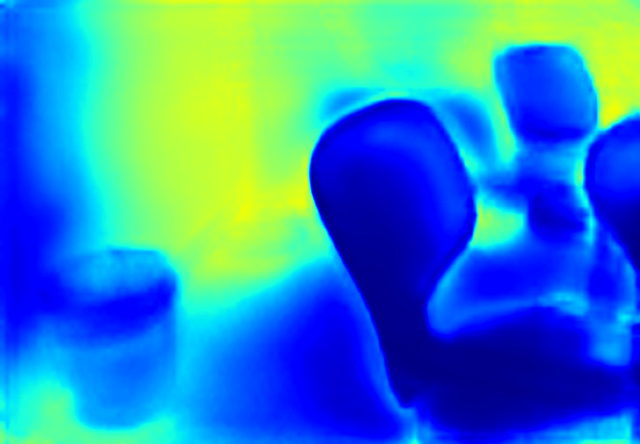}\\

% === 
\hspace{-2.7mm}\includegraphics[height =0.63in]{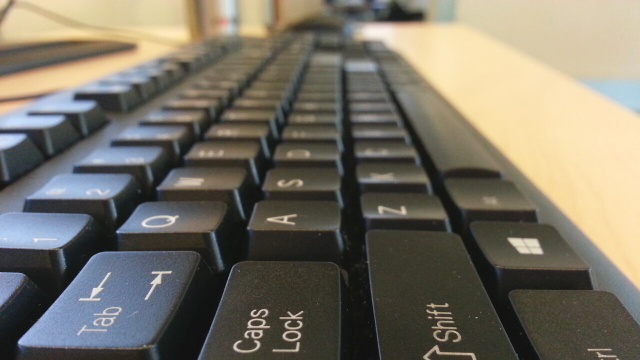}&
 \hspace{-2.7mm}\includegraphics[height =0.63in]{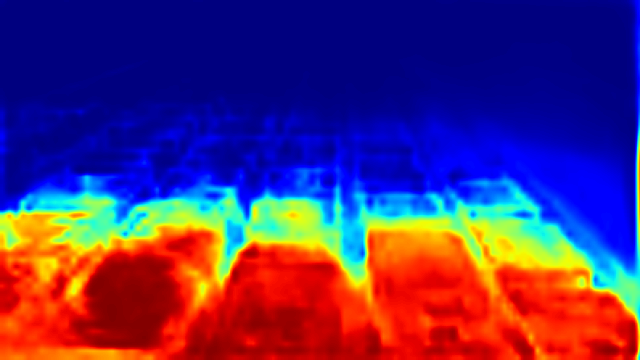}&
\hspace{-2.7mm}\includegraphics[height =0.63in]{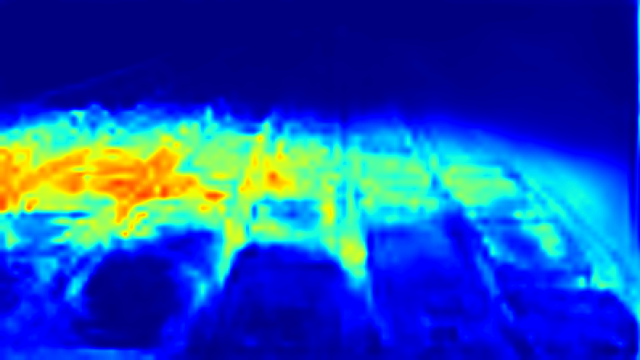}&
\hspace{-2.7mm}\includegraphics[height =.63in]{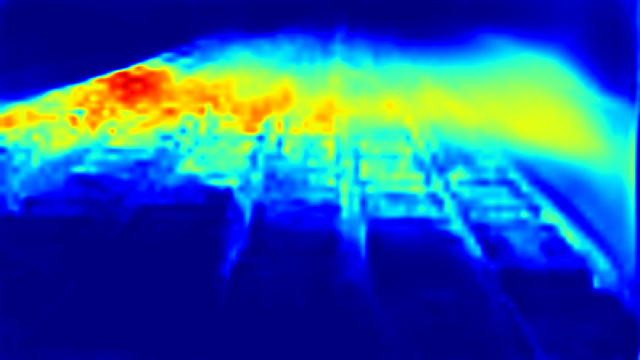}&
\hspace{-2.7mm}\includegraphics[height =.63in]{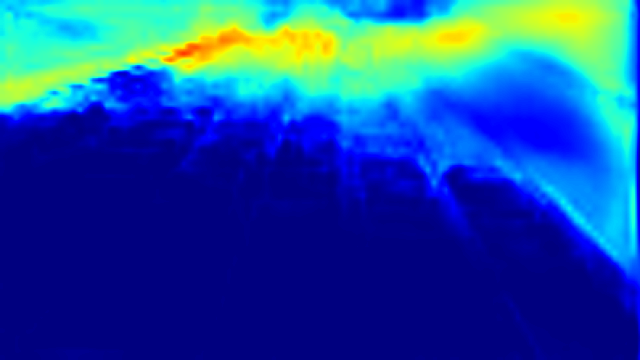} & 
\hspace{-2.7mm}\includegraphics[height =.63in]{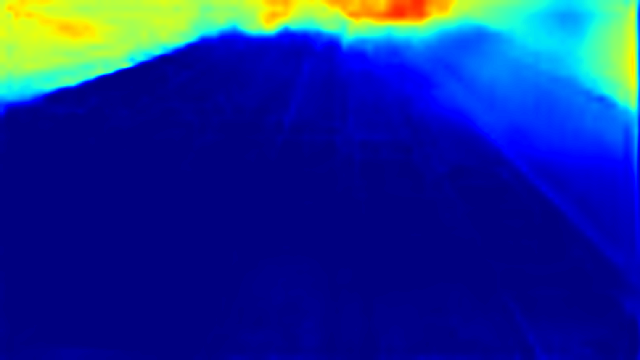}\\

\hspace{-2.7mm}\includegraphics[height =0.63in]{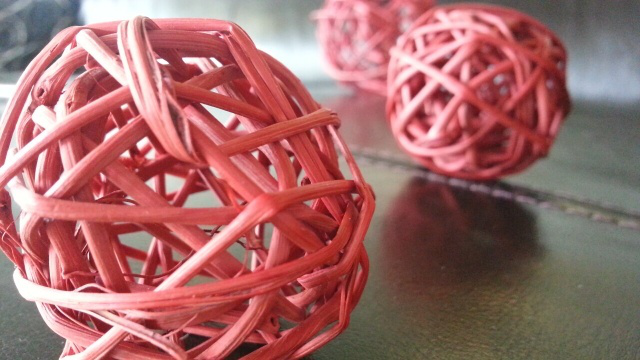}&
 \hspace{-2.7mm}\includegraphics[height =0.63in]{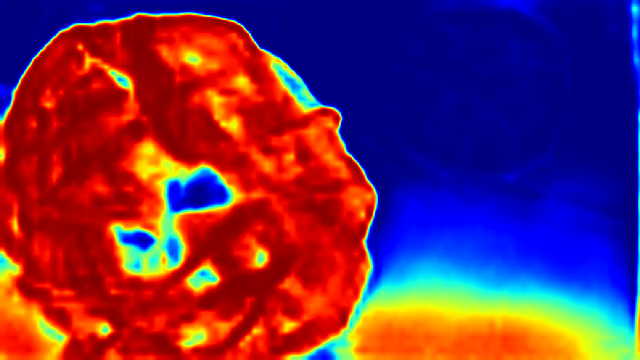}&
\hspace{-2.7mm}\includegraphics[height =0.63in]{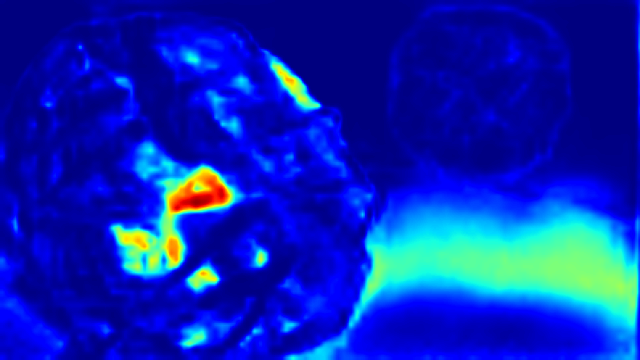}&
\hspace{-2.7mm}\includegraphics[height =.63in]{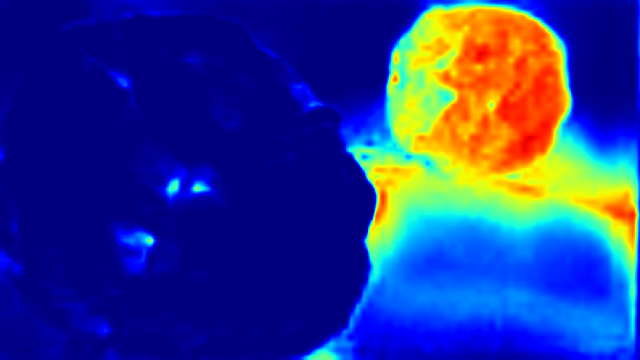}&
\hspace{-2.7mm}\includegraphics[height =.63in]{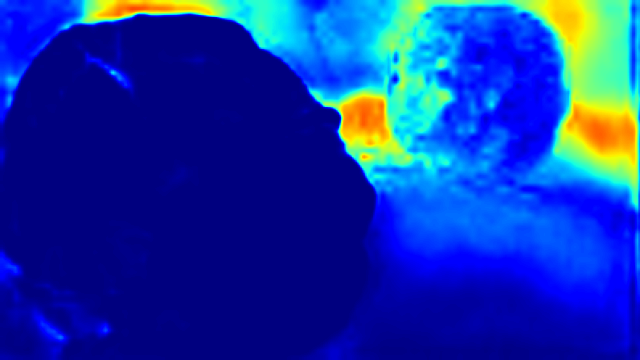} & 
\hspace{-2.7mm}\includegraphics[height =.63in]{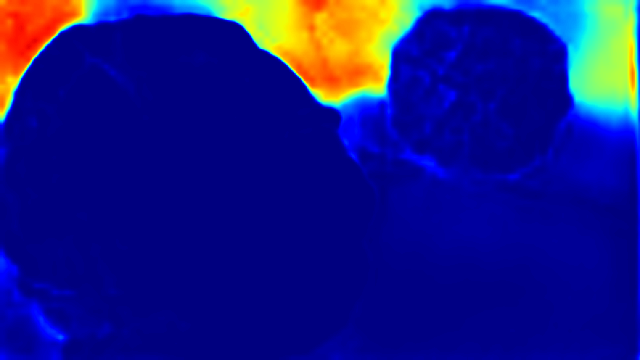}\\

\hspace{-2.7mm}\includegraphics[height =0.63in]{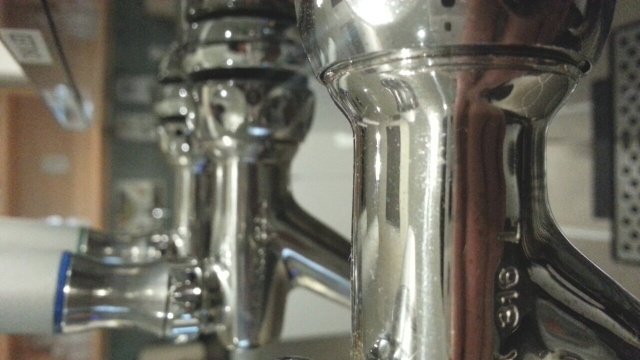}&
 \hspace{-2.7mm}\includegraphics[height =0.63in]{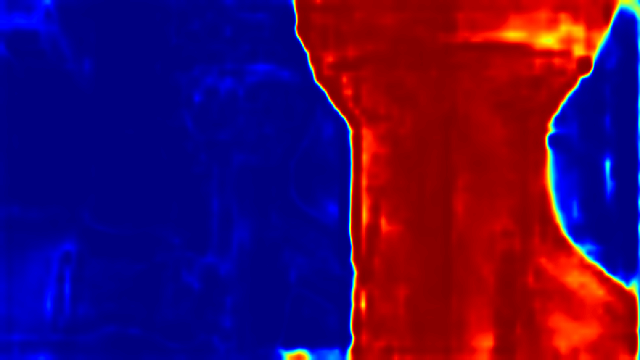}&
\hspace{-2.7mm}\includegraphics[height =0.63in]{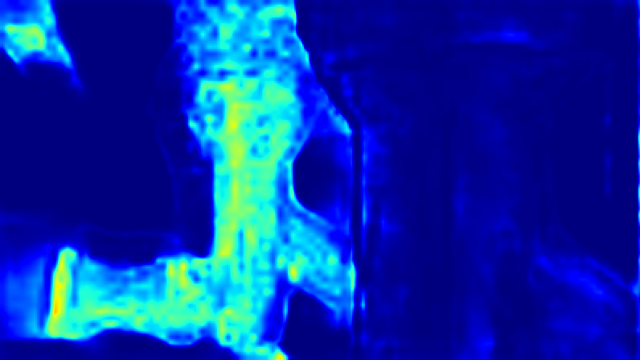}&
\hspace{-2.7mm}\includegraphics[height =0.63in]{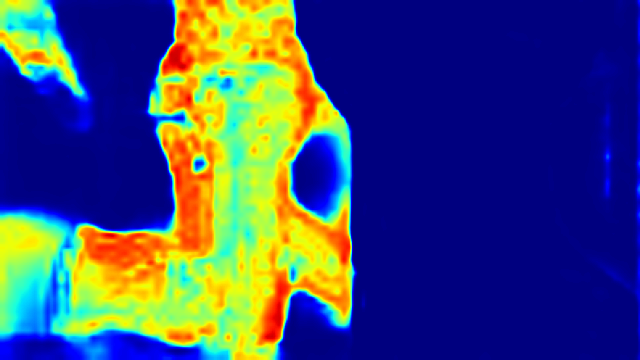}&
\hspace{-2.7mm}\includegraphics[height =0.63in]{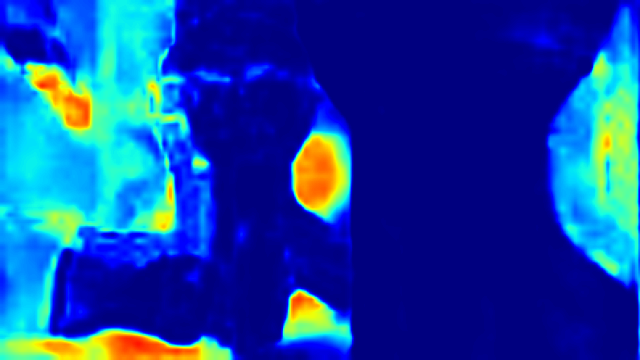} & 
\hspace{-2.7mm}\includegraphics[height =0.63in]{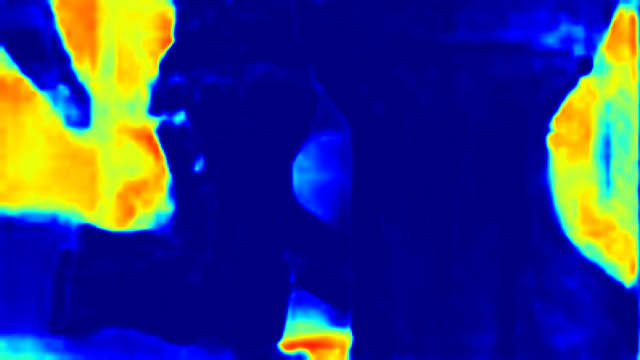}\\
\end{tabular}
\caption{Focus probability visualization. Rows 1-3, 4-6, and 7-9 are from FoD500, DDFF-12, and Mobile depth dataset, respectively. The redder the color, the higher the probability.}
\label{fig:focus_prob}
\end{figure*}

To empirically prove this, we visualize the focus probability distribution of our full method (Ours-DFV) in Figure~\ref{fig:focus_prob}. The first three rows are from FoD500 dataset~\cite{maximov2020focusondefocus}, the next three rows are from DDFF-12 dataset~\cite{hazirbas2018DFFdataset}, and the last three rows are from Mobile depth dataset~\cite{suwajanakorn2015mobiledepth}. The corresponding depth prediction results are available in Figures~\ref{fig:Fod500},~\ref{fig:ddff12}, and~\ref{fig:mobileDFF}, respectively. From Figure~\ref{fig:focus_prob}, we can observe, in all the samples, the peak of the best-focused pixel distribution move from the closest objects toward the farthest objects as frame ID increases. This aligns with our input frame order (ascending focal distances). 
%In the experiment, we also tried to randomly shuffle the frames, i.e. breaking the ascending focal distance order, but the results are not as good as the original setting. 

\subsection{Additional Qualitative Results}
\label{sec:more_res}
\begin{figure*}
\begin{tabular}{ccccc|c}

    \centering
    \hspace{-0.5mm}Image/GT &  \hspace{-0.5mm}DDFF & \hspace{-0.5mm} DefocusNet &\hspace{-0.5mm}  Ours-FV &\hspace{-0.5mm}  Ours-DFV &\hspace{-2.0mm}  {\small Uncer. Ours-FV/DFV} \\
      \hspace{-0.5mm}\includegraphics[height =1.00in]{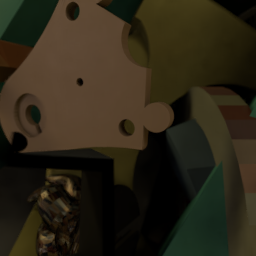} &
% \hspace{-0.5mm}\includegraphics[height =0.8in]{figures//unknown_res/telephone_img.png} &
\hspace{-0.5mm}\includegraphics[height =1.00in]{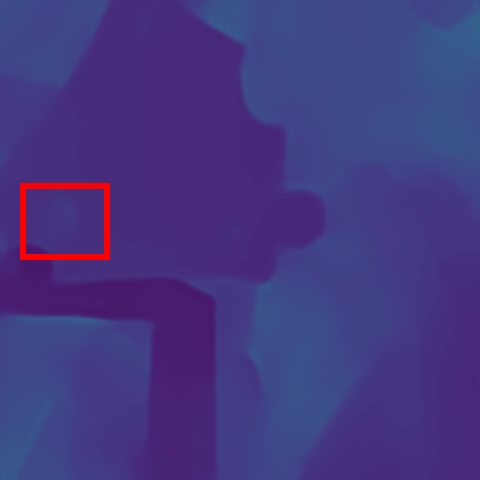} &
 \hspace{-0.5mm}\includegraphics[height =1.00in]{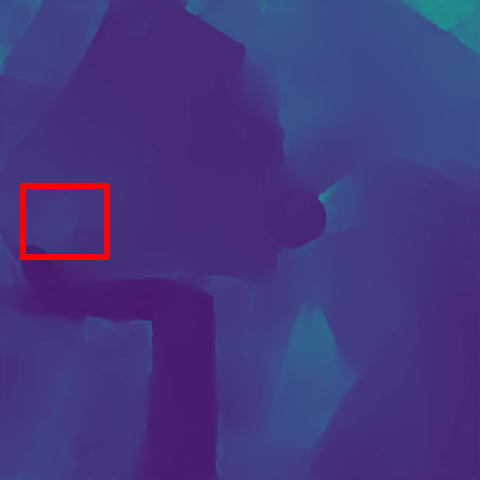}&
\hspace{-0.5mm}\includegraphics[height =1.00in]{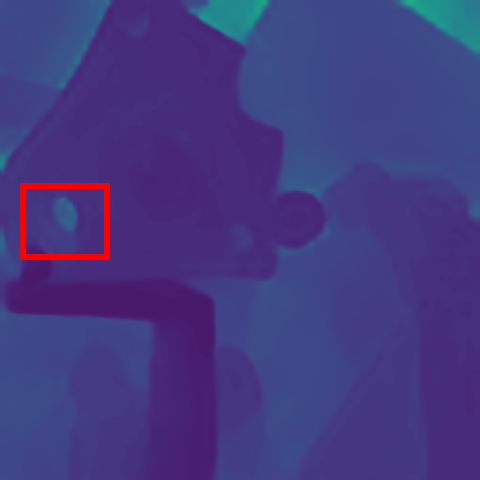} &
\hspace{-0.5mm}\includegraphics[height =1.00in]{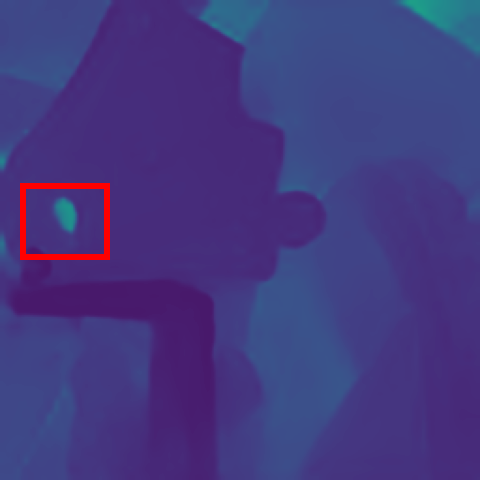} &
\hspace{-0.5mm}\includegraphics[height =1.00in]{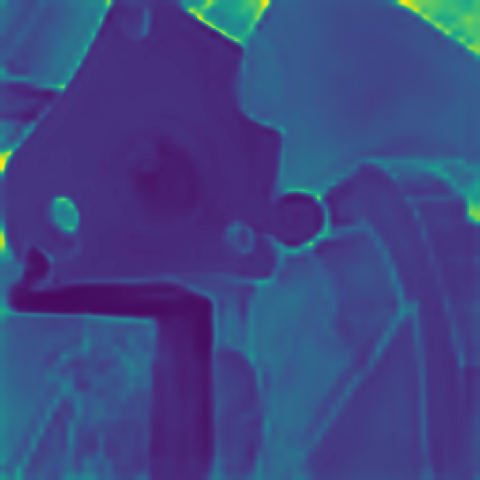}\\
\hspace{-0.5mm}\includegraphics[height= 1.00in]{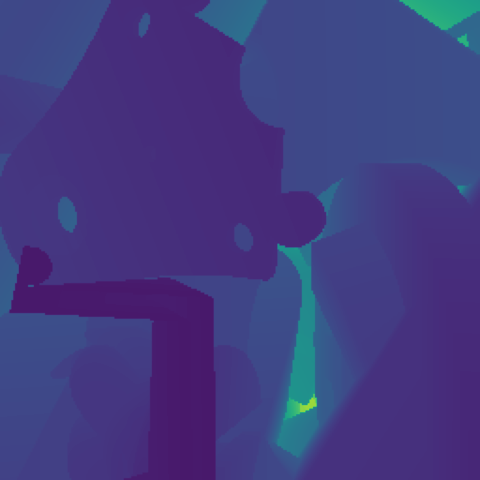} &
% \hspace{-0.5mm}\includegraphics[height= 0.8in]{figures/unknown_res/telephone_pred_viz.png} &
\hspace{-0.5mm}\includegraphics[height= 1.00in]{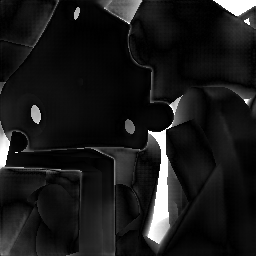} &
\hspace{-0.5mm}\includegraphics[height= 1.00in]{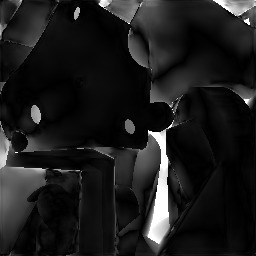} &
\hspace{-0.5mm}\includegraphics[height =1.00in]{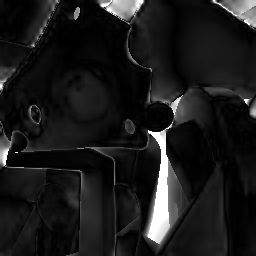} &
\hspace{-0.5mm}\includegraphics[height= 1.00in]{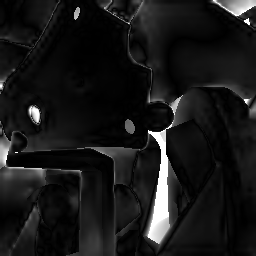} & 
\hspace{-0.5mm}\includegraphics[height =1.00in]{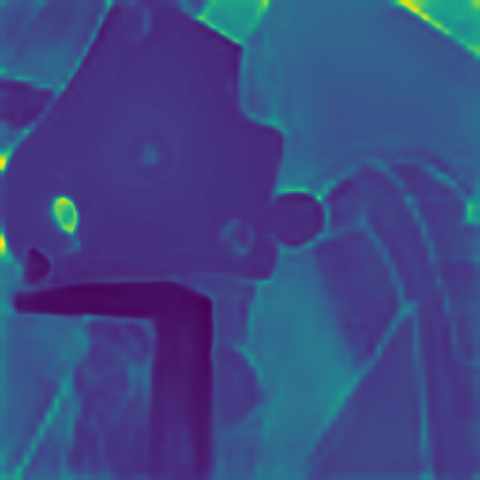}\\

\hspace{-0.5mm}\includegraphics[height =1.00in]{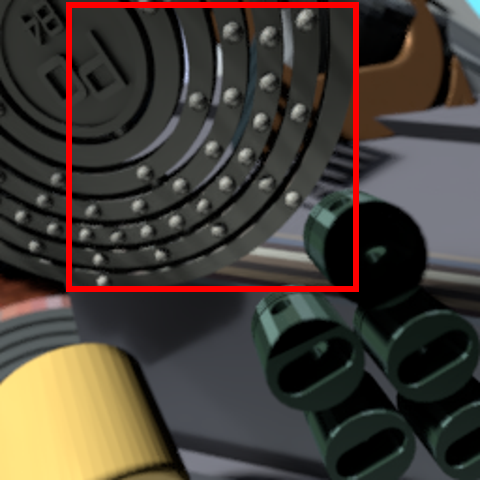} &
\hspace{-0.5mm}\includegraphics[height =1.00in]{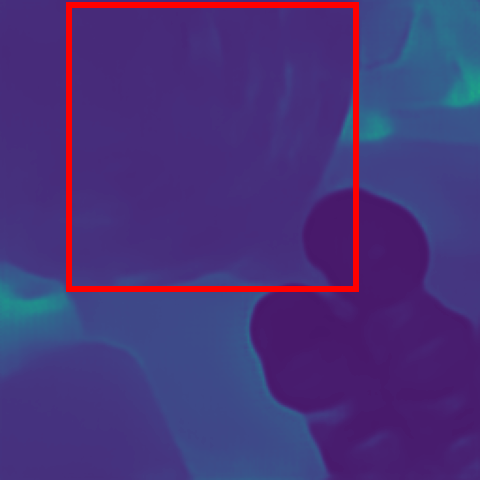} &
 \hspace{-0.5mm}\includegraphics[height =1.00in]{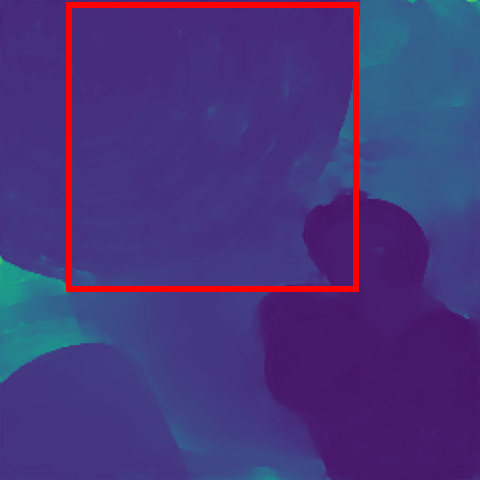}&
\hspace{-0.5mm}\includegraphics[height =1.00in]{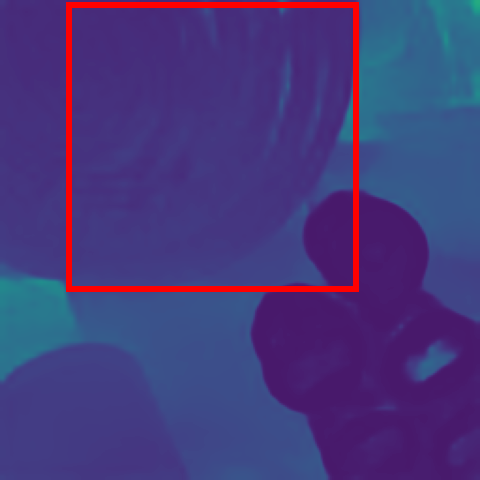} &
\hspace{-0.5mm}\includegraphics[height =1.00in]{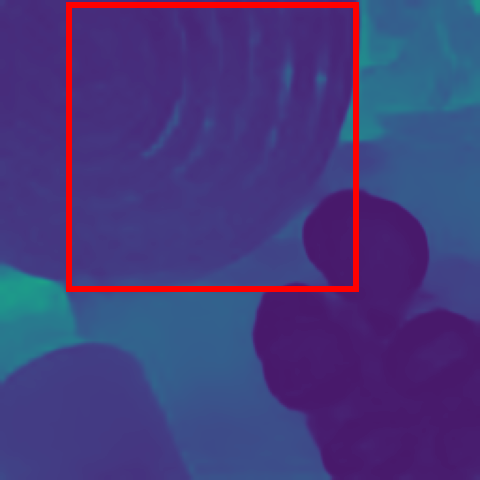} &
\hspace{-0.5mm}\includegraphics[height =1.00in]{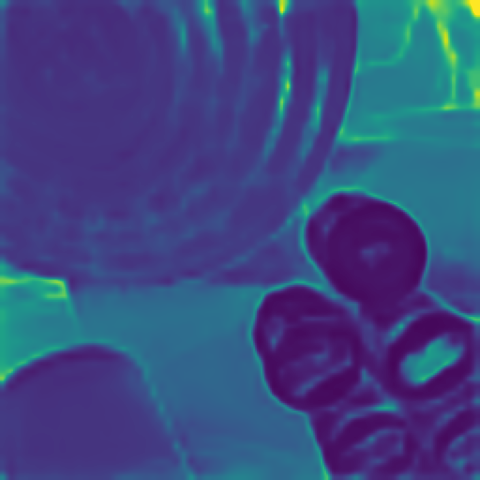}\\
\hspace{-0.5mm}\includegraphics[height= 1.00in]{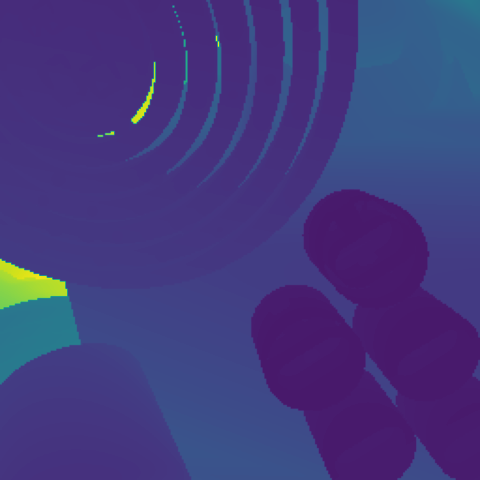} &
% \hspace{-0.5mm}\includegraphics[height= 0.8in]{figures/unknown_res/telephone_pred_viz.png} &
\hspace{-0.5mm}\includegraphics[height= 1.00in]{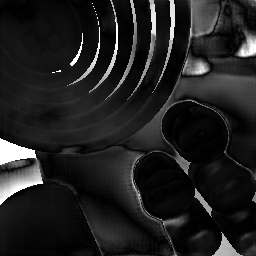} &
\hspace{-0.5mm}\includegraphics[height= 1.00in]{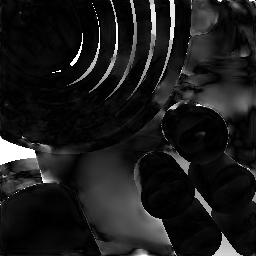} &
\hspace{-0.5mm}\includegraphics[height =1.00in]{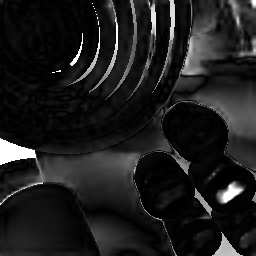} &
\hspace{-0.5mm}\includegraphics[height= 1.00in]{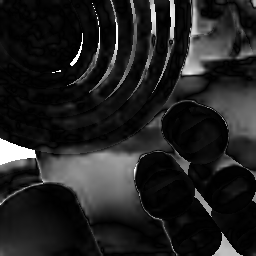} & 
\hspace{-0.5mm}\includegraphics[height =1.00in]{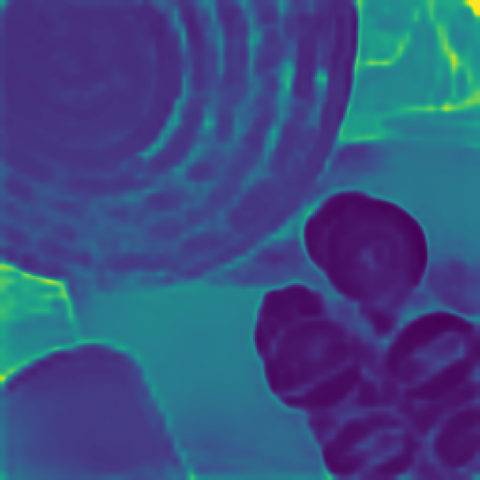}\\

\hspace{-0.5mm}\includegraphics[height =1.00in]{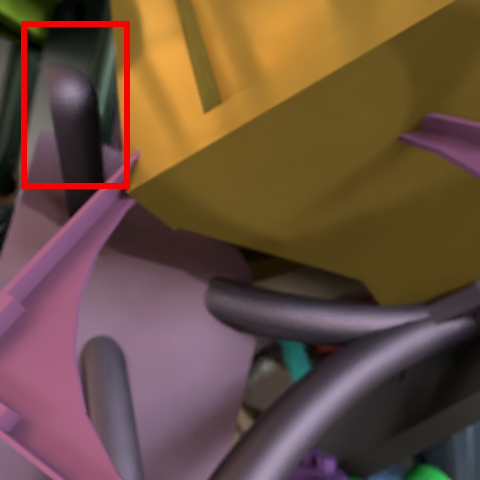} &
\hspace{-0.5mm}\includegraphics[height =1.00in]{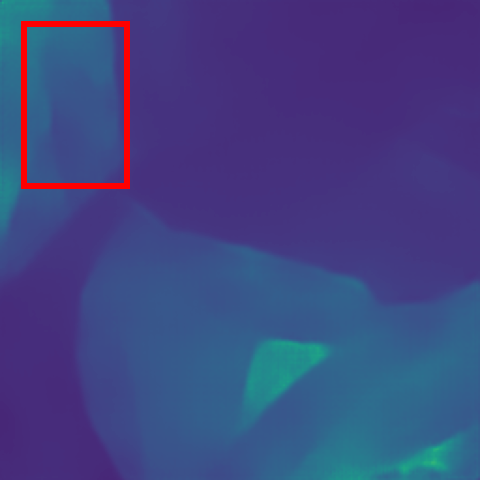} &
 \hspace{-0.5mm}\includegraphics[height =1.00in]{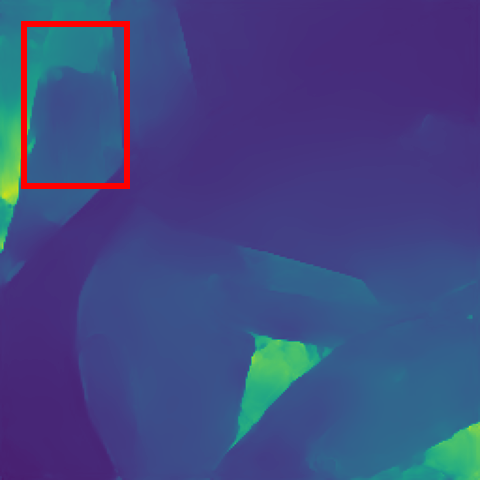}&
\hspace{-0.5mm}\includegraphics[height =1.00in]{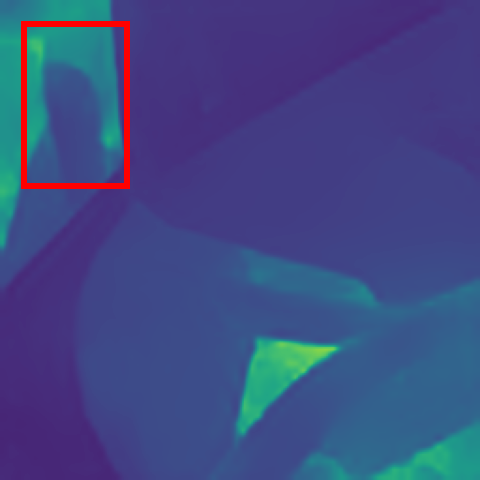} &
\hspace{-0.5mm}\includegraphics[height =1.00in]{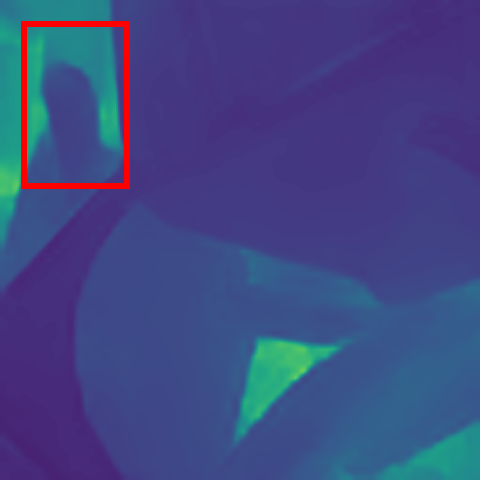} &
\hspace{-0.5mm}\includegraphics[height =1.00in]{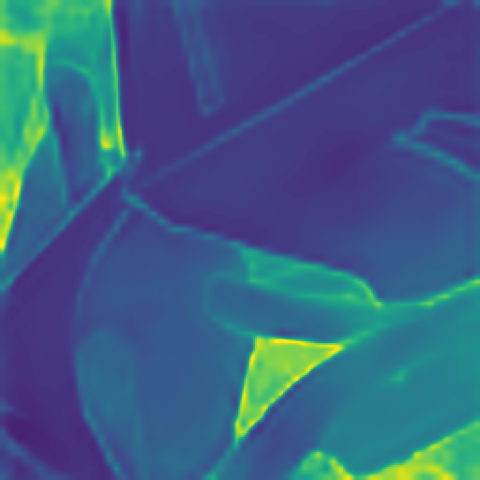}\\
\hspace{-0.5mm}\includegraphics[height= 1.00in]{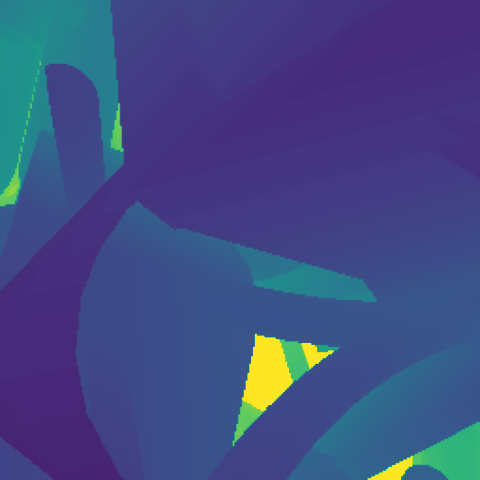} &
% \hspace{-0.5mm}\includegraphics[height= 0.8in]{figures/unknown_res/telephone_pred_viz.png} &
\hspace{-0.5mm}\includegraphics[height= 1.00in]{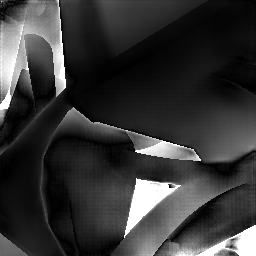} &
\hspace{-0.5mm}\includegraphics[height= 1.00in]{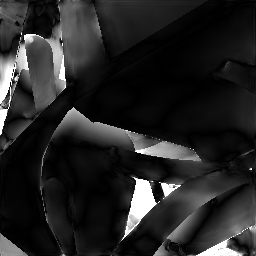} &
\hspace{-0.5mm}\includegraphics[height =1.00in]{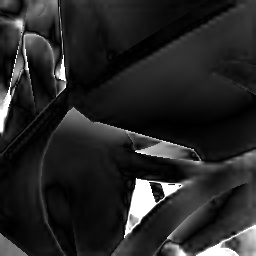} &
\hspace{-0.5mm}\includegraphics[height= 1.00in]{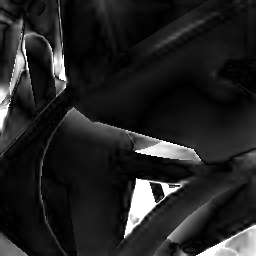} & 
\hspace{-0.5mm}\includegraphics[height =1.00in]{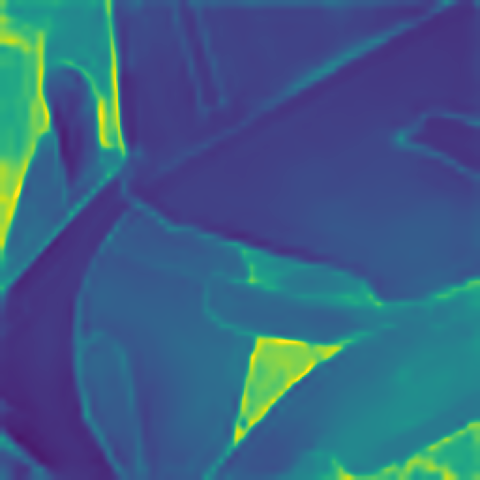}\\

\hspace{-0.5mm}\includegraphics[height =1.00in]{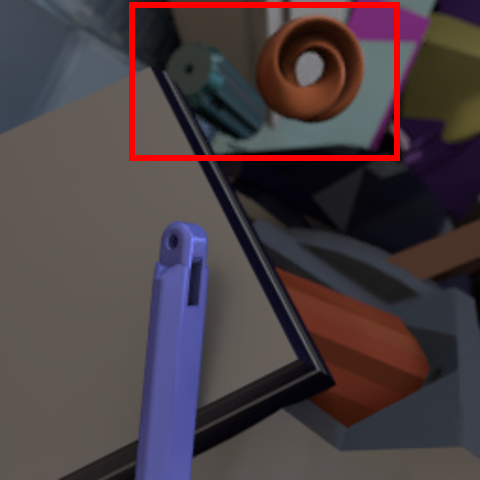} &
\hspace{-0.5mm}\includegraphics[height =1.00in]{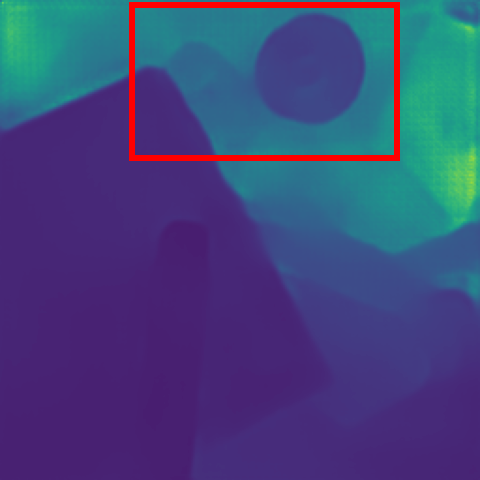} &
 \hspace{-0.5mm}\includegraphics[height =1.00in]{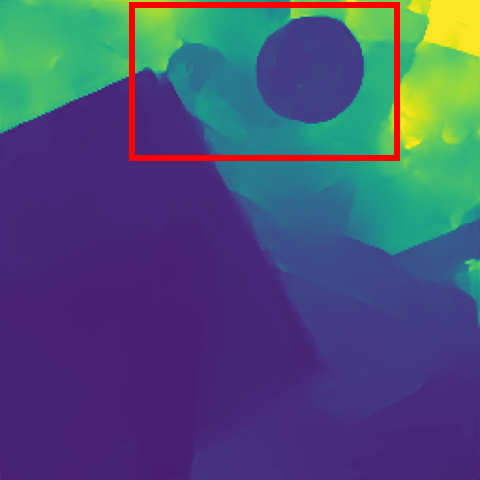}&
\hspace{-0.5mm}\includegraphics[height =1.00in]{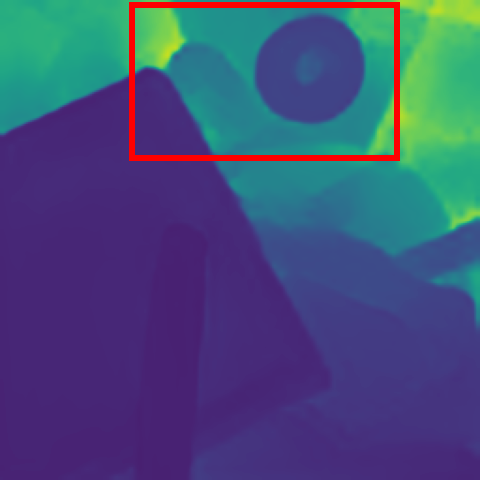} &
\hspace{-0.5mm}\includegraphics[height =1.00in]{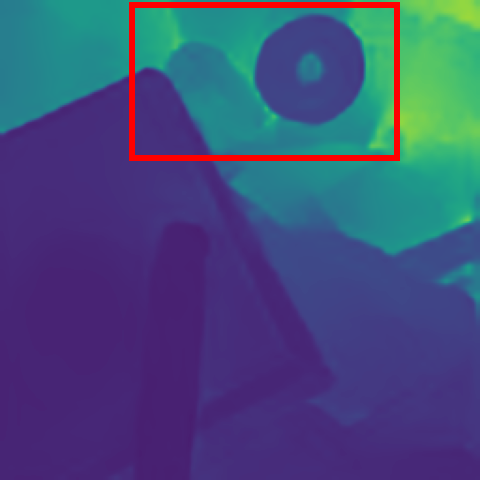} &
\hspace{-0.5mm}\includegraphics[height =1.00in]{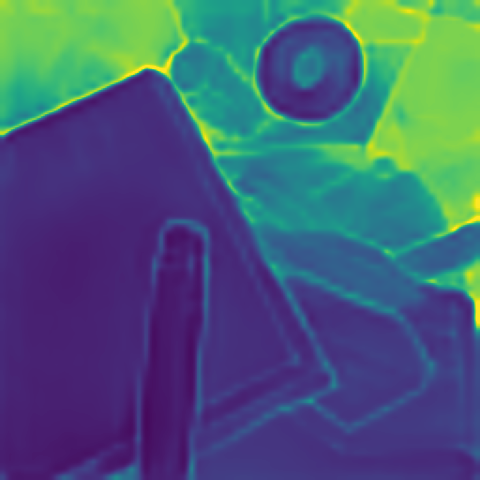}\\
\hspace{-0.5mm}\includegraphics[height= 1.00in]{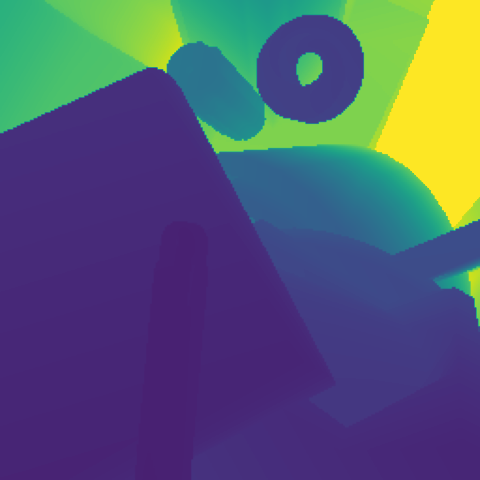} &
% \hspace{-0.5mm}\includegraphics[height= 0.8in]{figures/unknown_res/telephone_pred_viz.png} &
\hspace{-0.5mm}\includegraphics[height= 1.00in]{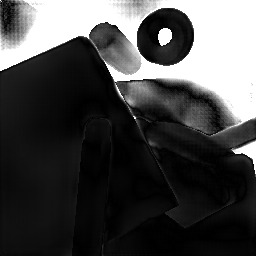} &
\hspace{-0.5mm}\includegraphics[height= 1.00in]{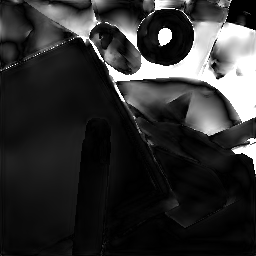} &
\hspace{-0.5mm}\includegraphics[height =1.00in]{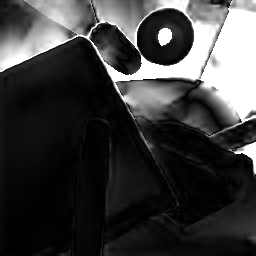} &
\hspace{-0.5mm}\includegraphics[height= 1.00in]{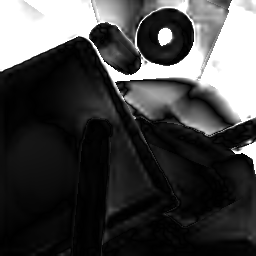} & 
\hspace{-0.5mm}\includegraphics[height =1.00in]{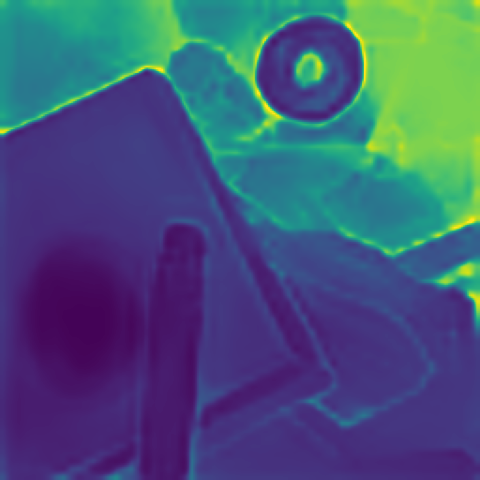}\\

\end{tabular}
\caption{Qualitative results on FoD500 test set. The first column shows the first image in the input focal stack and the corresponding ground truth. The next 4 columns show the depth predictions (rows 1, 3, 5, and 7) and the corresponding error map (rows 2, 4, 6, and 8). The last column presents the corresponding uncertainty maps of Ours-FV (rows 1, 3, 5, and 7) and Ours-DFV (rows 2, 4, 6, and 8). The warmer the color, the higher the value.} % For error map, the darker the color, the lower the value.}
\label{fig:Fod500}
\end{figure*}

\begin{figure*}
    \centering
    \begin{tabular}{ccccc|c}
%\hspace{-2mm}\includegraphics[height =0.41in]{figures/scheme1.pdf} &
\hspace{-2.5mm}Image/GT &  \hspace{-2.5mm}DDFF & \hspace{-2.5mm} DefocusNet &\hspace{-2.5mm}  Ours-FV &\hspace{-2.5mm}  Ours-DFV &\hspace{-2.0mm}  {\small Uncer. Ours-FV/DFV} \\

\hspace{-2.5mm}\includegraphics[height=0.75in]{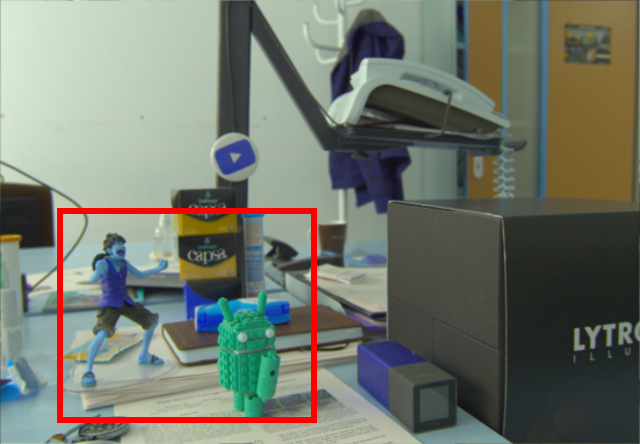} &
\hspace{-2.5mm}\includegraphics[height=0.75in]{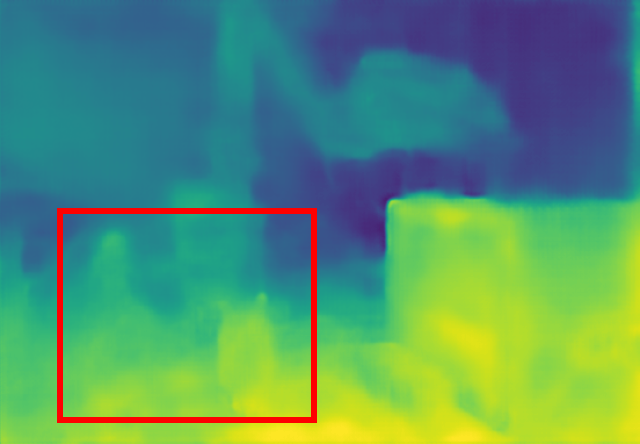} &
 \hspace{-2.5mm}\includegraphics[height=0.75in]{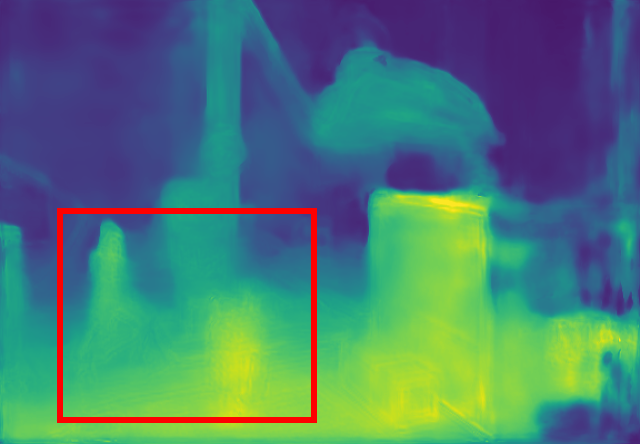}&
\hspace{-2.5mm}\includegraphics[height =0.75in]{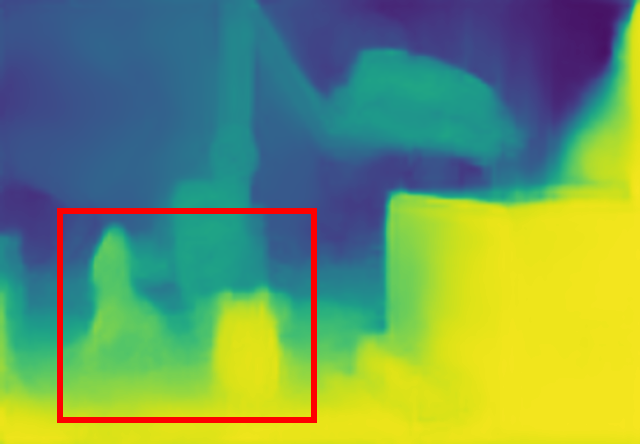} &
\hspace{-2.5mm}\includegraphics[height =0.75in]{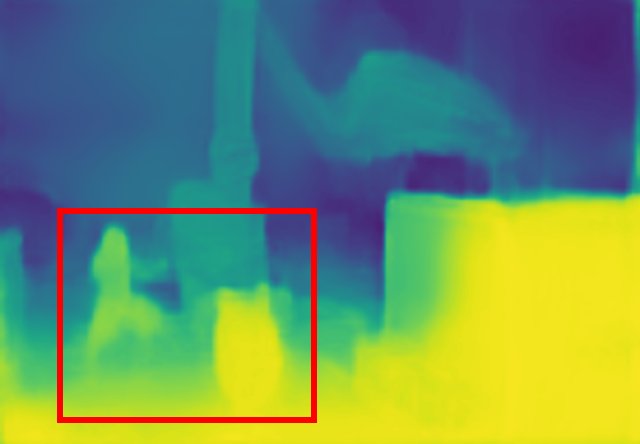} & 
\hspace{-1.2mm}\includegraphics[height =0.75in]{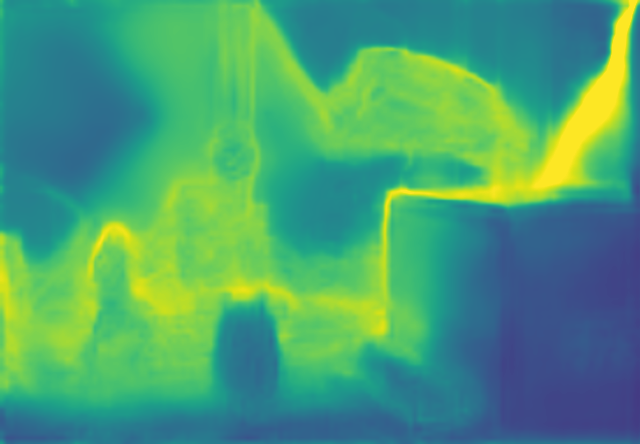}\\

\hspace{-2.5mm}\includegraphics[height= 0.75in]{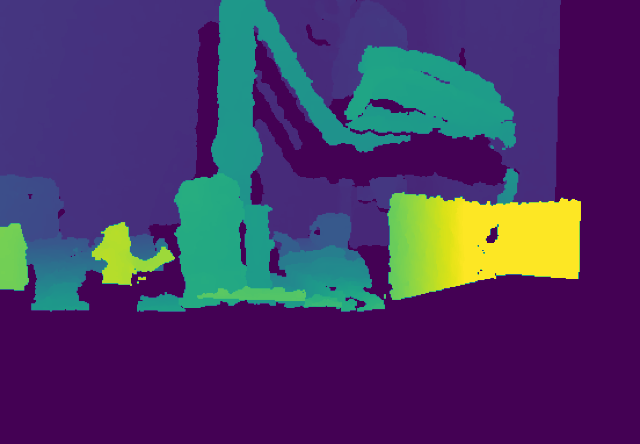} &
% \hspace{-2.5mm}\includegraphics[height= 0.8in]{figures/unknown_res/telephone_pred_viz.png} &
\hspace{-2.5mm}\includegraphics[height= 0.75in]{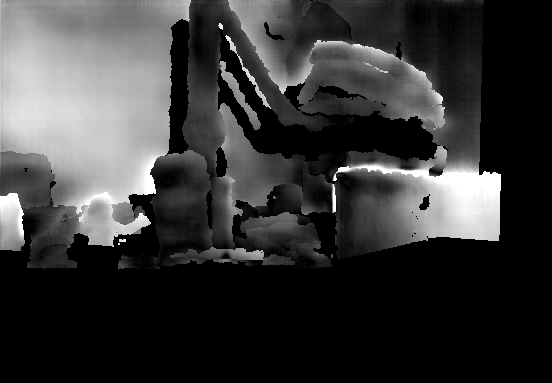} &
\hspace{-2.5mm}\includegraphics[height= 0.75in]{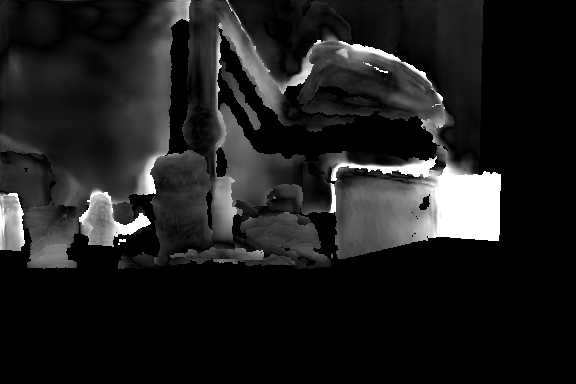} &
\hspace{-2.5mm}\includegraphics[height =0.75in]{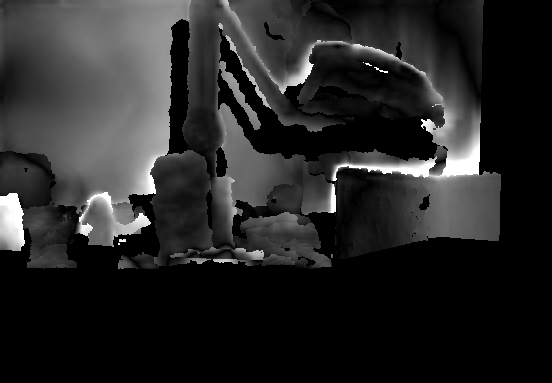} &
\hspace{-2.5mm}\includegraphics[height= 0.75in]{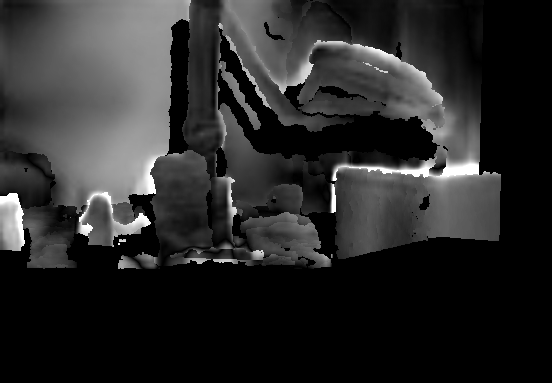} & 
\hspace{-1.2mm}\includegraphics[height =0.75in]{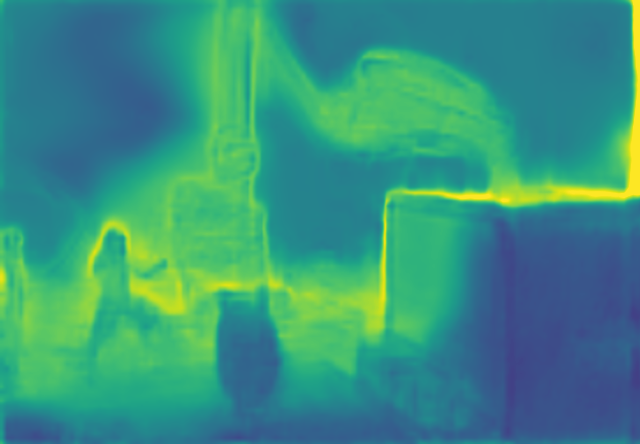}\\

\hspace{-2.5mm}\includegraphics[height =0.75in]{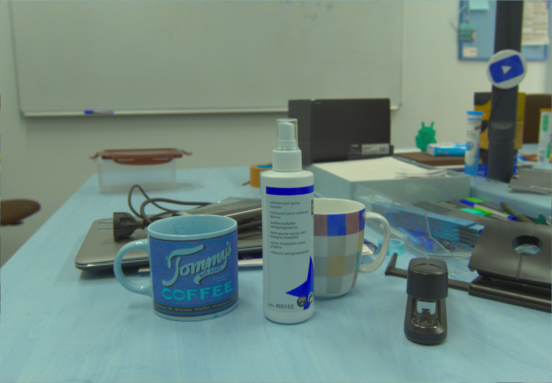} &
\hspace{-2.5mm}\includegraphics[height=0.75in]{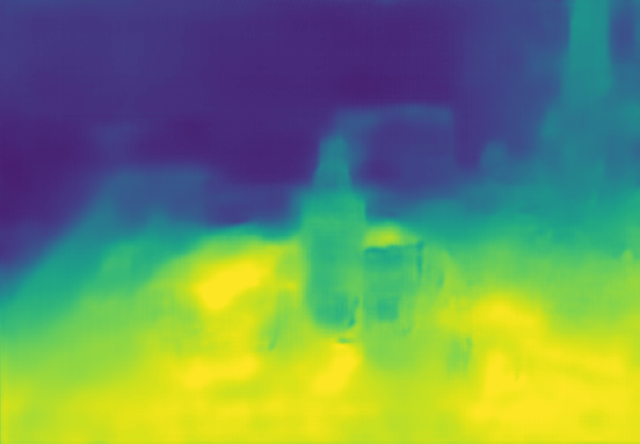} &
 \hspace{-2.5mm}\includegraphics[height=0.75in]{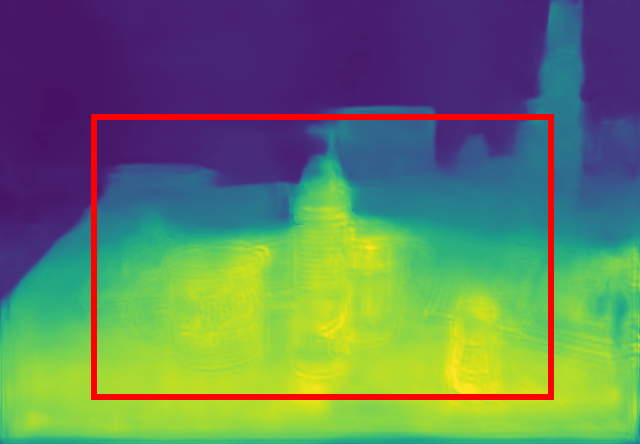}&
\hspace{-2.5mm}\includegraphics[height =0.75in]{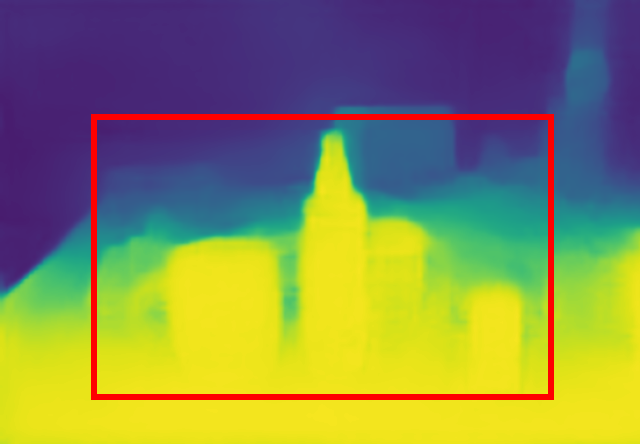} &
\hspace{-2.5mm}\includegraphics[height =0.75in]{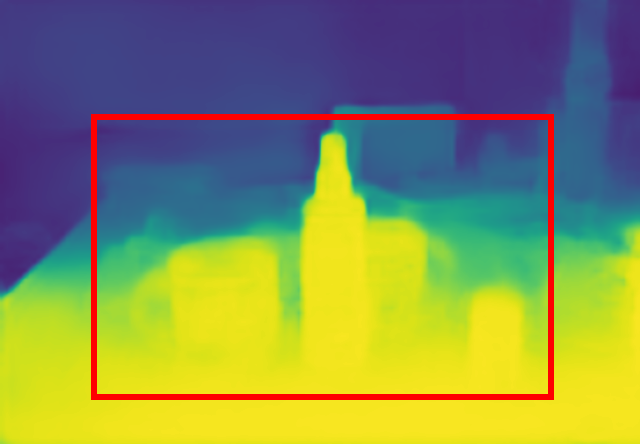} & 
\hspace{-1.2mm}\includegraphics[height =0.75in]{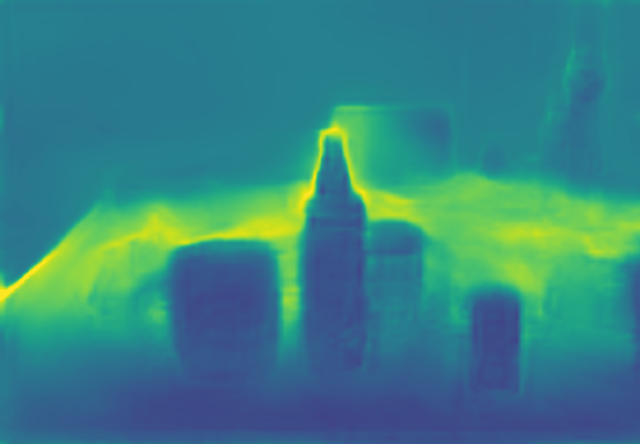}\\

\hspace{-2.5mm}\includegraphics[height= 0.75in]{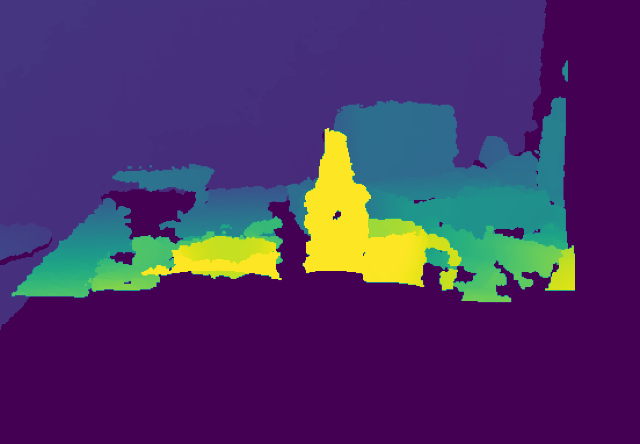} &
% \hspace{-2.5mm}\includegraphics[height= 0.8in]{figures/unknown_res/telephone_pred_viz.png} &
\hspace{-2.5mm}\includegraphics[height= 0.75in]{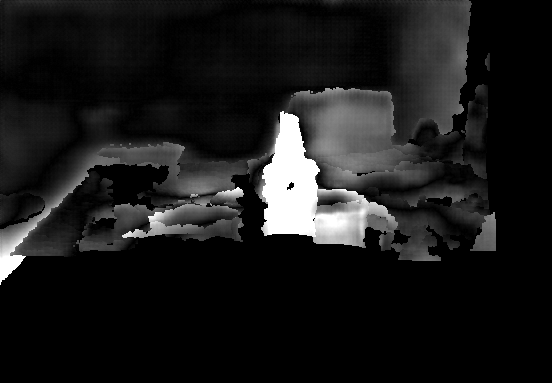} &
\hspace{-2.5mm}\includegraphics[height= 0.75in]{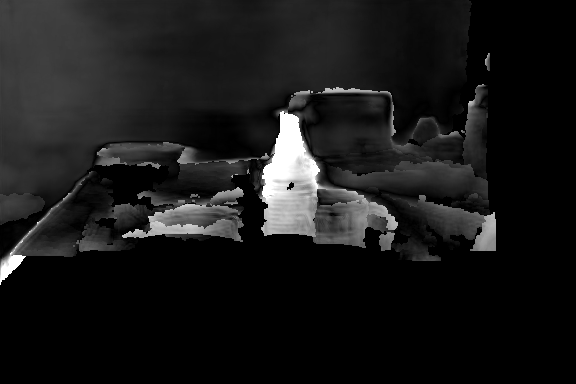} &
\hspace{-2.5mm}\includegraphics[height =0.75in]{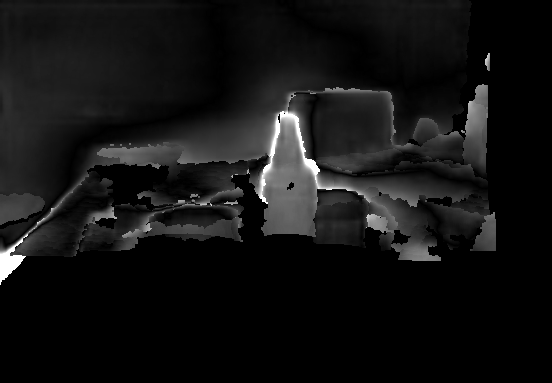} &
\hspace{-2.5mm}\includegraphics[height= 0.75in]{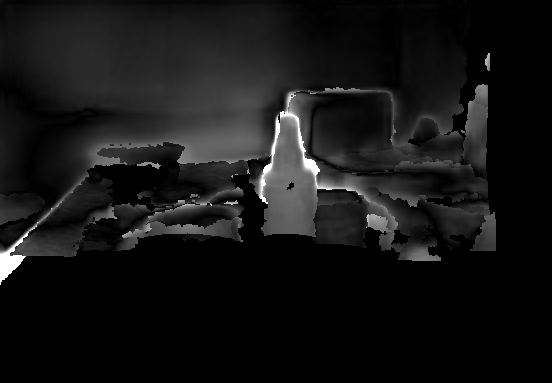} & 
\hspace{-1.2mm}\includegraphics[height =0.75in]{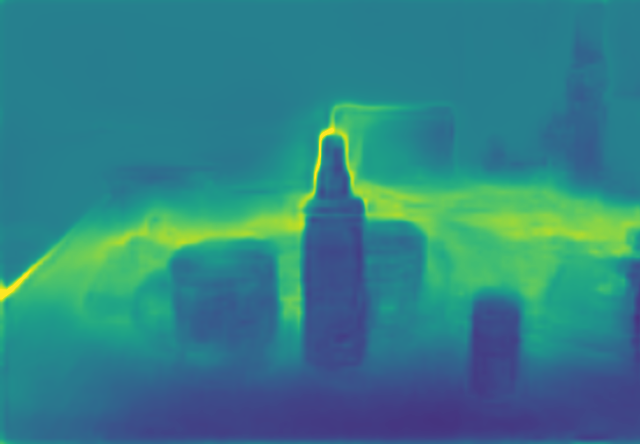}\\

\hspace{-2.5mm}\includegraphics[height =0.75in]{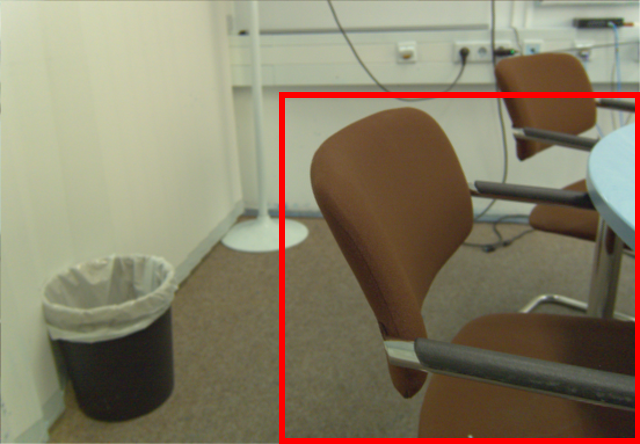} &
\hspace{-2.5mm}\includegraphics[height=0.75in]{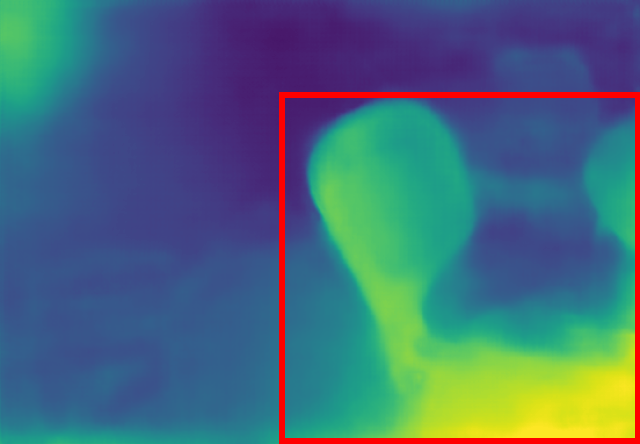} &
 \hspace{-2.5mm}\includegraphics[height=0.75in]{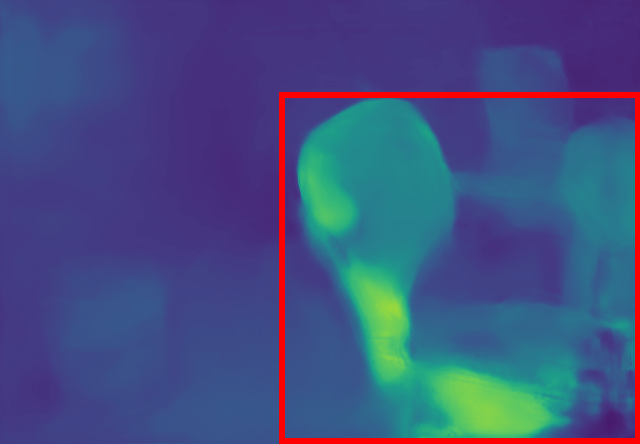}&
\hspace{-2.5mm}\includegraphics[height =0.75in]{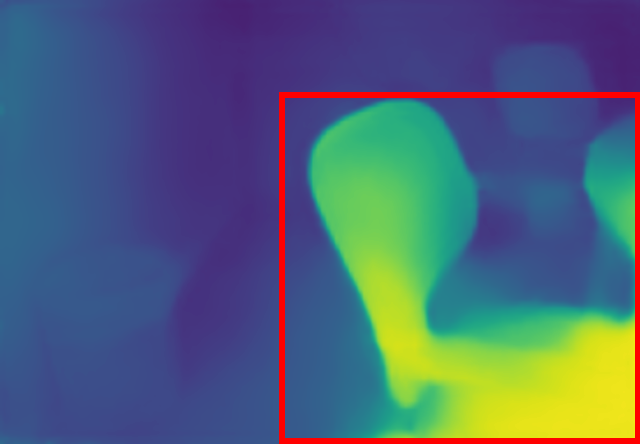} &
\hspace{-2.5mm}\includegraphics[height =0.75in]{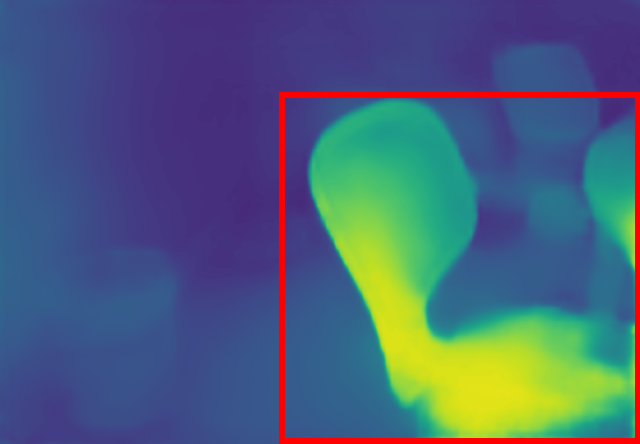} & 
\hspace{-1.2mm}\includegraphics[height =0.75in]{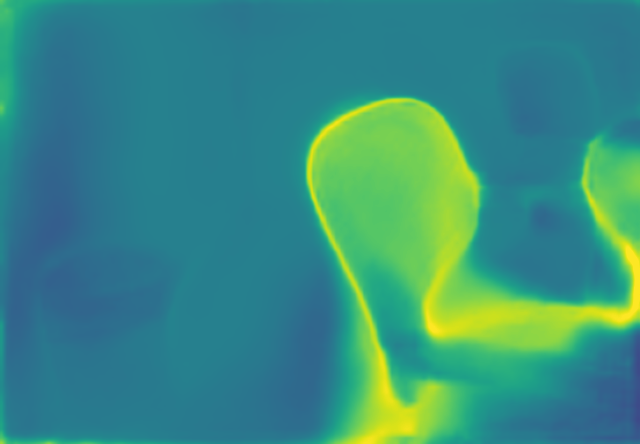}\\

\hspace{-2.5mm}\includegraphics[height= 0.75in]{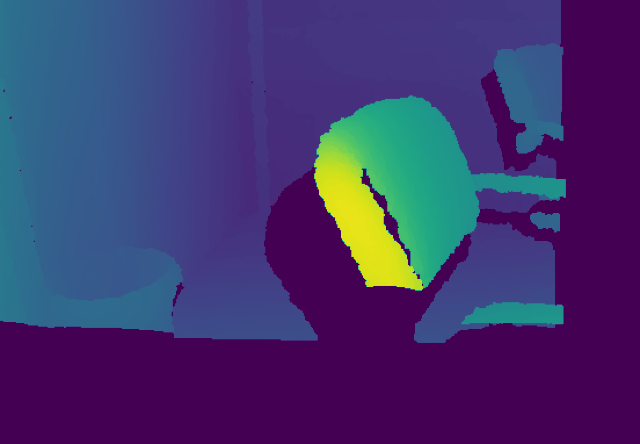} &
% \hspace{-2.5mm}\includegraphics[height= 0.8in]{figures/unknown_res/telephone_pred_viz.png} &
\hspace{-2.5mm}\includegraphics[height= 0.75in]{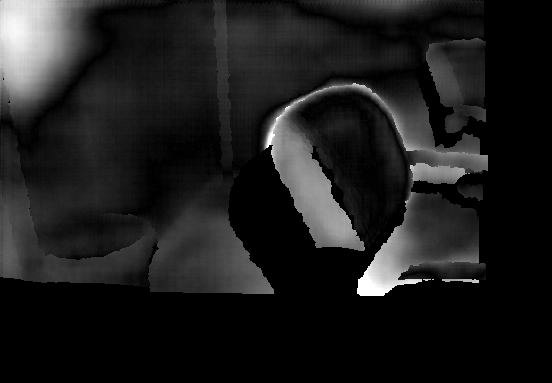} &
\hspace{-2.5mm}\includegraphics[height= 0.75in]{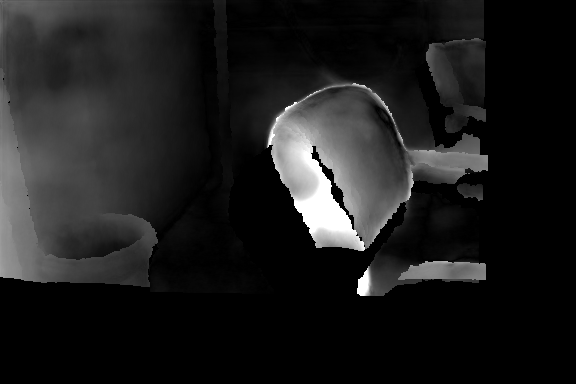} &
\hspace{-2.5mm}\includegraphics[height =0.75in]{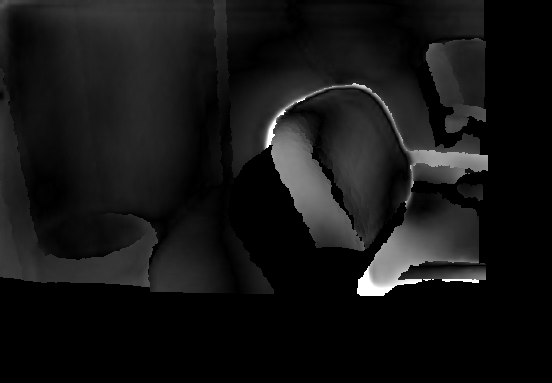} &
\hspace{-2.5mm}\includegraphics[height= 0.75in]{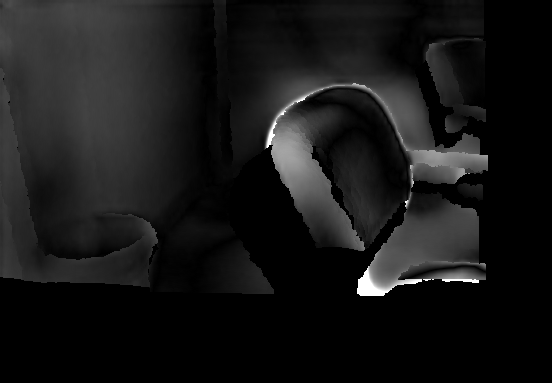} & 
\hspace{-1.2mm}\includegraphics[height =0.75in]{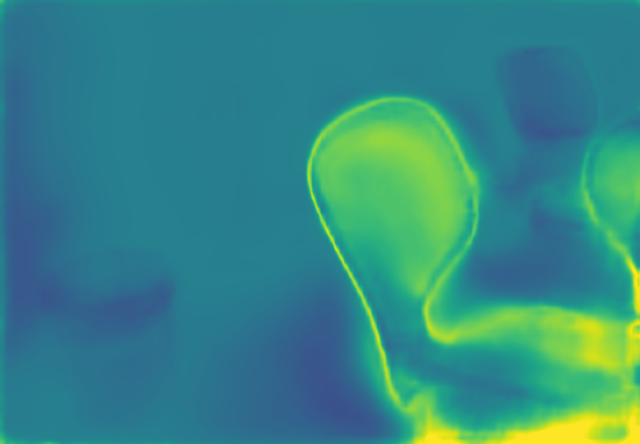}\\

\hspace{-2.5mm}\includegraphics[height=0.75in]{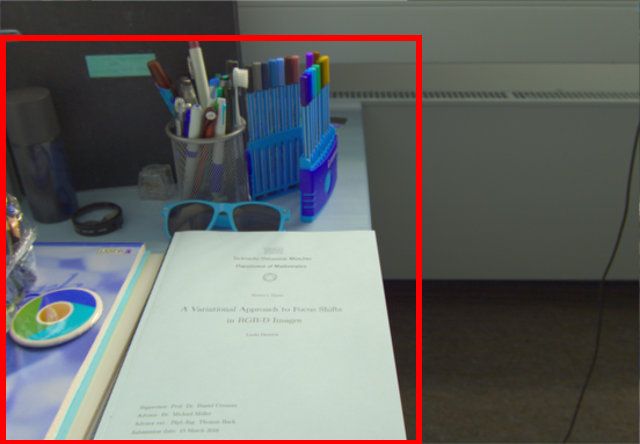} &
\hspace{-2.5mm}\includegraphics[height=0.75in]{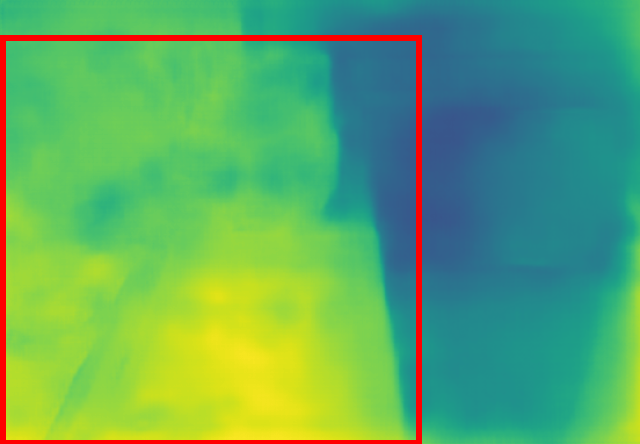} &
 \hspace{-2.5mm}\includegraphics[height=0.75in]{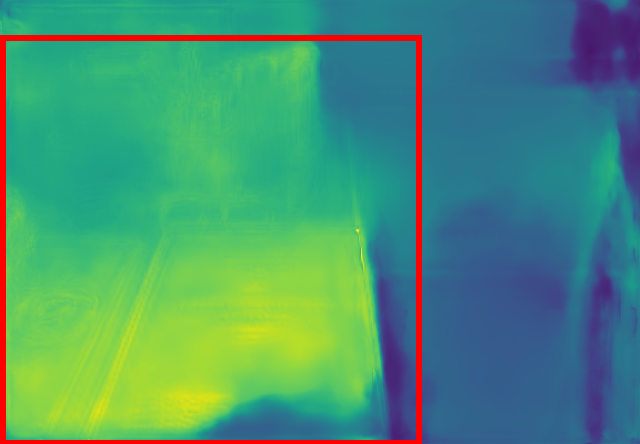}&
\hspace{-2.5mm}\includegraphics[height =0.75in]{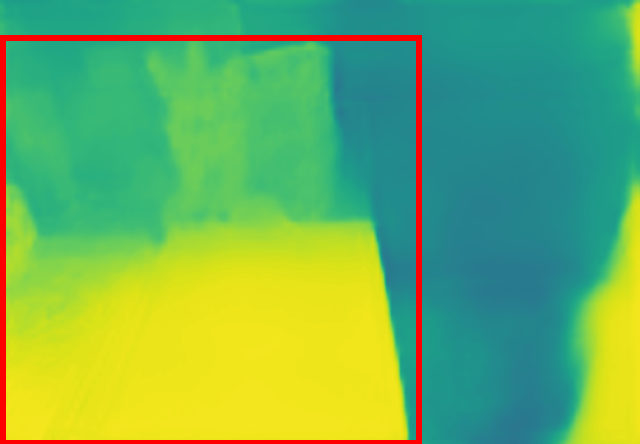} &
\hspace{-2.5mm}\includegraphics[height =0.75in]{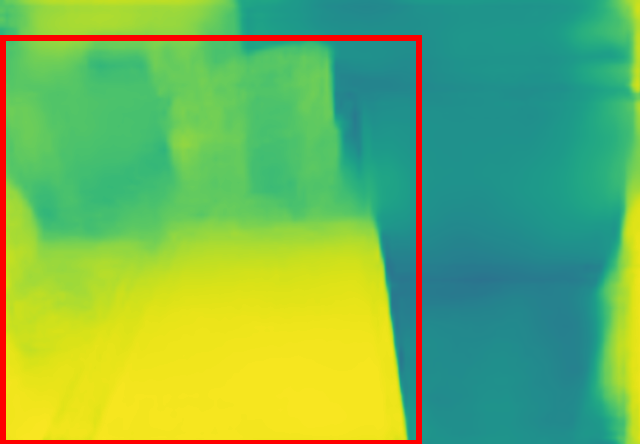} & 
\hspace{-1.2mm}\includegraphics[height =0.75in]{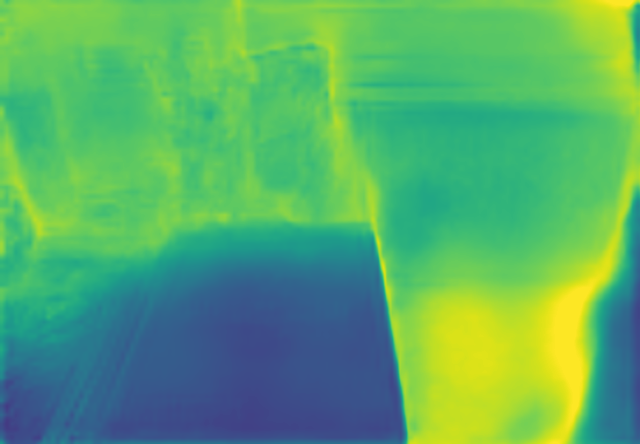}\\

\hspace{-2.5mm}\includegraphics[height= 0.75in]{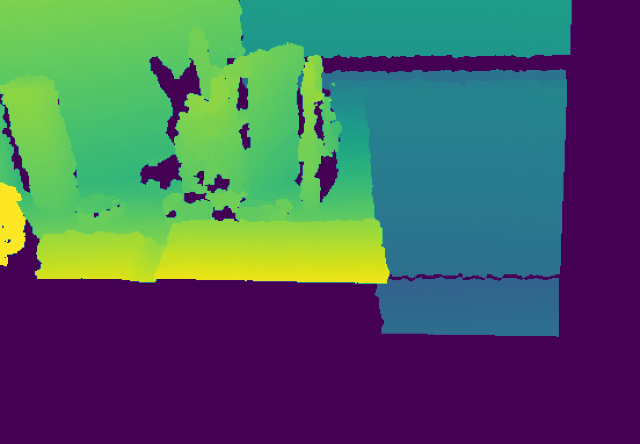} &
% \hspace{-2.5mm}\includegraphics[height= 0.8in]{figures/unknown_res/telephone_pred_viz.png} &
\hspace{-2.5mm}\includegraphics[height= 0.75in]{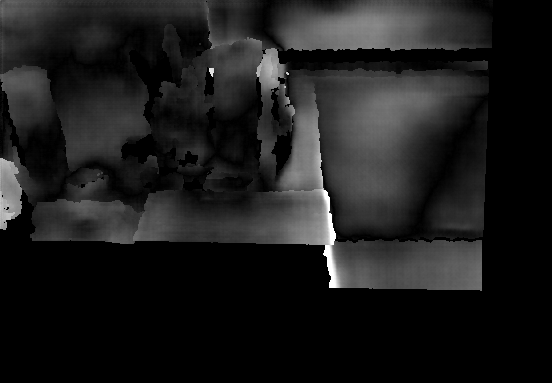} &
\hspace{-2.5mm}\includegraphics[height= 0.75in]{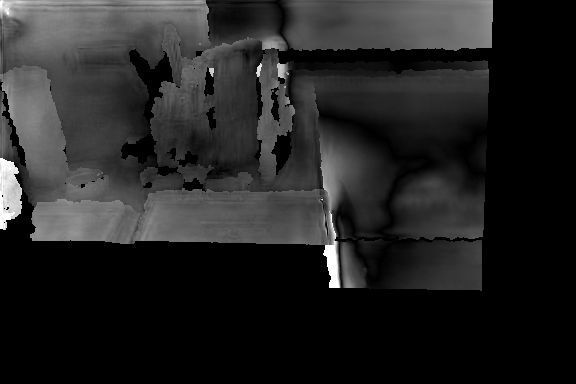} &
\hspace{-2.5mm}\includegraphics[height =0.75in]{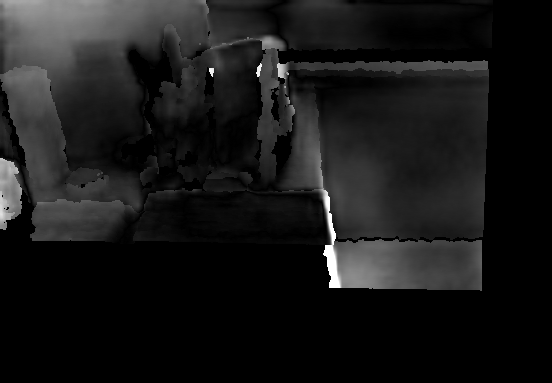} &
\hspace{-2.5mm}\includegraphics[height= 0.75in]{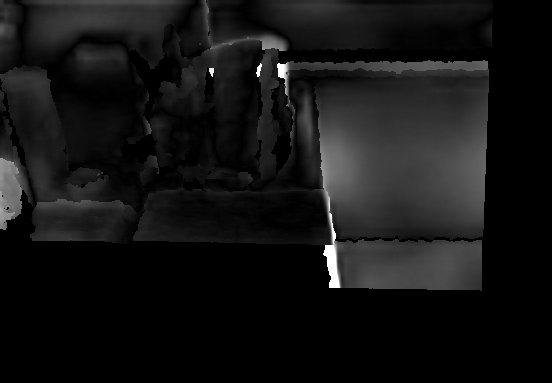} & 
\hspace{-1.2mm}\includegraphics[height =0.75in]{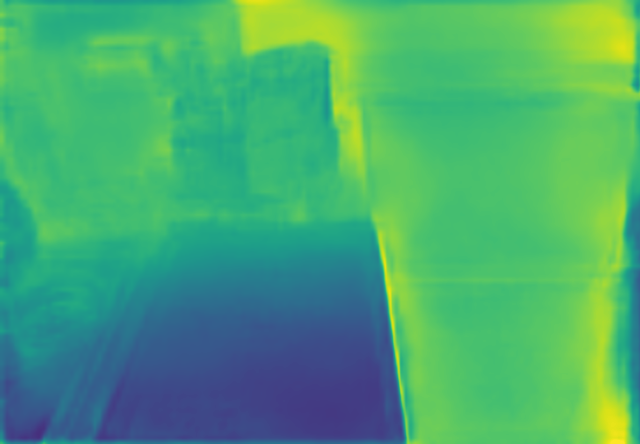}\\

    \hspace{-2.5mm}\includegraphics[height =0.75in]{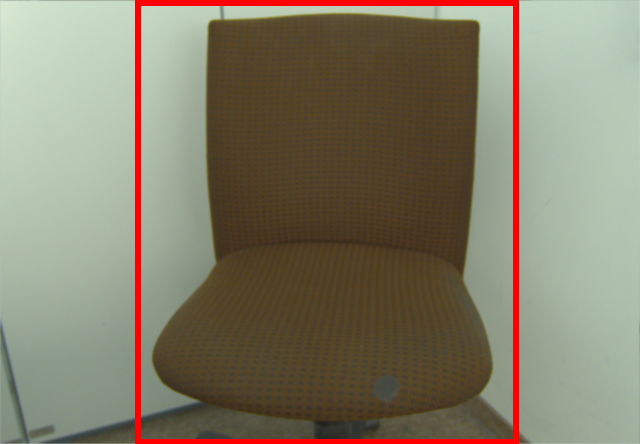} &
\hspace{-2.5mm}\includegraphics[height=0.75in]{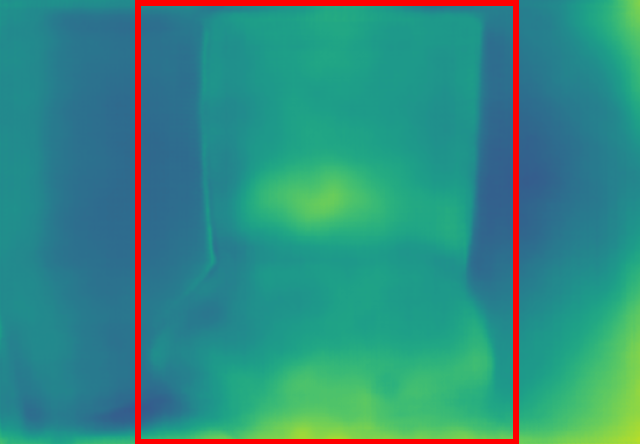} &
 \hspace{-2.5mm}\includegraphics[height=0.75in]{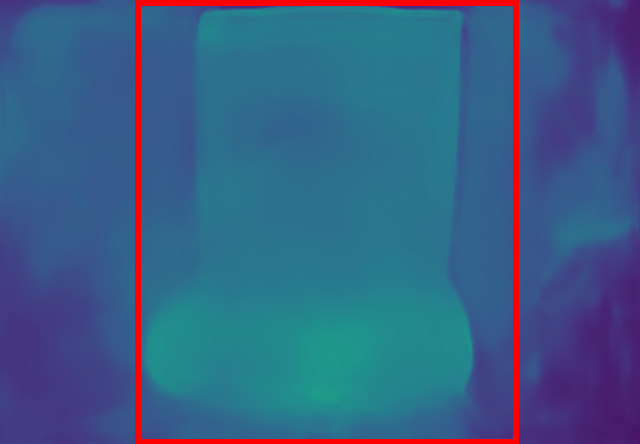}&
\hspace{-2.5mm}\includegraphics[height =0.75in]{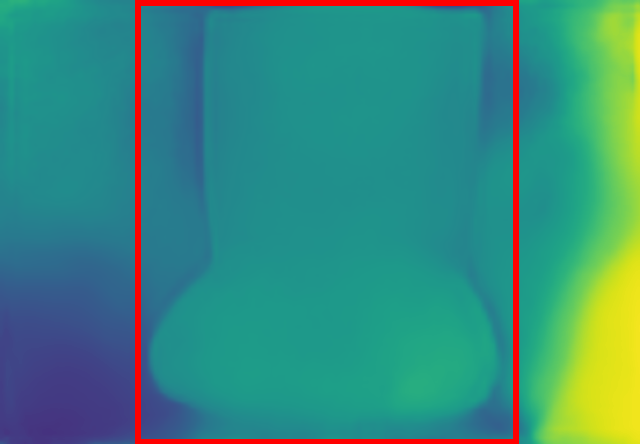} &
\hspace{-2.5mm}\includegraphics[height =0.75in]{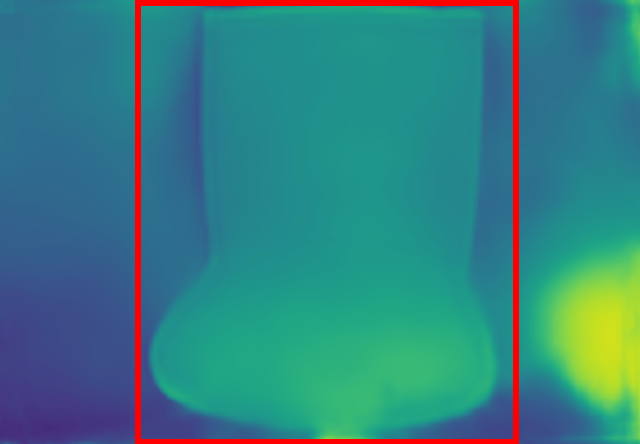} & 
\hspace{-1.2mm}\includegraphics[height =0.75in]{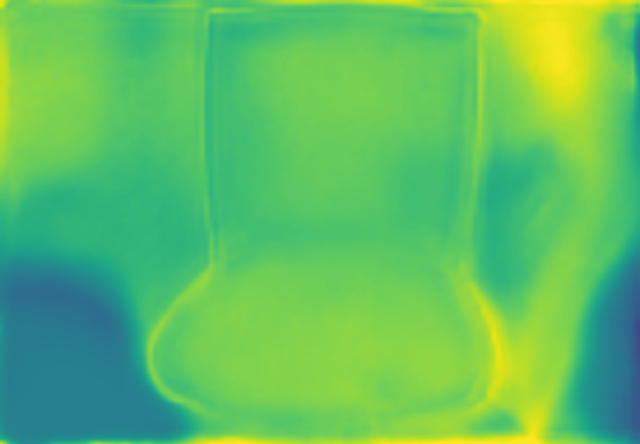}\\

\hspace{-2.5mm}\includegraphics[height= 0.75in]{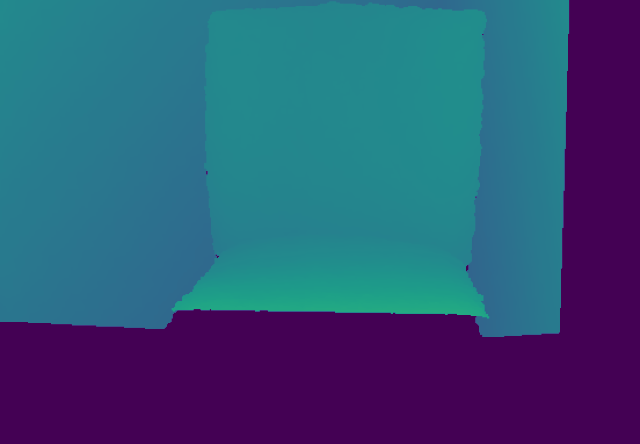} &
% \hspace{-2.5mm}\includegraphics[height= 0.8in]{figures/unknown_res/telephone_pred_viz.png} &
\hspace{-2.5mm}\includegraphics[height= 0.75in]{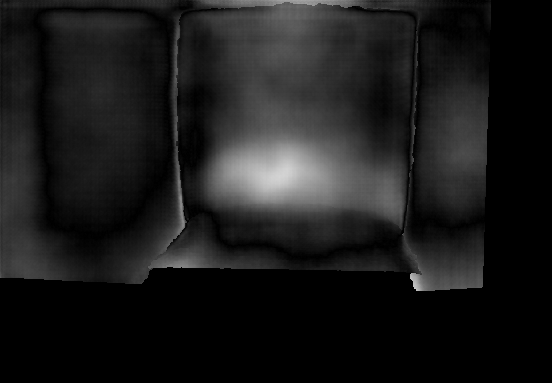} &
\hspace{-2.5mm}\includegraphics[height= 0.75in]{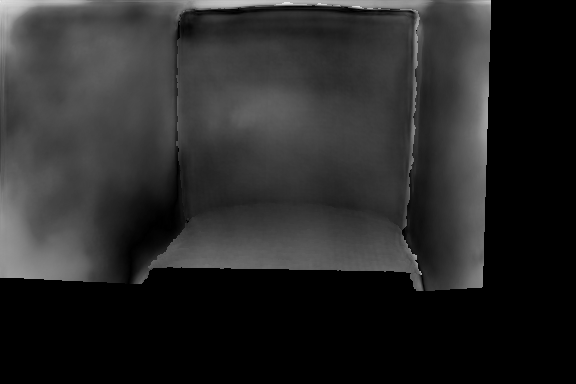} &
\hspace{-2.5mm}\includegraphics[height =0.75in]{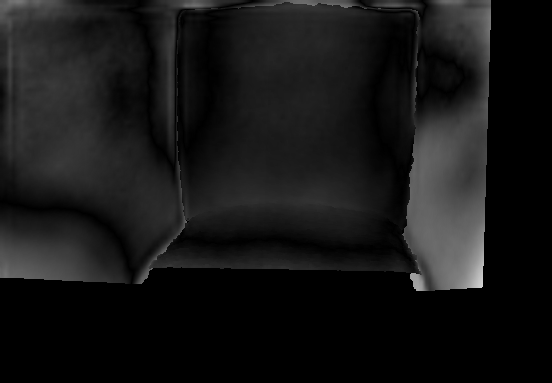} &
\hspace{-2.5mm}\includegraphics[height= 0.75in]{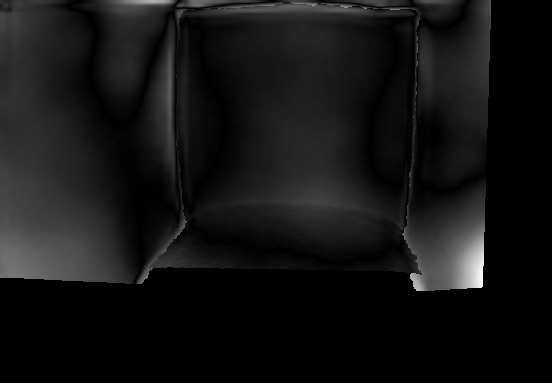} & 
\hspace{-1.2mm}\includegraphics[height =0.75in]{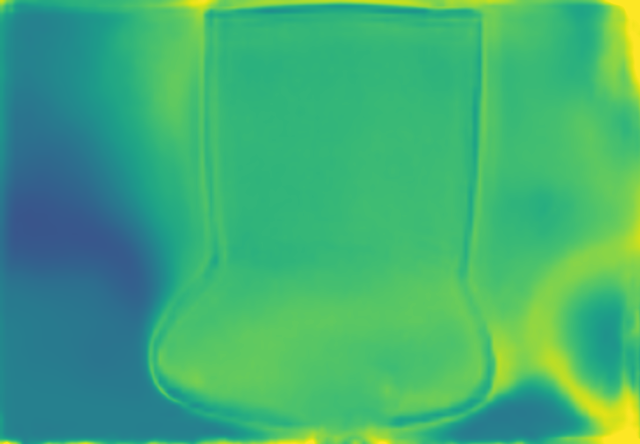}\\

 \end{tabular}
    \caption{Qualitative results on DDFF12 validation set. The first column shows the first image in the input focal stack and the corresponding ground truth. The next 4 columns show the disparity predictions (rows 1, 3, 5, and 7) and the corresponding error map  (rows 2, 4, 6, and 8). The last column presents the corresponding uncertainty maps of Ours-FV (rows 1, 3, 5, and 7) and Ours-DFV (rows 2, 4, 6, and 8). The warmer the color, the higher the value.} 
   
    \label{fig:ddff12}
\end{figure*}

Figure~\ref{fig:Fod500} and Figure~\ref{fig:ddff12} show additional qualitative results on FoD500 and DDFF-12 datasets. Compared to DDFF~\cite{hazirbas2018DFFdataset} and DefocusNet~\cite{maximov2020focusondefocus}, our methods, especially Ours-DFV, better preserve object boundaries and provide more smooth depth estimations. Some examples are highlighted with the red boxes. For uncertainty maps, the network turns to be more confident in the closer objects and less confident in objects that are farther or have weak textures. The high uncertainty is also frequently observed in  object boundaries. 

\begin{figure*}[t]
\centering

\begin{tabular}{ccccccc}
\hspace{-2.7mm}Image & \hspace{-2.7mm}MobileDFF &  \hspace{-2.5mm}DDFF & \hspace{-2.5mm} DefocusNet & \hspace{-2.5mm} AiFDepthNet  & \hspace{-2.5mm} Ours-FV  &\hspace{-2.5mm} Ours-DFV \\
\hspace{-2.7mm}\includegraphics[height =0.53in]{figures/horizontal_mobile/keyboard_img.png} &
\hspace{-2.7mm}\includegraphics[height= 0.50in]{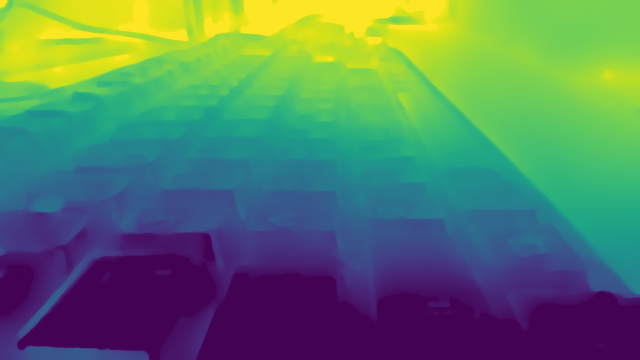} &
\hspace{-2.7mm}\includegraphics[height= 0.53in]{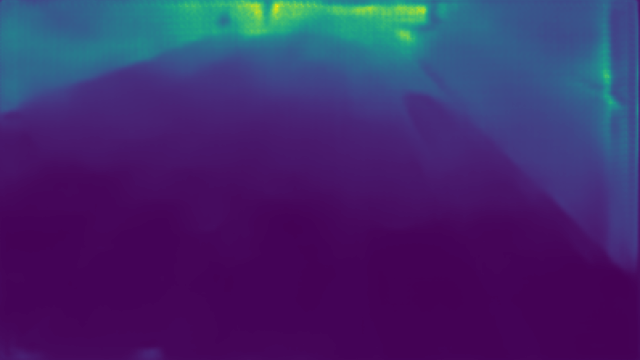}&
\hspace{-2.7mm}\includegraphics[height= 0.53in]{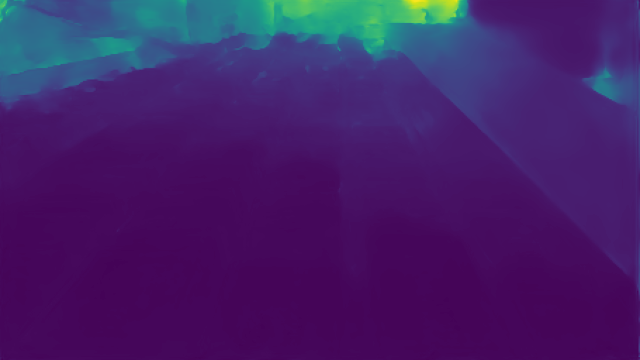}&
\hspace{-2.7mm}\includegraphics[height= 0.53in]{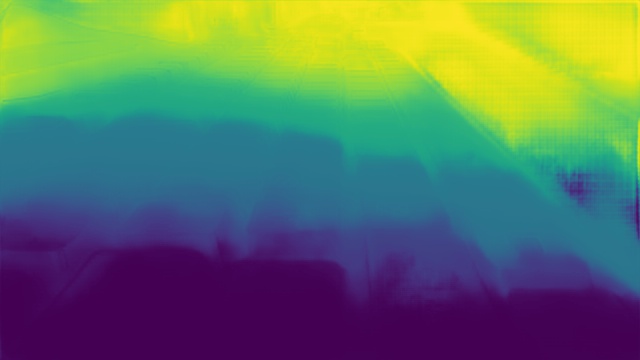}&
\hspace{-2.7mm}\includegraphics[height= 0.53in]{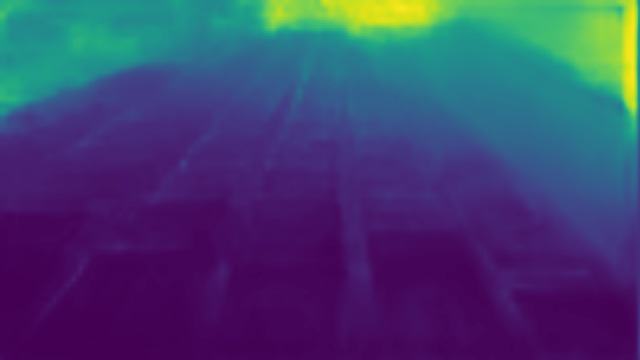}&
\hspace{-2.7mm}\includegraphics[height= 0.53in]{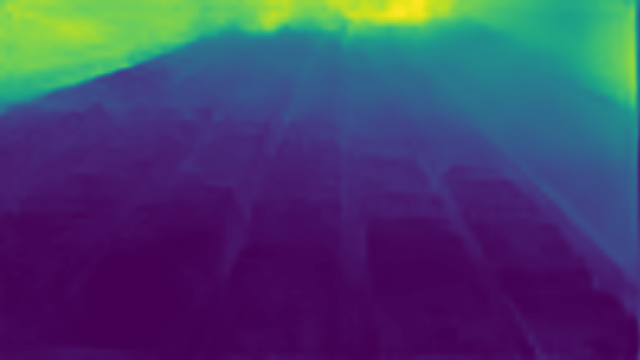}\\

\hspace{-2.7mm}\includegraphics[height =0.53in]{figures/horizontal_mobile/balls_img.png} &
\hspace{-2.7mm}\includegraphics[height= 0.53in]{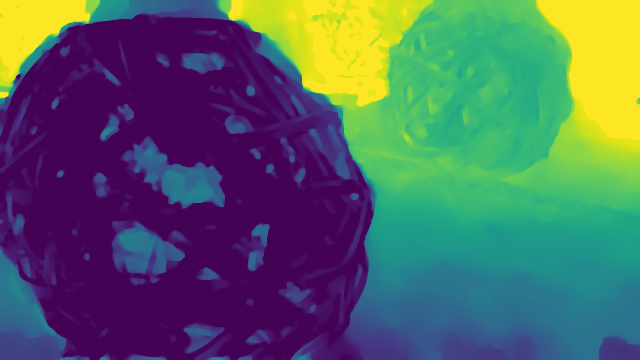} &
\hspace{-2.7mm}\includegraphics[height= 0.53in]{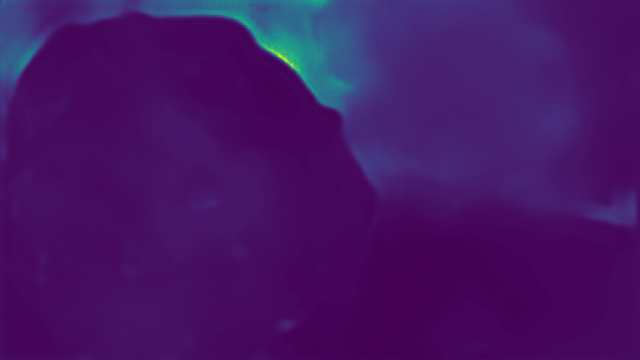}&
\hspace{-2.7mm}\includegraphics[height= 0.53in]{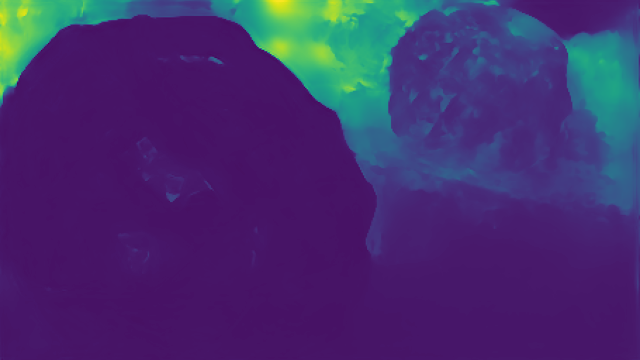}&
\hspace{-2.7mm}\includegraphics[height= 0.53in]{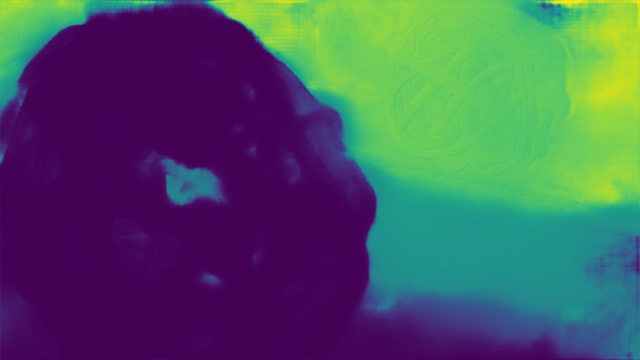}&
\hspace{-2.7mm}\includegraphics[height= 0.53in]{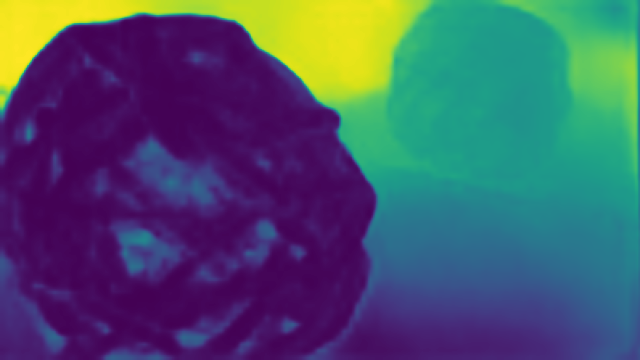}&
\hspace{-2.7mm}\includegraphics[height= 0.53in]{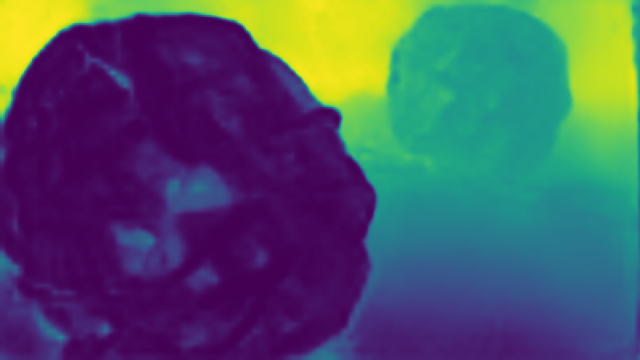}\\

\hspace{-2.7mm}\includegraphics[height =0.53in]{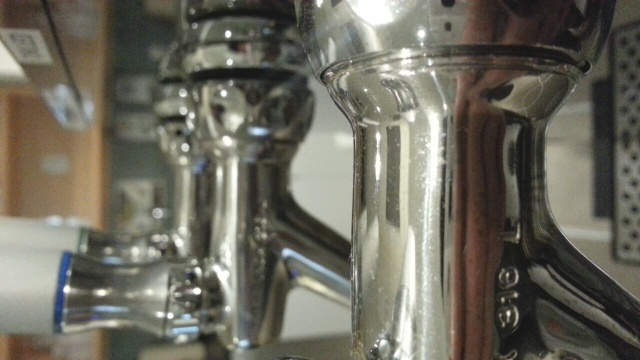} &
\hspace{-2.7mm}\includegraphics[height= 0.53in]{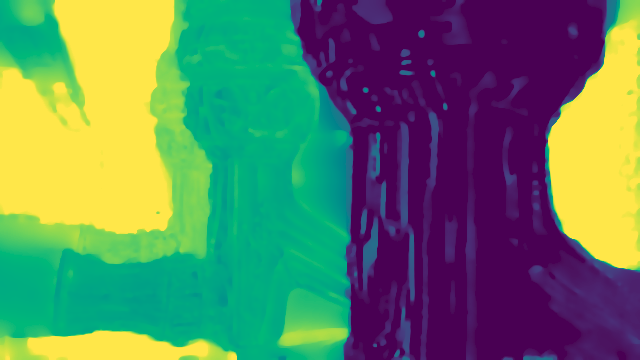} &
\hspace{-2.7mm}\includegraphics[height= 0.53in]{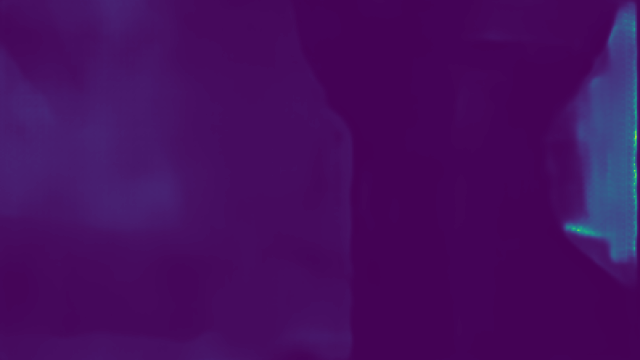}&
\hspace{-2.7mm}\includegraphics[height= 0.53in]{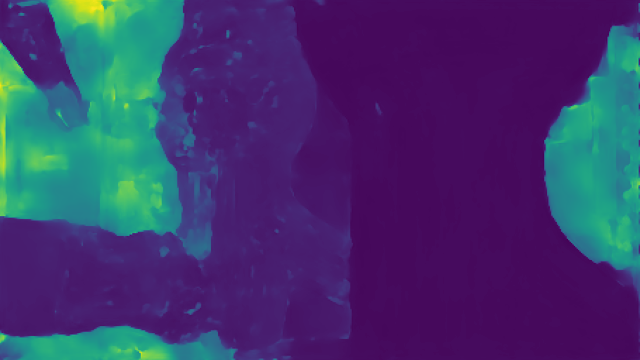}&
\hspace{-2.7mm}\includegraphics[height= 0.53in]{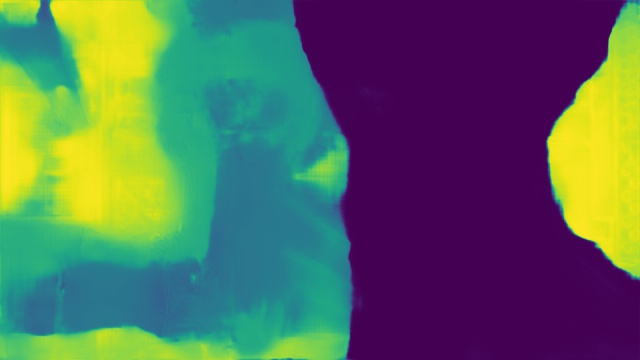}&
\hspace{-2.7mm}\includegraphics[height= 0.53in]{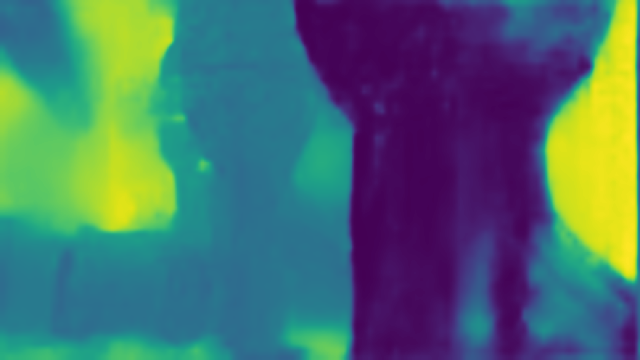}&
\hspace{-2.7mm}\includegraphics[height= 0.53in]{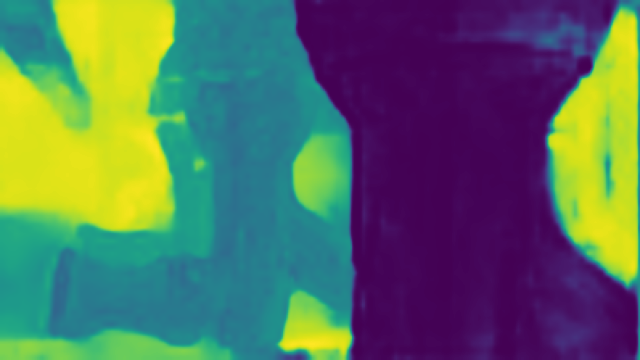}\\

\hspace{-2.5mm}\includegraphics[height =1.15in]{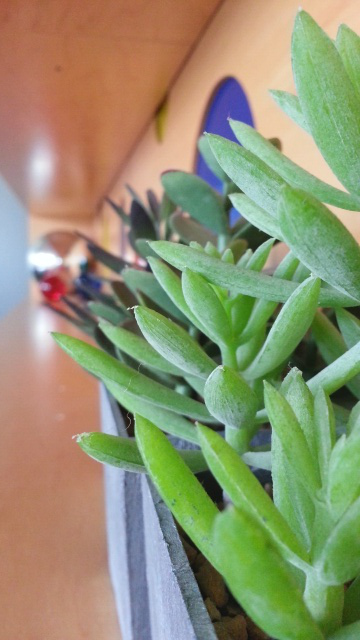} &
\hspace{-2.5mm}\includegraphics[height =1.15in]{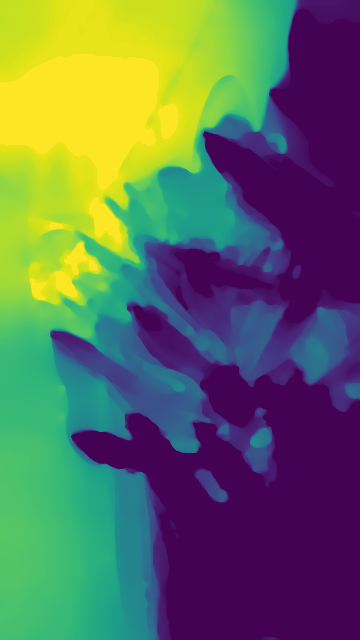} &
 \hspace{-2.5mm}\includegraphics[height =1.15in]{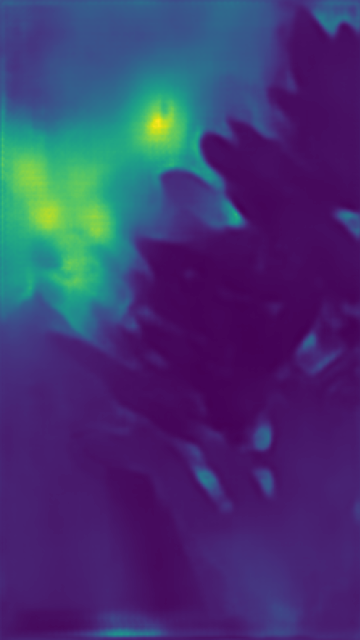}&
\hspace{-2.5mm}\includegraphics[height =1.15in]{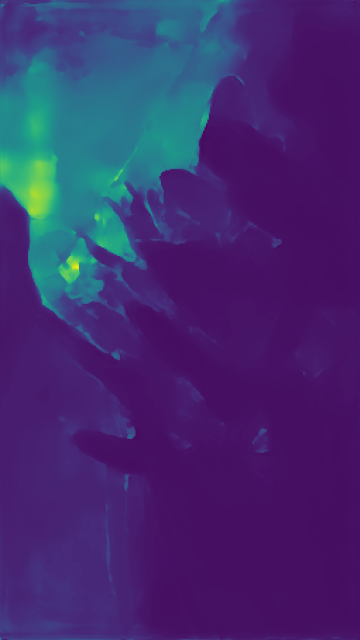} &
\hspace{-2.5mm}\includegraphics[height =1.15in]{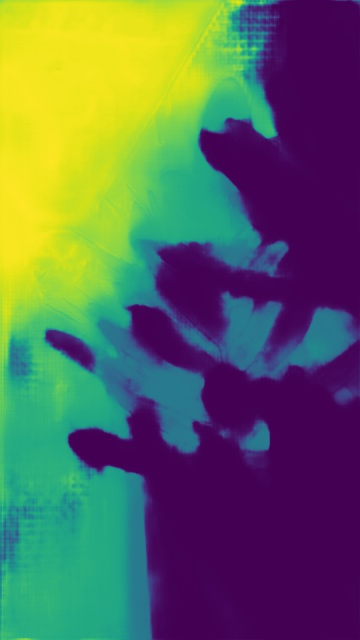}& 
\hspace{-2.5mm}\includegraphics[height =1.15in]{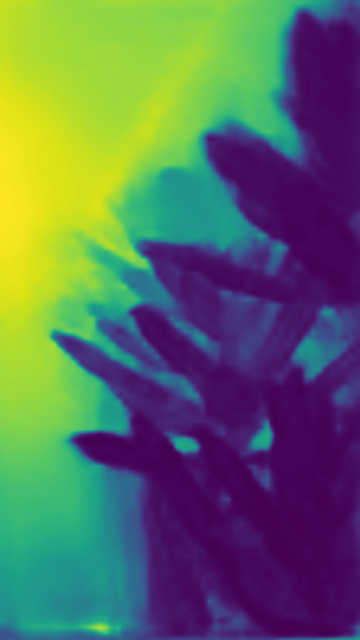}& 
\hspace{-2.5mm}\includegraphics[height =1.15in]{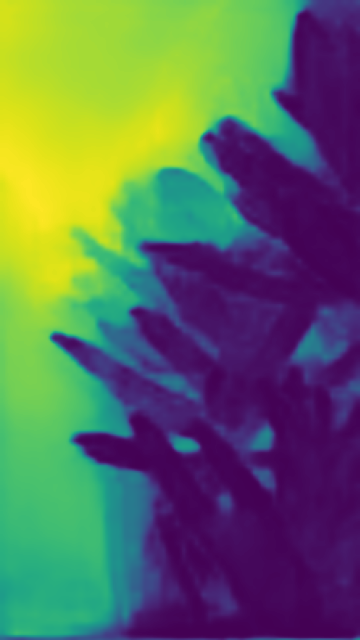}
\\

\hspace{-2.5mm}\includegraphics[height =1.15in]{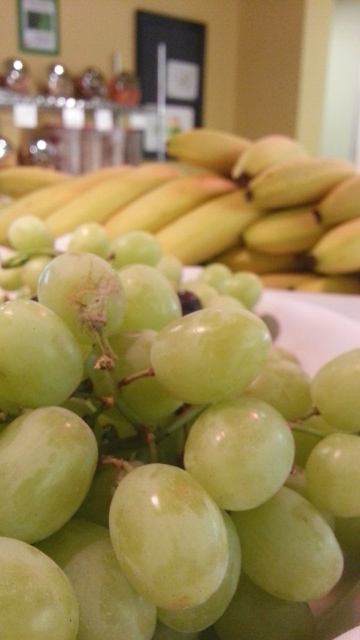} &
\hspace{-2.5mm}\includegraphics[height =1.15in]{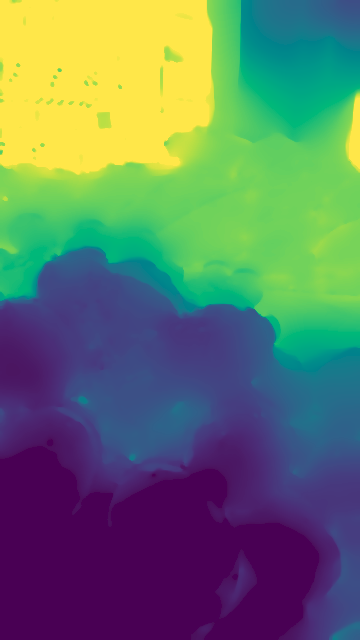} &
 \hspace{-2.5mm}\includegraphics[height =1.15in]{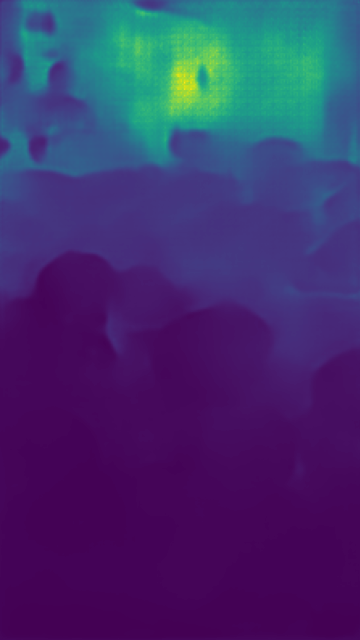}&
\hspace{-2.5mm}\includegraphics[height =1.15in]{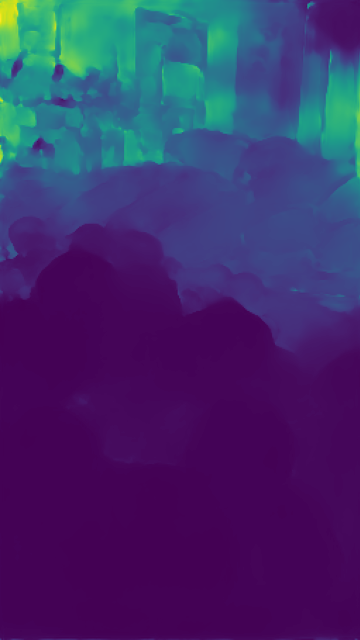} &
\hspace{-2.5mm}\includegraphics[height =1.15in]{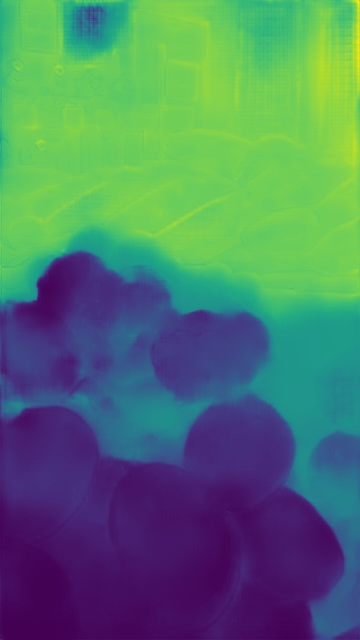}& 
\hspace{-2.5mm}\includegraphics[height =1.15in]{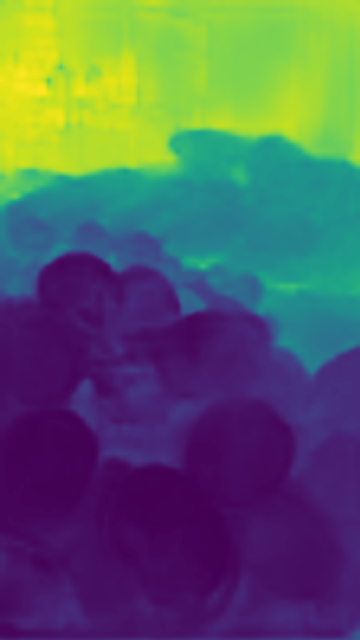}& 
\hspace{-2.5mm}\includegraphics[height =1.15in]{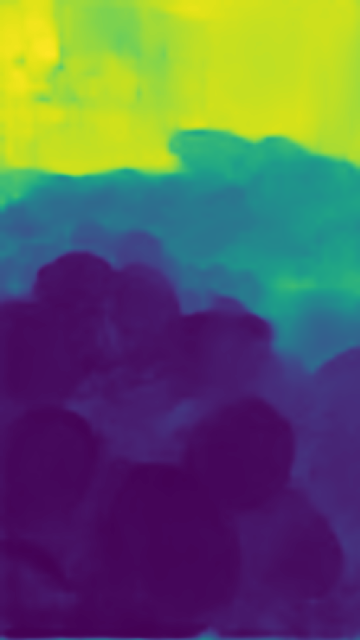}
\\

\hspace{-2.7mm}\includegraphics[height =0.63in]{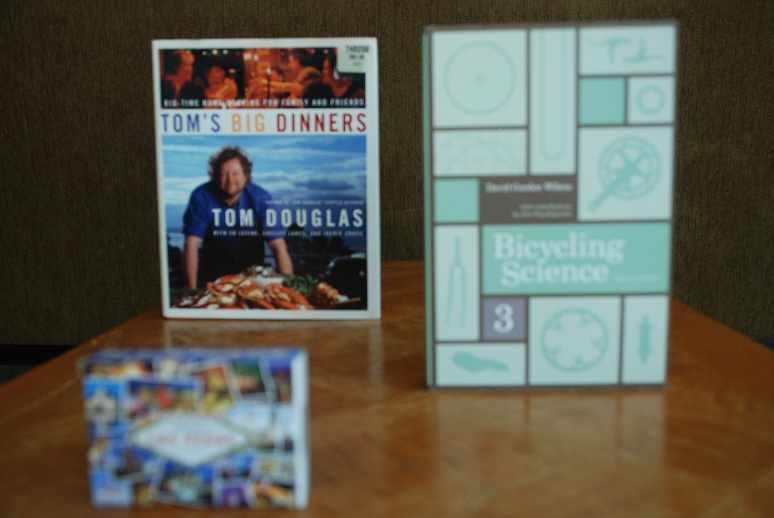} &
\hspace{-2.7mm}\includegraphics[height= 0.63in]{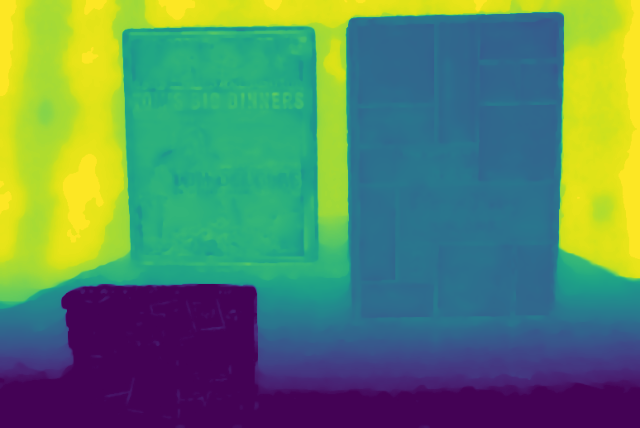} &
\hspace{-2.7mm}\includegraphics[height=0.63in]{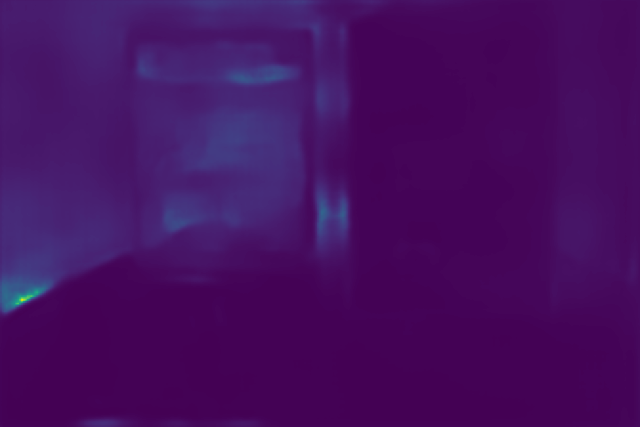}&
\hspace{-2.7mm}\includegraphics[height= 0.63in]{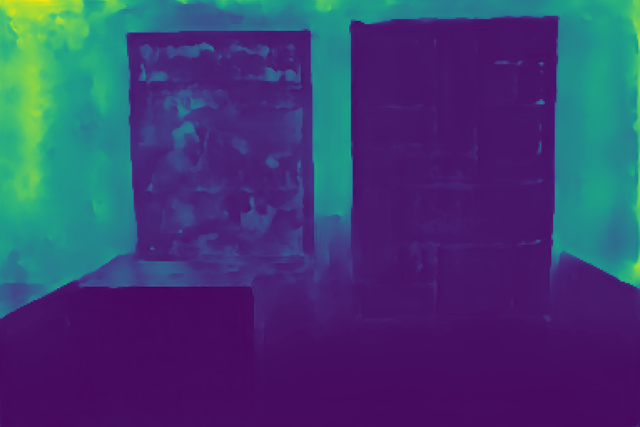}&
\hspace{-2.7mm}\includegraphics[height= 0.63in]{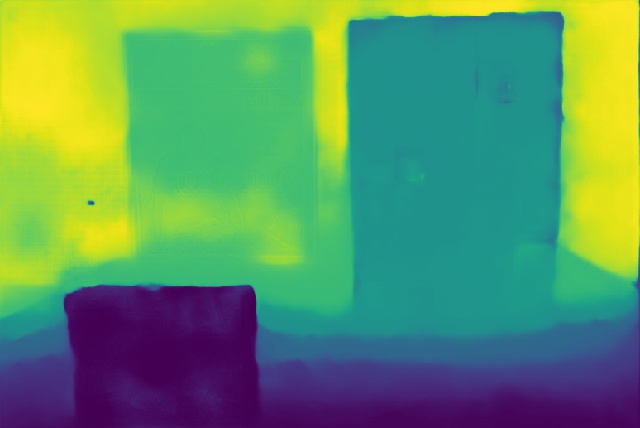}&
\hspace{-2.7mm}\includegraphics[height= 0.63in]{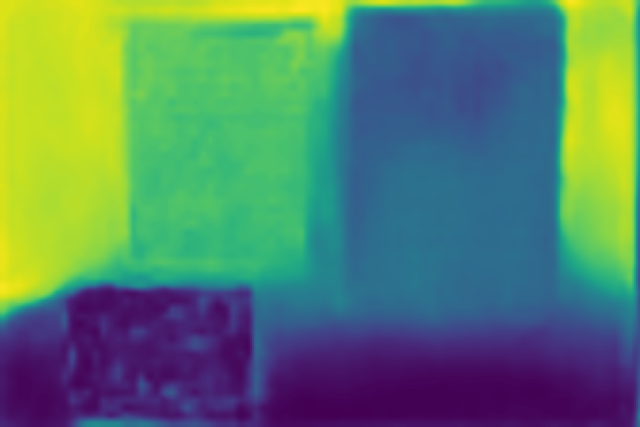}&
\hspace{-2.7mm}\includegraphics[height=0.63in]{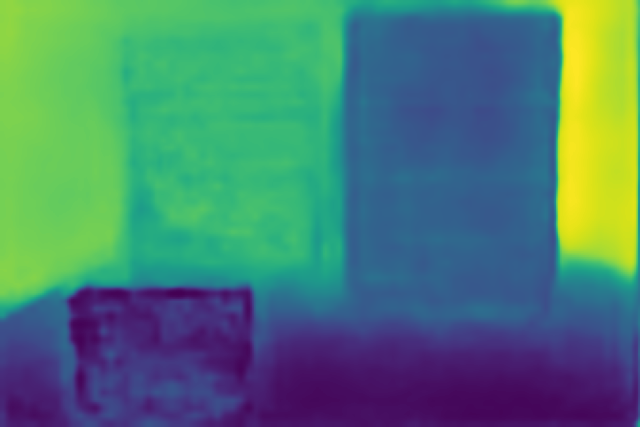}\\

\hspace{-2.7mm}\includegraphics[height =0.63in]{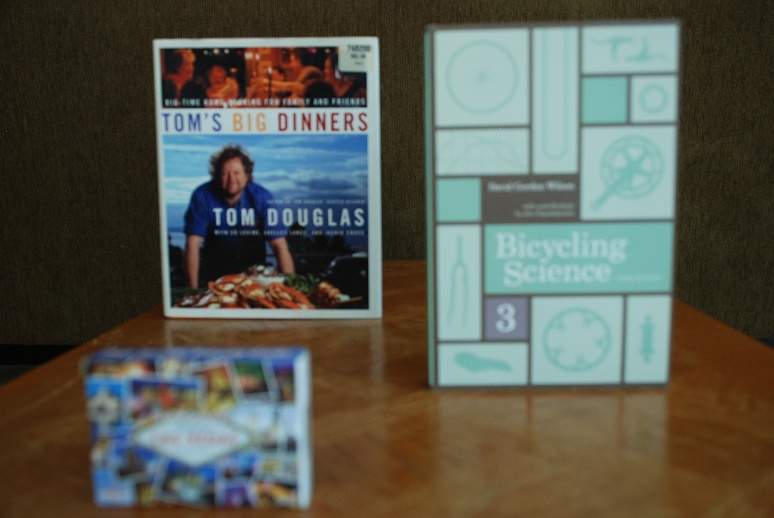} &
\hspace{-2.7mm}\includegraphics[height=0.63in]{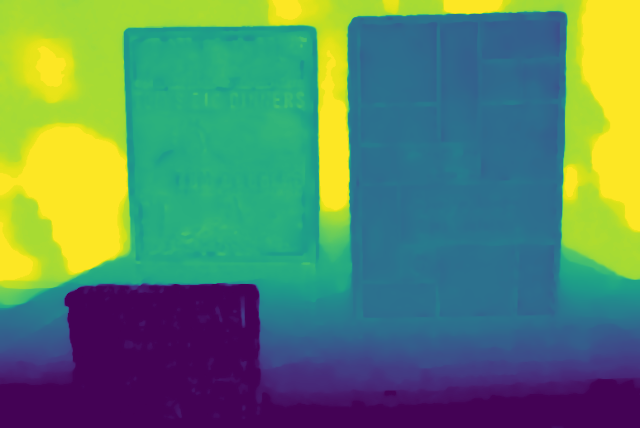} &
\hspace{-2.7mm}\includegraphics[height=0.63in]{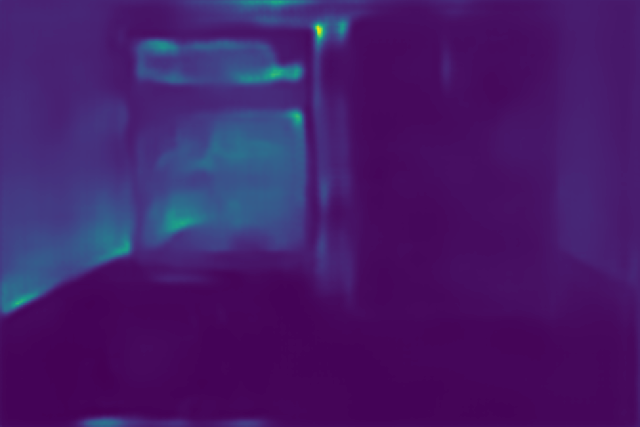}&
\hspace{-2.7mm}\includegraphics[height=0.63in]{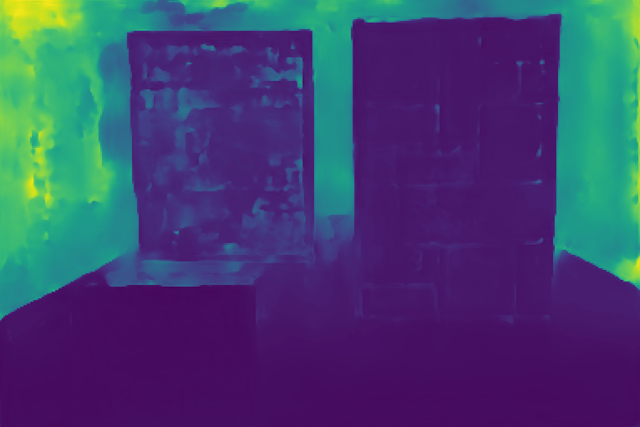}&
\hspace{-2.7mm}\includegraphics[height=0.63in]{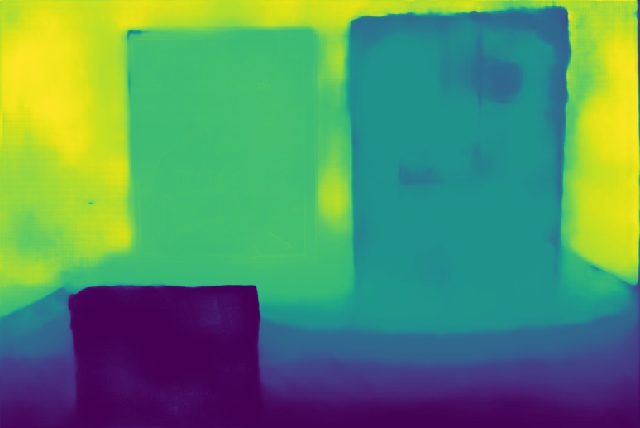}&
\hspace{-2.7mm}\includegraphics[height=0.63in]{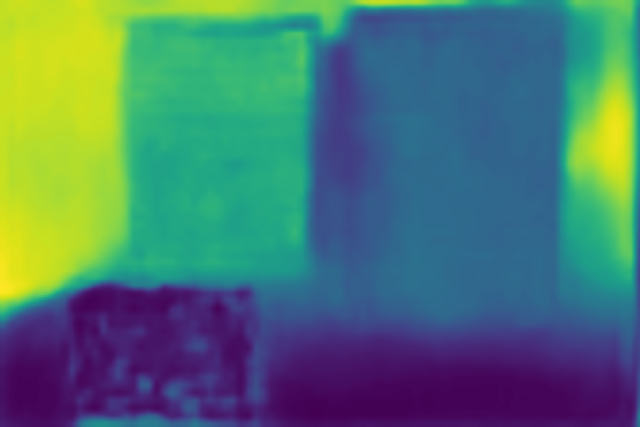}&
\hspace{-2.7mm}\includegraphics[height=0.63in]{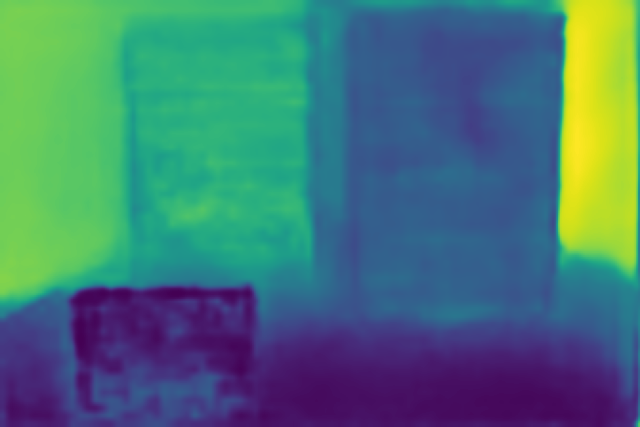}\\

\hspace{-2.7mm}\includegraphics[height =0.63in]{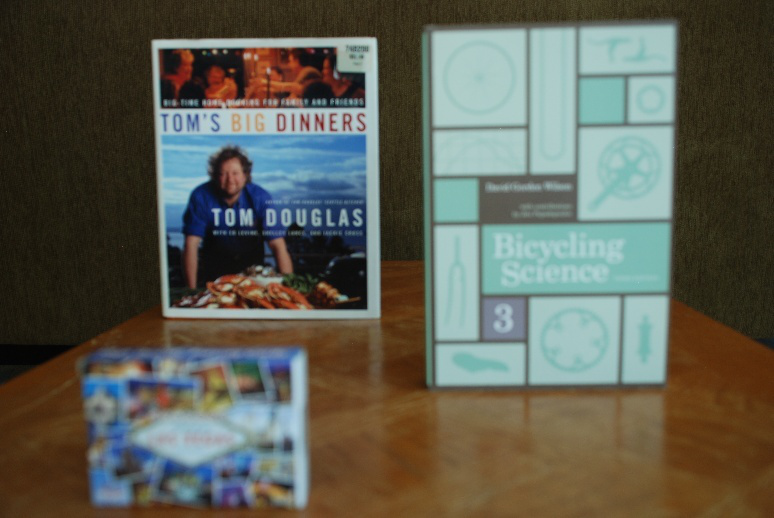} &
\hspace{-2.7mm}\includegraphics[height=0.63in]{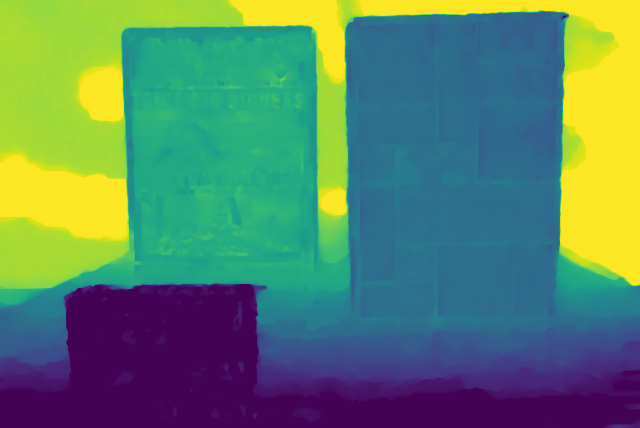} &
\hspace{-2.7mm}\includegraphics[height=0.63in]{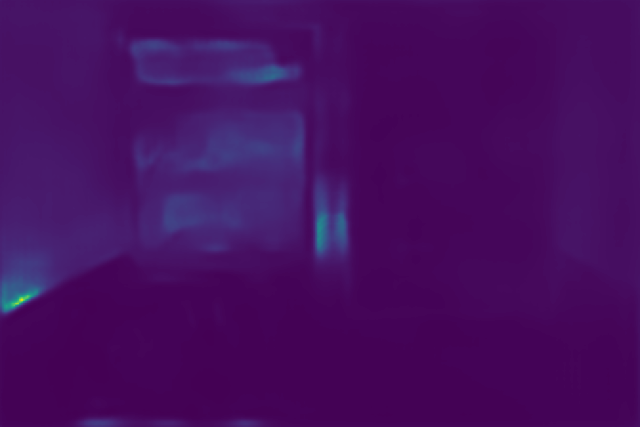}&
\hspace{-2.7mm}\includegraphics[height=0.63in]{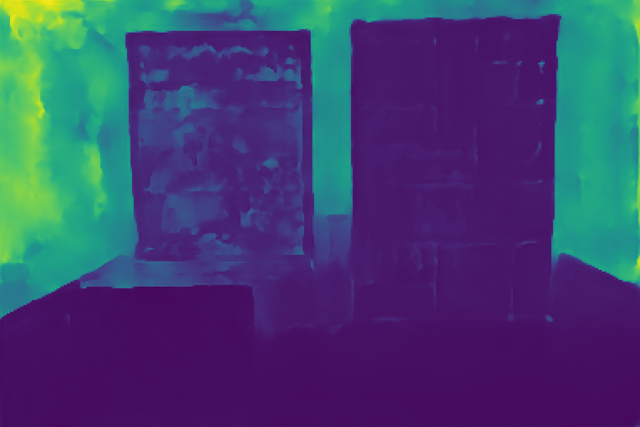}&
\hspace{-2.7mm}\includegraphics[height=0.63in]{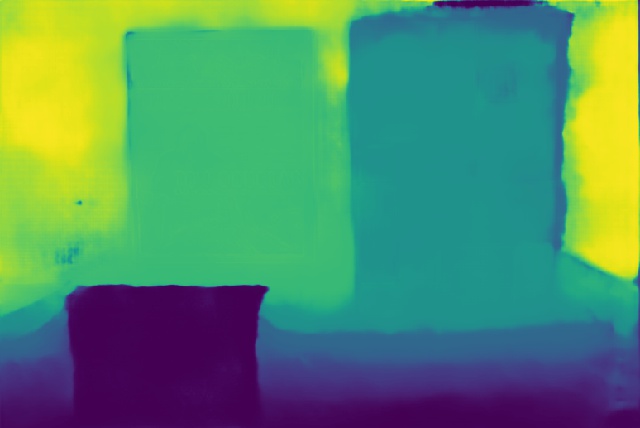}&
\hspace{-2.7mm}\includegraphics[height=0.63in]{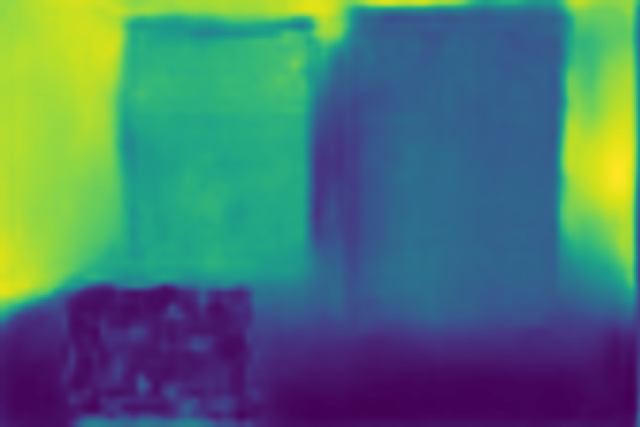}&
\hspace{-2.7mm}\includegraphics[height=0.63in]{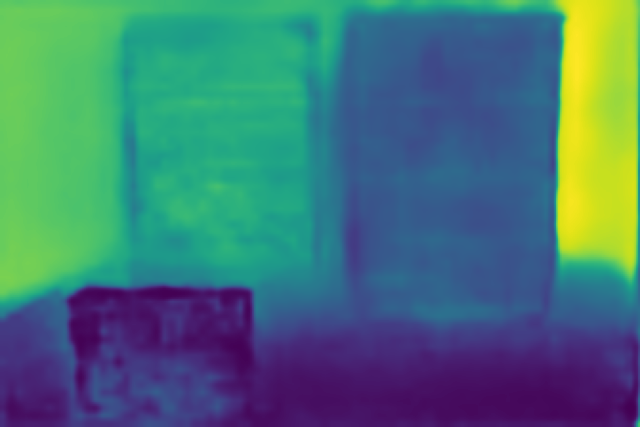}\\

\end{tabular}
\caption{Qualitative results on Mobile depth dataset. The warmer the color, the larger the depth value.}
\label{fig:mobileDFF}
\end{figure*}

Figure~\ref{fig:mobileDFF} illustrates the rest of the results on the aligned scenes of the Mobile depth dataset. Differing from the other samples, the focal stacks in rows 6-8 are taken from the same scenes with different camera motions (zero, small, and large), therefore have slightly different frame alignment. We refer readers to~\cite{suwajanakorn2015mobiledepth} for more details of this dataset. From the results, we can see that all deep methods generalize well in these scenes without any fine-tuning. In rows 6-8, even though some bumpy predictions can be observed in the books regions, all deep methods present reasonable and consistent depth estimation regardless of the alignment difference, respectively, which shows a certain degree of robustness to the alignment. Compared to other deep methods,  Ours-DFV consistently provides more smooth estimations with better boundary preservation, such as the front objects in the first five rows. %than the others

\subsection{Effect of Focal Stack Size}
\label{sec:focal_size_sec}
Most traditional DFF methods~\cite{nayar1994SFF,pertuz2013overview,moeller2015variational,surh2017noise, jeon2019ring} focus on finding the best-focused pixels in the given focal stack and are restricted to the frame-level accuracy for focus analysis. The input focal stack size can greatly affect their depth estimation accuracy. To maintain good accuracy, those methods usually take 10 - 30 frames per stack as input. In contrast, our methods estimate the best-focus distribution and can achieve sub-frame accuracy. This characteristic helps our model to deliver accurate depth estimation with fewer frames. 

To study the effect of focal stack size on our model, we train Ours-DFV model on DDFF-12 training set and test it on its validation set with different stack sizes, $N=2, ..., 10$. We also retrain DDFF~\cite{hazirbas2018DFFdataset} and DefocusNet~\cite{maximov2020focusondefocus} from scratch on the same training set as references. The reason that we exclude FoD500 dataset in this experiment is because it only has 5-frame focal stacks. 

The evaluation metrics are the same as the experiment metrics in the main text, which are adopted from~\cite{hazirbas2018DFFdataset}. They are MSE, RMS, log RMS, absolute relative (Abs. rel.), squared relative (Sqr. rel.), three accuracy percentages ($\delta$, $\delta^2$, $\delta^3$), bumpiness (Bump.), and  average uncertainty (avgUnc.). The first 8 metrics reflect the estimation accuracy from absolute and relative perspectives,  the Bump. metric evaluates the smoothness of results, and the avgUnc. is proposed by us to compare the prediction confidence between {Ours-CV} and {Ours-DCV}. $avgUnc. = \frac{1}{M}\sum^M_{j=1}\phi_j$, where $\phi_j$ is the uncertainty of the depth estimation of pixel $x_j$.  The lower the value, the higher the confidence. 

Table~\ref{tab:var_nfrm} shows the evaluation result. Ours-DFV is able to provide fairly accurate results with only 3-frame input stacks. The MSE error of 3-frame inputs is only 12.9\% higher than the best case (9-frame inputs), and already outperforms DDFF and DefocusNet which take 5 frames as input.  In general, the model delivers more accurate estimations as the input stack size increases. It is evident by that the best performances in terms of the first eight accuracy metrics are all achieved by the models with $N=\{8, 9, 10\}$ frames. We believe the fluctuation is due to the random process at the training time.  As the frame number increases, the model also turns to be more confident in terms of the average uncertainty, in the cost of the runtime.  Moreover, for all the cases, the model provides smooth depth predictions in terms of bumpiness. 

However, we do notice that the impact of additional
frames is diminishing as the frame number increases. DDFF~\cite{hazirbas2018DFFdataset} also reports that their network performance on DDFF-12 dataset stops improving after the frame number reaches $10$, which is the stack size they finally released to the public. Further studies with a new dataset containing more frames per stack are required to find the actual reason, and we leave it for future work.

%-------------------------------------------------------------------------

%-------------------------------------------------------------------------

%-------------------------------------------------------------------------

\end{document}